\documentclass[10pt,twocolumn,letterpaper]{article}

\usepackage[pagenumbers]{iccv} 

\usepackage[accsupp]{axessibility}  
\usepackage{multirow}
\usepackage[dvipsnames]{xcolor} 
\usepackage{graphicx}  
\usepackage{subcaption}  
\usepackage{algorithm}
\usepackage{amsmath}
\usepackage[noend]{algpseudocode}
\usepackage[most]{tcolorbox}
\usepackage{makecell}
\usepackage{booktabs}
\usepackage{diagbox}
\usepackage{marvosym}

\definecolor{iccvblue}{rgb}{0.21,0.49,0.74}
\definecolor{mygreen}{rgb}{0.0, 0.5, 0.0}

\makeatletter
\def\thanks#1{
    \protected@xdef\@thanks{
        \@thanks
        \protect\footnotetext{#1}
    }
}
\makeatother

\newcommand{\downred}[1]{\rlap{\textsubscript{\textcolor{red}{\textbf{$\downarrow$#1}}}}}
\newcommand{\downgreen}[1]{\rlap{\textsubscript{\textcolor{mygreen}{\textbf{$\downarrow$#1}}}}}
\newcommand{\upred}[1]{\rlap{\textsubscript{\textcolor{red}{\textbf{$\uparrow$#1}}}}}
\newcommand{\upgreen}[1]{\rlap{\textsubscript{\textcolor{mygreen}{\textbf{$\uparrow$#1}}}}}

\newcommand{\pmval}[1]{\rlap{\textsuperscript{\ensuremath{\pm #1}}}}
\newcommand{\mypara}[1]{\vspace{0.06cm}\noindent\textbf{#1}\hspace{0.02cm}}
\renewcommand{\vec}[1]{\boldsymbol{#1}}

\newcommand{\ul}[1]{\underline{#1}}

\definecolor{Coral}{RGB}{255,127,80}

\newcommand{\myeg}{\textit{e.g.},\ }

\newcommand{\nextline}{\\}

\usepackage[pagebackref,breaklinks,colorlinks,allcolors=iccvblue]{hyperref}

\def\paperID{10046}
\def\confName{ICCV}
\def\confYear{2025}

\title{Mitigating Object Hallucinations via Sentence-Level Early Intervention}

\author{
    \textbf{Shangpin Peng$^{1*}$ \quad\quad Senqiao Yang$^{2*}\thanks{*~Equal contribution.}$ \quad\quad Li Jiang$^{3}$ \quad\quad Zhuotao Tian$^{1}$}\textsuperscript{\Letter}\thanks{\Letter~Corresponding author (tianzhuotao@hit.edu.cn).}\\[3pt]
    $^1$Harbin Institute of Technology, Shenzhen\\[3pt]
    $^2$The Chinese University of Hong Kong \quad\quad $^3$The Chinese University of Hong Kong, Shenzhen\\
}

\begin{document}
\maketitle

\begin{abstract}
    Multimodal large language models (MLLMs) have revolutionized cross-modal understanding but continue to struggle with hallucinations - fabricated content contradicting visual inputs. Existing hallucination mitigation methods either incur prohibitive computational costs or introduce distribution mismatches between training data and model outputs. We identify a critical insight: hallucinations predominantly emerge at the early stages of text generation and propagate through subsequent outputs. To address this, we propose \textbf{SENTINEL} (\textbf{S}entence-level \textbf{E}arly i\textbf{N}tervention \textbf{T}hrough \textbf{IN}-domain pr\textbf{E}ference \textbf{L}earning), a framework that eliminates dependency on human annotations. Specifically, we first bootstrap high-quality in-domain preference pairs by iteratively sampling model outputs, validating object existence through cross-checking with two open-vocabulary detectors, and classifying sentences into hallucinated/non-hallucinated categories. Subsequently, we use context-coherent positive samples and hallucinated negative samples to build context-aware preference data iteratively. Finally, we train models using a context-aware preference loss (C-DPO) that emphasizes discriminative learning at the sentence level where hallucinations initially manifest. Experimental results show that SENTINEL can reduce hallucinations by over 90\% compared to the original model and outperforms the previous state-of-the-art method on both hallucination benchmarks and general capabilities benchmarks, demonstrating its superiority and generalization ability. The models, datasets, and code are available at \url{https://github.com/pspdada/SENTINEL}.
\end{abstract}
\section{Introduction}
\label{sec:intro}
\begin{figure}
   \centering
   \begin{subfigure}[b]{0.47\textwidth}
      \includegraphics[width=\textwidth]{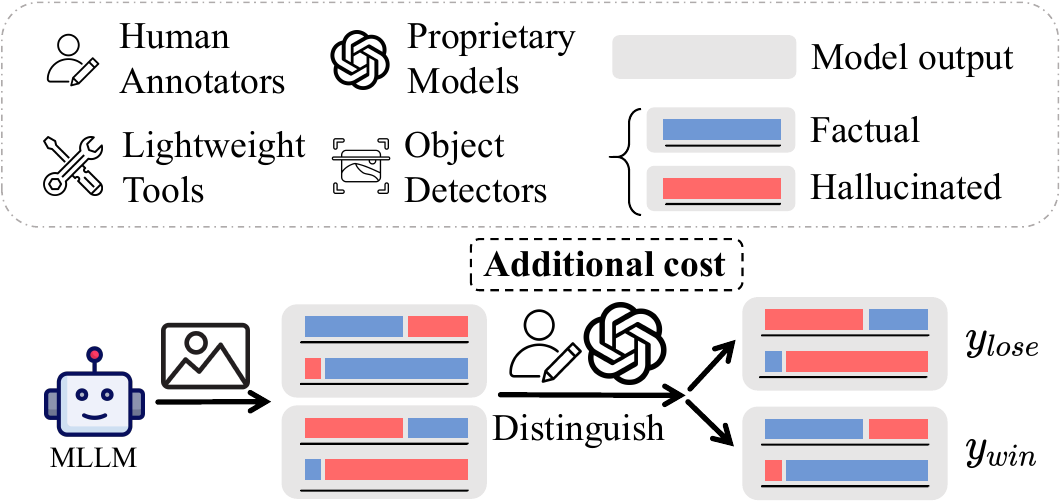}
      \vspace{-12pt}
      \caption{Ultra-large proprietary model/human annotator-dependent methods}
      \label{fig:method_a}
   \end{subfigure}
   \begin{subfigure}[b]{0.47\textwidth}
      \includegraphics[width=\textwidth]{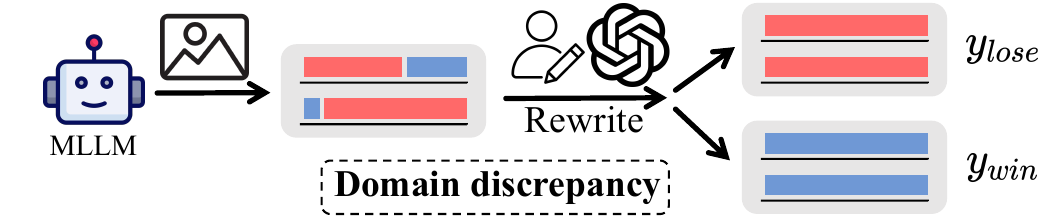}
      \vspace{-12pt}
      \caption{Response rewriting method}
      \label{fig:method_b}
   \end{subfigure}
   \begin{subfigure}[b]{0.47\textwidth}
      \includegraphics[width=\textwidth]{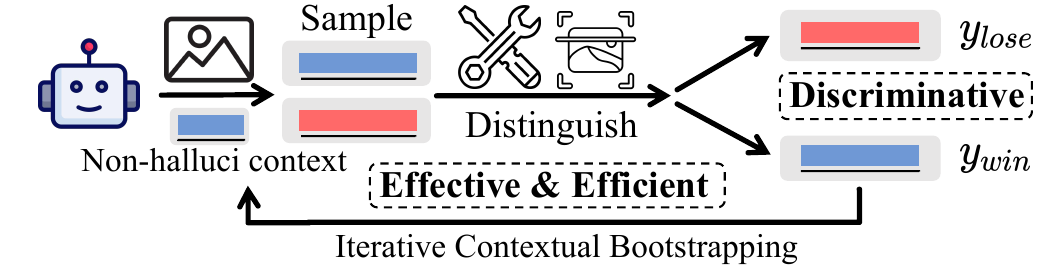}
      \vspace{-12pt}
      \caption{\textbf{SENTINEL} (Ours)}
      \label{fig:method_c}
   \end{subfigure}
   \vspace{-8pt}
   \caption{\textbf{Comparative analysis of data construction strategies for hallucination mitigation in MLLMs.} Our proposed approach demonstrates superior efficiency and effectiveness in generating high-quality, domain-specific preference learning datasets, offering a robust solution for reducing hallucination in MLLMs.}
   \label{fig:method_comparation}
   \vspace{-15pt}
\end{figure}

Recent advancements in multimodal large language models (MLLMs) have demonstrated significant progress in aligning visual and textual representations through cross-modal feature integration, marking a pivotal step toward the development of general-purpose AI systems~\cite{Qwen_VL_2023, Qwen2_VL_2024, LLaVA_v1_2023, LLaVA_v1_5_2024, LLaVA_NeXT_2024, InstructBLIP_2023, OpenAI_GPT4V_2023, MiniGPT_4_2024}. However, a critical challenge persists in multimodal settings: the phenomenon of hallucinations~\cite{HallucinationSurvey_2023, HallucinationSurvey_2024, HallucinationSurvey_202402}, wherein models generate factually inconsistent or fabricated information that deviates from the image content provided by users. This issue not only degrades user trust and experience but also poses substantial risks in real-world applications of MLLMs, thereby impeding the realization of trustworthy general AI systems~\cite{Advancing_Medical_Imaging_2023, Driving_with_LLMs_2024, Embodied_Task_Planning_2023}.

To address this challenge, recent work has explored enhanced decoding strategies~\cite{VCD_2023, OPERA_2024, DoLa_2024} as a means to mitigate hallucinations. While these approaches show promise, they often introduce trade-offs, including increased computational overhead during inference, higher latency, and reliance on specific dependencies, which may limit their scalability and practicality in resource-constrained scenarios.

On the other hand, preference alignment methods~\cite{DPO_2023, Fine_tuning_Factuality_2023, FLAME_2024} avoid additional inference costs but face other challenges. As shown in~\cref{fig:method_a}, many of them rely on large proprietary models (\myeg GPT~\cite{GPT4_2023})~\cite{AMP_2024, FGAIF_2024, HA_DPO_2023, HSA_DPO_2024, POVID_2024, Woodpecker_2023} or human annotators~\cite{RLHF_V_2024, M_HalDetect_2024}, incurring high costs. Additionally,~\cref{fig:method_b} highlights that output rewriting~\cite{HA_DPO_2023, HSA_DPO_2024, POVID_2024} can create distributional discrepancies, while~\citet{Step_DPO_2024} and our experiments in~\cref{tab:rewrite_sample_result} show that out-of-domain training data harms generalization.
\textit{Therefore, the high costs and the distribution disparities inherent in the curated training data may compromise hallucination mitigation efforts.}

\mypara{Key observations.}
To address hallucination with greater efficacy and efficiency, we investigate the dynamics of hallucination within the model's output. Our analysis reveals that hallucination intensity escalates with the length of generated text, while mitigating hallucinations at specific sentences significantly reduces their prevalence in subsequent outputs, as detailed in~\cref{subfig:word_position_distribution,subfig:sentence_frequency_distribution}. These findings suggest that early intervention—targeting hallucinations at their initial occurrence—is crucial to preventing their propagation in later generations. This raises a key question: \textit{How can we effectively implement an early intervention strategy to address hallucinations of MLLMs as they arise?}

\mypara{Our solution.}
In this work, we propose \textbf{SENTINEL} (\textbf{S}entence-level \textbf{E}arly i\textbf{N}tervention \textbf{T}hrough \textbf{IN}-domain pr\textbf{E}ference \textbf{L}earning), which provides early intervention for the initial occurrence of hallucinations during generation. Unlike existing methods, SENTINEL operates without relying on external large language models for rewriting, ensuring that the learning targets remain strictly within the domain of the model's original outputs. This approach preserves the model's intrinsic distribution and expression patterns while effectively curbing hallucination propagation.

Specifically, SENTINEL first employs an in-domain candidate bootstrapping strategy, which performs multiple sampling rounds on the current model, extracts objects from the outputs, and applies consistency cross-checking to classify objects as \textit{hallucinated}, \textit{uncertain}, or \textit{factual}. This is followed by a context-aware preference data generation process, which constructs preference pairs using non-hallucinated positive samples and hallucinated negative ones, enhanced by iterative contextual bootstrapping. Finally, context-aware preference learning is performed using the modified context-aware DPO loss, maximizing the likelihood of generating context-coherent positive samples while minimizing hallucinated negative ones. By focusing on captions where hallucinations first emerge, SENTINEL effectively halts their propagation in subsequent outputs.

Experimental results across various benchmarks demonstrate that SENTINEL effectively mitigates object hallucination while preserving the generalization capabilities of MLLMs. Specifically, on Object Halbench~\cite{Obj_HalBench_2018} and AMBER~\cite{AMBER_2023}, hallucinations are reduced by about 92\% and 65\%, respectively, with consistent improvements on HallusionBench~\cite{HallusionBench_2023}. Furthermore, SENTINEL preserves its performance on VQAv2~\cite{VQA_v2_2017} and TextVQA~\cite{TextVQA_2019}, and achieving decent gains on both ScienceQA~\cite{ScienceQA_2022} and MM-Vet~\cite{MM_Vet_2024}.

To summarize, our contributions are as follows:

\begin{itemize}[leftmargin=*]
  \item We demonstrate that early intervention at the first occurrence of hallucination is crucial for preventing its propagation in subsequent model outputs of MLLMs.
  \item We propose SENTINEL, which effectively and efficiently mitigates hallucinations without requiring extensive external resources or manual effort.
  \item The model-agnostic SENTINEL achieves state-of-the-art performance on hallucination benchmarks without compromising MLLMs' general capabilities.
\end{itemize}

\section{Background and Motivation}
\label{sec:preliminary}

In this section, we briefly introduce the foundational concepts and methods relevant to this study in \cref{subsec:preliminaries}, establishing the necessary background. Following this, in \cref{subsec:motivation}, we outline our key insights and elucidate the motivations behind our proposed designs.

\subsection{Related Work and Preliminaries}
\label{subsec:preliminaries}
Object Hallucination (OH) in Multimodal Large Language Models (MLLMs) is characterized by the generation of text that is semantically coherent yet inconsistent with the visual content of the provided image~\cite{HallucinationSurvey_2023, HallucinationSurvey_2024}. To mitigate this issue, recent advancements have focused on innovative decoding strategies, which aim to reduce the prevalence of OH by refining the generation process of MLLMs~\cite{VCD_2023,DoLa_2024,OPERA_2024,HALC_2024}.

Concurrently, preference learning has emerged as an alternative approach for addressing OH, leveraging its capacity to align MLLMs with human expectations for truthfulness and traceability~\cite{LRV_Instruction_2024, M_HalDetect_2024, Silkie_2023}. Notably, the Proximal Policy Optimization algorithm (PPO)~\cite{PPO_2017} enhances model reliability by training an auxiliary reward model to assess response quality and then guide the model in optimizing its outputs based on the reward signals. Moreover, Direct Preference Optimization (DPO)~\cite{DPO_2023} has emerged as a simpler alternative, learning directly from pre-collected feedback without requiring a reward model. The DPO loss is:

\vspace{-10pt}
\begin{equation}
    \begin{aligned}
        \mathcal{L}_{\text{DPO}}(\vec{\theta}) \! =\! & -\mathbb{E}_{(\vec{x},\vec{y}_{w},\vec{y}_l) \sim D}\! \Biggl[
        \log \sigma \Bigl(\beta \log \frac{\pi_{\vec{\theta}}(\vec{y}_{w}|\vec{x})}{\pi_{\text{ref}}(\vec{y}_{w}|\vec{x})}                                                                        \\
                                                      & \qquad\qquad\qquad\qquad - \beta \log \frac{\pi_{\vec{\theta}}(\vec{y}_l|\vec{x})}{\pi_{\text{ref}}(\vec{y}_l|\vec{x})} \Bigr) \!\Biggr], \\
        \text{where } \vec{x}\! =                     & [\vec{v},\vec{q}].
    \end{aligned}
    \label{eq:dpo}
\end{equation}
\vspace{-10pt}

Here, $D$ represents the preference dataset for learning, $\sigma$ denotes the sigmoid function, $\pi_{\vec{\theta}}$ indicates the policy model under training, $\pi_{\text{ref}}$ represents the unchanged reference model, $\vec{y}_w$ stands for the positive sample, and $\vec{y}_l$ represents the negative sample, both based on the input $\vec{x}$, which includes the image $\vec{v}$ and prompt $\vec{q}$.
The hyperparameter $\beta$ governs the separation between the policy model and the reference model.
Many recent methods~\cite{HA_DPO_2023,HSA_DPO_2024,RLAIF_V_2024,Topic_level_SC_2024,POVID_2024} leverage DPO to mitigate hallucinations by curating preference data to guide the models. More related works of this study are discussed in~\cref{sec:supp_related_works}.

\subsection{Motivation}
\label{subsec:motivation}
This section outlines the motivations behind this work. The implementation details of related experiments are provided in~\cref{sec:motivation_details}.

\begin{figure}[t]
    \centering
    \captionsetup[subfigure]{}
    \begin{subfigure}{0.16\textwidth}
        \centering
        \includegraphics[width=\linewidth]{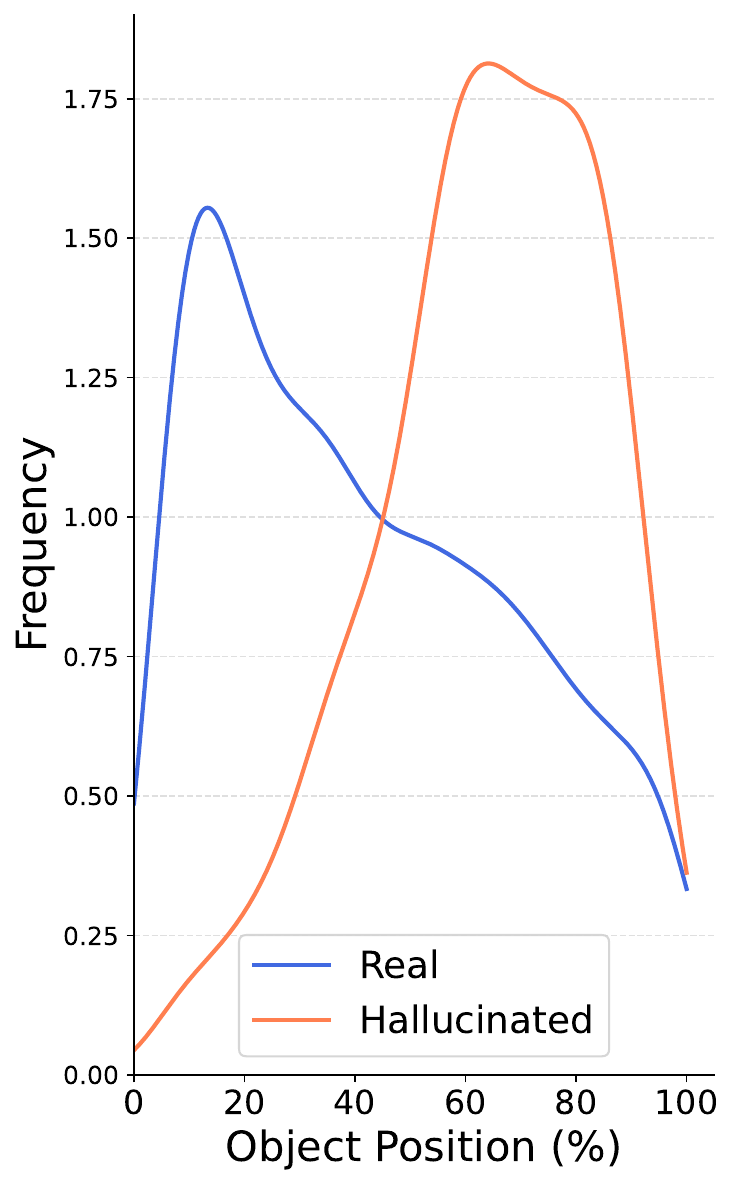}
        \caption{}
        \label{subfig:word_position_distribution}
    \end{subfigure}
    \hfill
    \begin{subfigure}{0.31\textwidth}
        \centering
        \includegraphics[width=\linewidth]{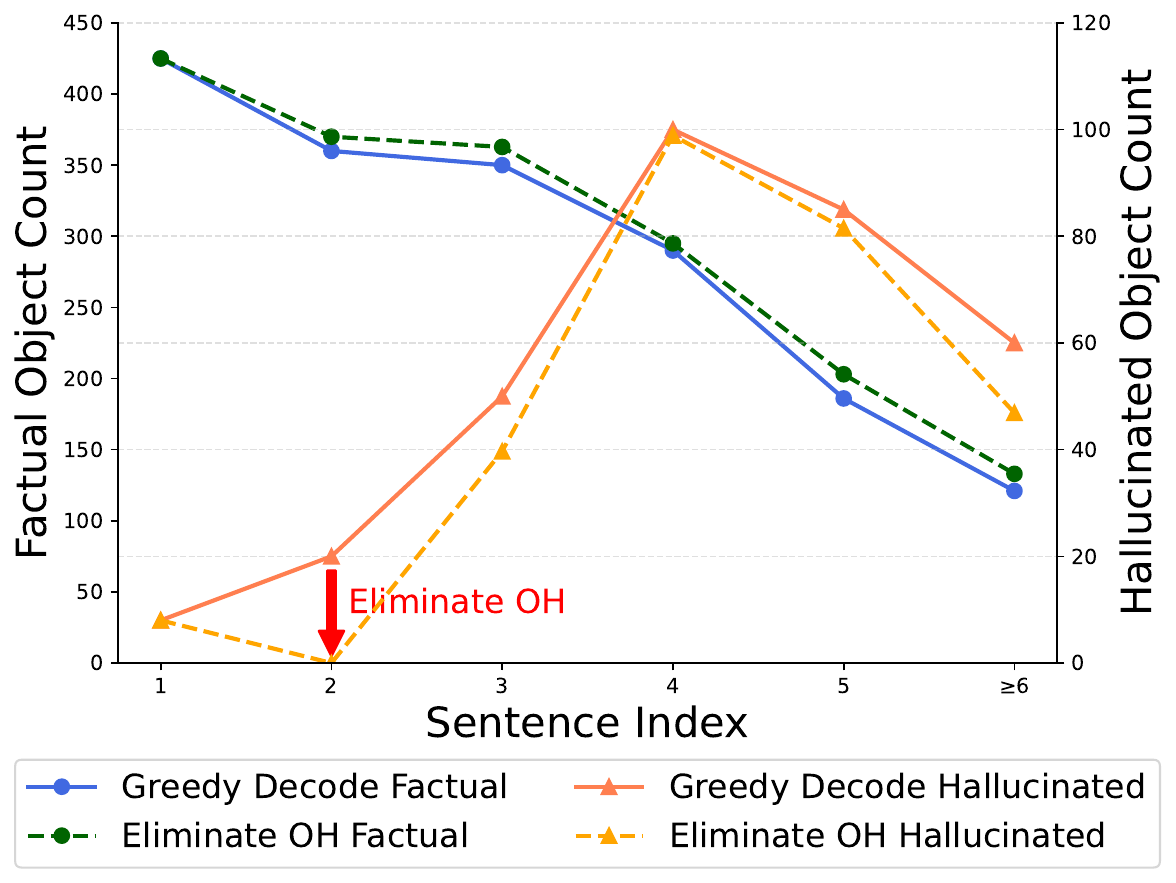}
        \caption{}
        \label{subfig:sentence_frequency_distribution}
    \end{subfigure}
    \vspace{-20pt}
    \caption{\textbf{Object position distribution in MLLM hallucination analysis.} (a) illustrates the progressive deterioration of hallucination effects in Multimodal Large Language Models (MLLMs) with increasing description length in the image captioning task, while (b) demonstrates the effectiveness of early-stage intervention in mitigating the propagation of hallucination.}
    \label{fig:object_position_distribution}
    \vspace{-10pt}
\end{figure}

\mypara{Hallucination grows with text length.}
To better understand the causes of Object Hallucination (OH), we analyze the distributions of hallucinated and factual objects in image captions generated by MLLMs.
Specifically, as shown in~\cref{subfig:word_position_distribution}, where the horizontal axis represents the normalized position of an object in the caption (as a percentage), while the vertical axis denotes the normalized frequency (probability density), the \textcolor{RoyalBlue}{blue curve} corresponds to objects present in the image, and the \textcolor{Coral}{orange curve} represents hallucinated objects. The comparison reveals that as caption length increases, the model becomes more prone to hallucinations, with fewer factual objects described and more hallucinated ones introduced. This trend is further corroborated by sentence-level analysis in~\cref{subfig:sentence_frequency_distribution}.
These findings lead us to hypothesize that \textit{intervening at the initial occurrence of hallucination could be critical in reducing its recurrence in subsequent model outputs}.

\mypara{Early intervention mitigates hallucinations.}
To evaluate the effectiveness of early intervention in curbing hallucination propagation, we analyze the impact of addressing hallucinations at the sentence level in image captioning tasks. Specifically, as illustrated in~\cref{subfig:sentence_frequency_distribution}, eliminating hallucinated objects in the second sentence—compared to vanilla greedy decoding—significantly reduces the likelihood of hallucinated objects in subsequent sentences while increasing the probability of factual objects present in the image. Similar results are observed when addressing hallucinations in the third sentence, as shown in~\cref{subsec:decode_based_early_intervention}. These findings underscore the necessity of early intervention to mitigate hallucinations effectively.

To enable early intervention, an open-vocabulary object detector~\cite{Grounding_DINO_2024,YOLO_World_2024} could be employed during inference to verify the presence of the objects generated by the model within the image. While this method effectively reduces hallucinations without sacrificing caption diversity, as demonstrated in~\cref{subsec:decode_based_early_intervention}, it is time-consuming; despite the object detector being efficient, the model's sampling process incurs significant computational overhead.

Consequently, we opt for a preference learning strategy during model training, which mitigates hallucinations without compromising the original inference efficiency.

\begin{figure*}[t]
    \centering
    \includegraphics[width=\textwidth]{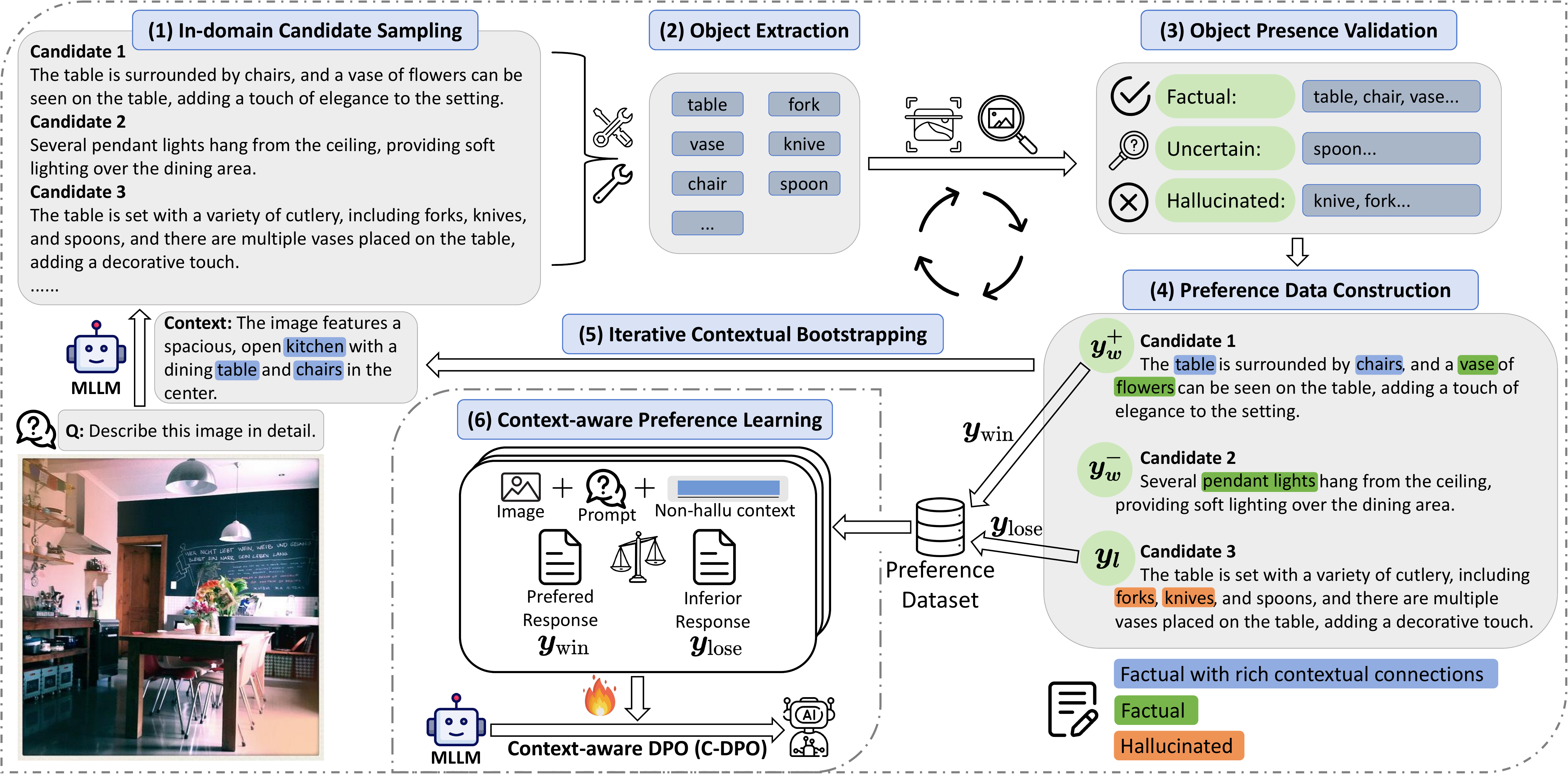}
    \vspace{-15pt}
    \caption{
        \textbf{The overview of SENTINEL.} The proposed SENTINEL takes six essential steps:
        \textbf{(1)}
        Generate multiple in-domain responses conditioned on the input image, prompt, and context $\vec{c}$.
        \textbf{(2)}
        Identify and extract all mentioned objects from each generated sentence.
        \textbf{(3)}
        Utilizing two object detectors to validate the existence of extracted objects through cross-referencing.
        \textbf{(4)}
        Categorize generated sentences into hallucinated and non-hallucinated groups based on detection results.
        \textbf{(5)}
        Extend the generation context with verified non-hallucinated sentences to guide subsequent outputs.
        \textbf{(6)}
        Fine-tune the model using the context-aware DPO (C-DPO) loss with the in-domain, style-consistent, and context-varying preference data.
    }
    \label{fig:method}
    \vspace{-10pt}
\end{figure*}

\section{Method}
\label{sec:method}

\subsection{Overview}
\label{sec:overview}
Existing preference learning methods may use an external model to rewrite sentences or rely on model-generated responses as training data. However, these methods may introduce discrepancies in distribution and expression patterns between the training data and the model's original output. Hence, we propose SENTINEL, which performs sentence-level early intervention to mitigate object hallucinations through preference learning with in-domain data, without manual effort or dependence on extensive LLMs.

As shown in~\cref{fig:method}, the proposed SENTINEL method takes six essential steps. Specifically, \cref{subsec:candidate-gen} presents the process of generating the in-domain candidates containing the factual and hallucinated objects. Subsequently, \cref{subsec:data-prepare} introduces the construction of preference data pairs derived from these in-domain candidates. These two steps can be integrated into the In-domain Preference Data Construction phase (shown in \cref{alg:gen_data}). Finally, in~\cref{sec:sentence-level-training} we elaborate on how SENTINEL leverages the curated preference data to achieve preference learning.

\subsection{In-domain Candidate Bootstrapping}
\label{subsec:candidate-gen}
To construct positive and negative preference data pairs without relying on external models for rewriting, we perform multiple sampling rounds on the current model and extract objects from the outputs. We then apply a consistency cross-checking method to classify the model's output objects into three categories: \textit{hallucinated}, \textit{uncertain}, and \textit{factual}, which are used to construct preference data in subsequent steps. This process is termed In-domain Candidate Booststrapping, as illustrated in~\cref{fig:method} (1)-(3).

\mypara{In-domain candidate sampling.}
In our approach, we use sampling-based decoding to obtain $n$ candidate samples. This ensures that the positive ($\vec{y}_w$) and negative ($\vec{y}_l$) samples are drawn from the same distribution as the current model, preserving consistency in textual styles and linguistic structures. The generation halts upon sentence completion (e.g., detection of a period), at which point sentences are automatically segmented for subsequent discrimination.

\mypara{Object extraction.}
After generating candidate sentences, we extract the mentioned objects from the text for hallucination detection. To achieve this, we utilize the SceneGraphParser~\cite{FACTUAL_2023} model to transform the textual descriptions into a series of triplet-based scene graphs. By parsing these scene graphs, we identify specific noun entities from the subjects and objects, which are subsequently used as candidate objects for existence verification.

\mypara{Object presence validation.}
Following object extraction, we apply cross-checking to validate the presence of candidate objects in the image. Specifically, we utilize two open-vocabulary object detectors, GroundingDINO~\cite{Grounding_DINO_2024} and Yolo World~\cite{YOLO_World_2024}, for cross-validation. This approach demonstrates superior performance compared to using a single detector, as shown in~\cref{fig:data_scale} of the ablation study.

The cross-checking results are categorized into three types: (1) \textit{hallucinated} (both models confirm absence), (2) \textit{factual} (both models confirm presence), and (3) \textit{uncertain} (conflicting results). Sentences containing hallucinated objects are tagged as ``\textit{hallucinated}'', whereas those only containing factual objects are tagged as ``\textit{non-hallucinated}'', forming positive-negative sample pairs for preference learning.
To ensure data quality and minimize detector bias, we ignore uncertain objects.

\begin{algorithm}[ht]
    \footnotesize
    \centering
    \caption{In-domain Preference Data Construction}
    \begin{algorithmic}[1]
        \Statex \textbf{Input:} Image $\vec{v}$, prompt $\vec{q}$, context $\vec{c}$ (initially empty)
        \Statex \textbf{Output:} Training samples $(\vec{v}, \vec{q}, \vec{c}, \vec{y}^{+}_{w}, \vec{y}_{l})$
        \Statex
        \While{Model $M$ does not generate $\texttt{</s>}$}
        \State Sample $n$ in-domain candidates  $s_i$ using $\vec{v}$, $\vec{q}$, and $\vec{c}$
        \For{each sample $s_i$}
        \State Extract entities from the sample
        \State Validate the presence of entities using object detectors
        \EndFor
        \State Select $\vec{y}^{+}_{w}$ as context-coherent non-hallucinated sample
        \State Select $\vec{y}_{l}$ as hallucinated sample
        \State Construct preference samples $(\vec{v}, \vec{q}, \vec{c}, \vec{y}^{+}_{w}, \vec{y}_{l})$
        \State Append a non-hallucinated sample $\vec{y}^{+}_{w}$ to the context $\vec{c}$
        \EndWhile
    \end{algorithmic}
    \label{alg:gen_data}
\end{algorithm}

\subsection{Context-aware Preference Data Generation}
\label{subsec:data-prepare}
With sentences labeled as ``\textit{hallucinated}'' or ``\textit{non-hallucinated}'' from \cref{subsec:candidate-gen}, this section introduces context-aware preference data generation.
As illustrated in~\cref{fig:method} (4)-(5), this process extracts contextually relevant data, ensuring the training data better represents the model's output distribution. The specifics are elaborated below.

\mypara{Preference data construction.}
The preference data is typically composed of the image, the corresponding prompt, the positive sample, the negative sample, and the context (i.e., all generated sentences excluding the current one). In the construction of sample pairs, positive samples $\vec{y}_w$ are selected from the \textit{non-hallucinated} sentences, while negative samples $\vec{y}_l$ are derived from the \textit{hallucinated} sentences.

\begin{figure}[t]
    \centering
    \includegraphics[width=0.47\textwidth]{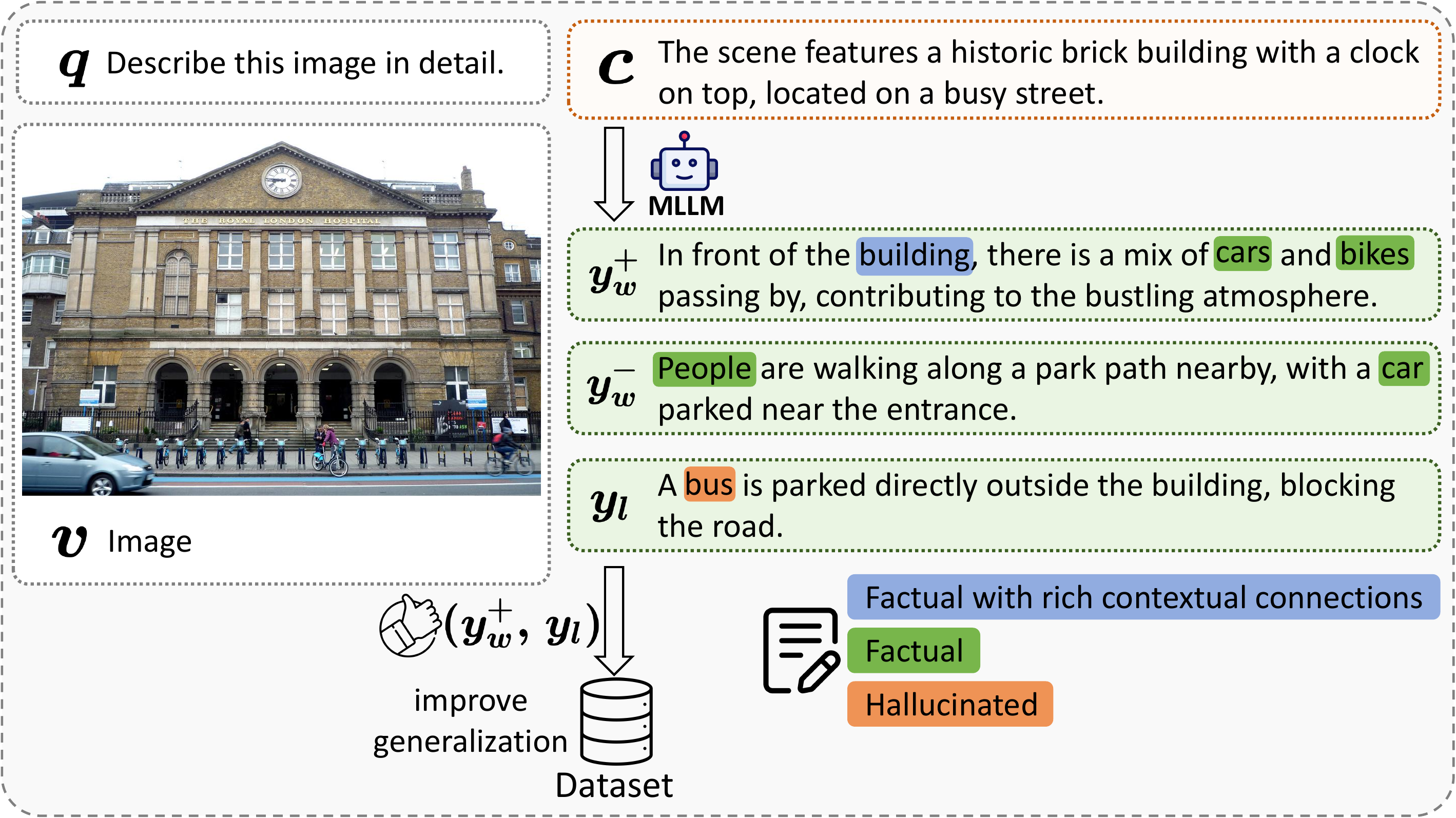}
    \vspace{-5pt}
    \caption{\textbf{Categories of in-domain candidates.} The in-domain candidates fall into three types. Employing non-hallucinated, context-coherent descriptions ($\vec{y}^{+}_{w}$) as positive samples, paired with hallucinated descriptions ($\vec{y}_l$), enhances the model's generalization performance and robustness.}
    \label{fig:type_of_y_win}
    \vspace{-10pt}
\end{figure}

Subsequently, we partition the positive samples $\vec{y}_w$ into two categories: (1) the context-coherent positive sample $\vec{y}^{+}_{w}$, wherein some of the described objects are explicitly referenced in the context, and (2) the context-agnostic positive sample $\vec{y}^{-}_{w}$, where none of the objects are mentioned in the context.
In essence, the objects described in $\vec{y}^{+}_{w}$ exhibit a strong correlation with the context, while those in $\vec{y}^{-}_{w}$ display a weaker or negligible correlation. Illustrative examples are provided in~\cref{fig:method,fig:type_of_y_win}.

We observe that the context-coherent sample $\vec{y}^{+}_{w}$ can effectively mitigate hallucinations without compromising the model's generalization capabilities, and incorporating $\vec{y}^{-}_{w}$ as the positive samples results in performance reduction, as shown in~\cref{tab:different_yw_type}. This observation underscores the importance of contextual signals in guiding the model's generation process. Specifically, the richer contextual information in $\vec{y}^{+}_{w}$ samples appears to enhance the model's ability to preserve contextual coherence and prioritize salient content, resulting in performance improvements~\cite{Skipn_2024}.

\begin{figure}[t]
    \centering
    \includegraphics[width=0.47\textwidth]{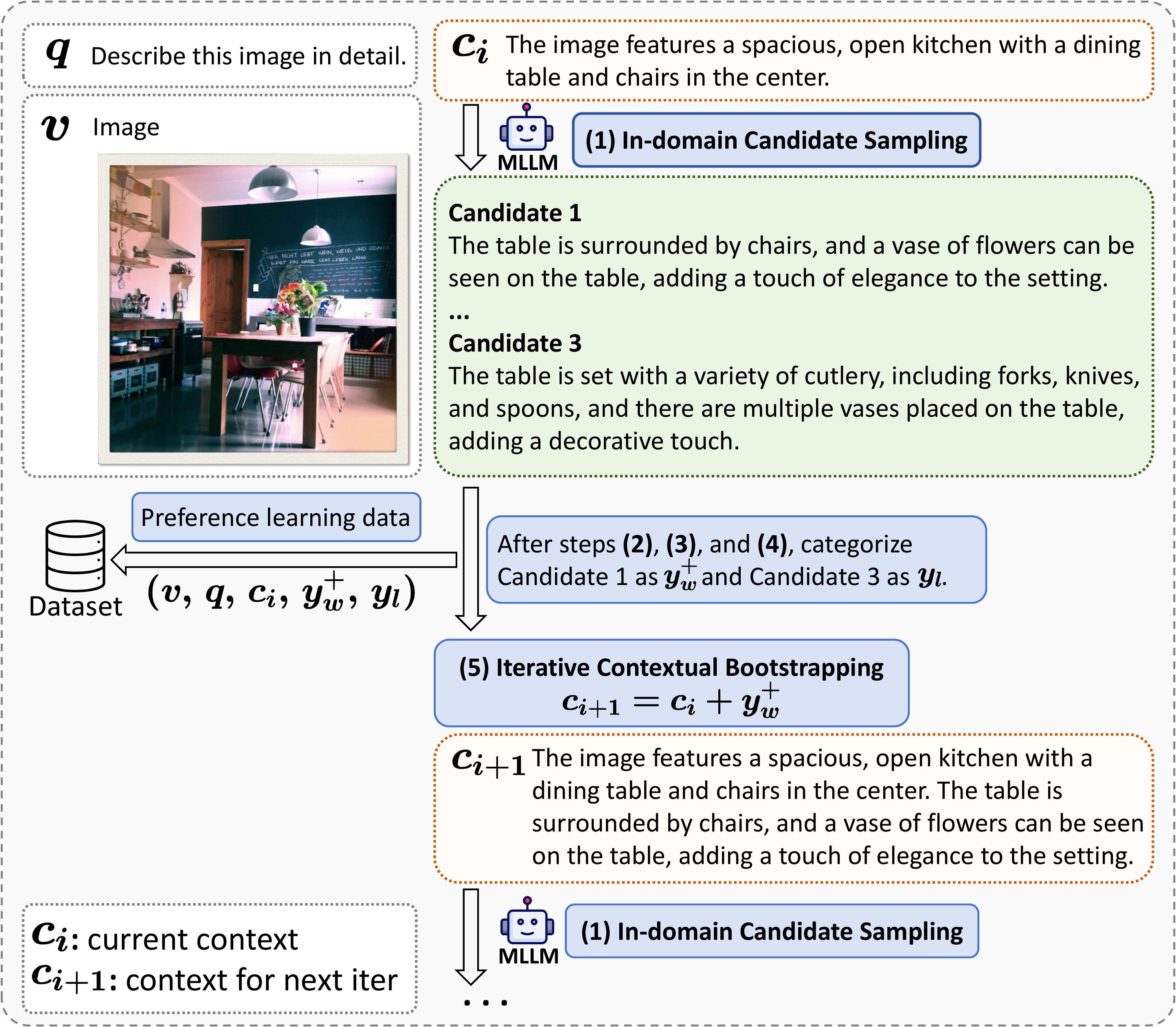}
    \vspace{-5pt}
    \caption{\textbf{Visualization of the Iterative Contextual Bootstrapping (ICB) framework.} Given an input image and corresponding question, this pipeline iteratively generates diverse contextual samples, enabling robust hallucination mitigation across varying contexts and significantly improving model generalization.}
    \label{fig:iteration}
    \vspace{-10pt}
\end{figure}

\mypara{Iterative Contextual Bootstrapping (ICB).}
The proposed SENTINEL framework is designed to enable early intervention for mitigating hallucinations in generative models. Given the context $\vec{c}$, which represents the hallucination-free content preceding the current output, the model is trained to distinguish between a non-hallucinated positive sample $\vec{y}^{+}_{w}$ and a hallucinated negative sample $\vec{y}_{l}$. To enhance robustness across diverse contexts, we introduce the Iterative Contextual Bootstrapping (ICB) strategy, as depicted in~\cref{fig:iteration}.

Specifically, given the query $\vec{q}$, the input image $\vec{v}$, and the current context $\vec{c}_i$, we generate multiple candidate outputs by repeatedly sampling from the MLLM. These candidates are then processed through a structured pipeline consisting of (2) object extraction, (3) object presence validation, and (4) preference data construction, as illustrated in~\cref{fig:method}. This pipeline is designed to identify a non-hallucinated positive sample $\vec{y}^{+}_{w}$ and a hallucinated negative sample $\vec{y}_{l}$. By aggregating $\vec{v}, \vec{q}, \vec{c}_i, \vec{y}^{+}_{w}$ and $\vec{y}_{l}$, we construct a preference data pair $(\vec{v}, \vec{q}, \vec{c}_i, \vec{y}^{+}_{w},\vec{y}_{l})$, which is subsequently appended to the dataset for preference learning.

Furthermore, to bootstrap the preference data with different hallucination-free contexts, we construct $\vec{c}_{i+1}=\vec{c}_i + \vec{y}^{+}_{w}$ for the next iteration by appending the positive sample $\vec{y}^{+}_{w}$ to the current context $\vec{c}_i$. The updated context $\vec{c}_{i+1}$ is then processed through the same procedure as described above to generate a new preference data pair. This iterative approach ensures that the preference data is enriched with progressively more complex and varied contexts, enabling the model to generalize its hallucination mitigation capabilities across different scenarios.
The effectiveness of this pipeline is validated and discussed in~\cref{subsec:iterative_contextual_booststrapping}.
\begin{table*}
  \centering
  \resizebox{\textwidth}{!} {%
    \begin{tabular}{ll cccccc|cccc}
      \toprule
      \multirow{3}{*}{\vspace{-2mm}\textbf{Model}}                            &
      \multirow{3}{*}{\vspace{-2mm}\textbf{Method}}                           &
      \multicolumn{6}{c|}{\textbf{Hallucination benchmarks}}                  &
      \multicolumn{4}{c}{\textbf{General benchmarks}}
      \\
                                                                              &
                                                                              &
      \multicolumn{2}{c}{\textbf{Object HalBench}~\cite{Obj_HalBench_2018}}   &
      \multicolumn{3}{c}{\textbf{AMBER}~\cite{AMBER_2023}}                    &
      \multicolumn{1}{c|}{\textbf{HallusionBench}~\cite{HallusionBench_2023}} &
      \multicolumn{1}{c}{\textbf{VQAv2}~\cite{VQA_v2_2017}}                   &
      \multicolumn{1}{c}{\textbf{TextVQA}~\cite{TextVQA_2019}}                &
      \multicolumn{1}{c}{\textbf{ScienceQA}~\cite{ScienceQA_2022}}            &
      \multicolumn{1}{c}{\textbf{MM-Vet}~\cite{MM_Vet_2024}}
      \\
      \cmidrule(lr){3-4}
      \cmidrule(lr){5-7}
      \cmidrule(lr){8-8}
      \cmidrule(lr){9-9}
      \cmidrule(lr){10-10}
      \cmidrule(lr){11-11}
      \cmidrule(lr){12-12}
                                                                              &
                                                                              & Resp.\,$\downarrow$
                                                                              & Ment.\,$\downarrow$
                                                                              & CHAIR\,$\downarrow$
                                                                              & Hal.\,$\downarrow$
                                                                              & Cog.\,$\downarrow$
                                                                              & Question Acc.\,$\uparrow$
                                                                              & Acc.\,$\uparrow$
                                                                              & Acc.\,$\uparrow$
                                                                              & Image Acc.\,$\uparrow$
                                                                              & Overall\,$\uparrow$
      \\

      \midrule
      \multirow{11}{*}{LLaVA-v1.5-7B}                                         & baseline                        & 52.7         & 28.0         & 8.4          & 35.5          & 4.0          & 46.86          & \textbf{78.5} & \textbf{58.2} & 66.8          & 31.0          \\
                                                                              & VCD~\cite{VCD_2023}             & 51.3         & 25.9         & 9.1          & 39.8          & 4.2          & -              & 77.0          & 56.1          & 68.7          & 29.8          \\
                                                                              & OPERA~\cite{OPERA_2024}         & 45.3         & 22.9         & 6.5          & 28.5          & 3.1          & -              & 78.2          & \textbf{58.2} & 68.2          & 30.3          \\
                                                                              & DoLa~\cite{DoLa_2024}           & 44.0         & 25.1         & 6.2          & 27.7          & 2.9          & -              & 76.3          & 56.6          & 67.5          & 30.8          \\
                                                                              & EFUF~\cite{EFUF_2024}           & 39.3         & 22.6         & 5.8          & 28.2          & 3.1          & 47.03          & 78.1          & 57.2          & 66.4          & 31.2          \\
                                                                              & HA-DPO~\cite{HA_DPO_2023}       & 37.0         & 20.9         & 6.7          & 30.9          & 3.3          & \textbf{47.74} & 77.6          & \ul{56.7}     & \textbf{69.7} & 30.6          \\
                                                                              & POVID~\cite{POVID_2024}         & 33.4         & 16.6         & 5.3          & 28.7          & 3.0          & 46.59          & 77.2          & 56.6          & 68.8          & \ul{31.8}     \\
                                                                              & CLIP-DPO~\cite{CLIP_DPO_2024}   & -            & -            & 3.7          & 16.6          & 1.3          & -              & -             & 56.4          & 67.6          & -             \\
                                                                              & RLAIF-V~\cite{RLAIF_V_2024}     & 7.8          & 4.2          & \textbf{2.8} & \ul{15.7}     & \textbf{0.9} & 35.43          & 75.2          & 55.1          & 68.2          & 29.9          \\
                                                                              & TPO~\cite{Topic_level_SC_2024}  & 5.6          & 3.2          & 3.6          & 20.5          & 1.6          & 40.12          & 75.9          & 55.3          & 67.1          & 25.7          \\
                                                                              & \textbf{Ours}                   & \textbf{4.3} & \textbf{2.6} & \ul{2.9}     & \textbf{14.6} & \ul{1.2}     & \ul{47.56}     & \ul{78.4}     & \textbf{58.2} & \ul{69.2}     & \textbf{32.6} \\
      \midrule
      \multirow{5}{*}{LLaVA-v1.5-13B}                                         & baseline                        & 46.0         & 23.0         & 6.9          & 31.9          & 3.3          & \ul{46.43}     & \textbf{80.0} & \textbf{61.2} & 71.6          & \ul{36.0}     \\
                                                                              & VCD~\cite{VCD_2023}             & 43.7         & 21.6         & 7.8          & 36.2          & 3.7          & -              & 78.5          & 59.5          & \ul{72.0}     & 33.7          \\
                                                                              & vanilla-DPO~\cite{HSA_DPO_2024} & 6.7          & 3.6          & 2.8          & 15.5          & 1.6          & 46.41          & 79.2          & 60.4          & 71.8          & 35.0          \\
                                                                              & HSA-DPO~\cite{HSA_DPO_2024}     & \ul{5.3}     & \ul{3.2}     & \textbf{2.1} & \ul{13.4}     & \ul{1.2}     & 46.14          & 78.3          & 60.0          & 71.3          & 33.7          \\
                                                                              & \textbf{Ours}                   & \textbf{3.3} & \textbf{1.9} & \ul{2.7}     & \textbf{11.7} & \textbf{0.9} & \textbf{46.77} & \ul{79.9}     & \ul{61.0}     & \textbf{72.8} & \textbf{36.2} \\
      \bottomrule
    \end{tabular}
  }
  \vspace{-2.5mm}
  \caption{
    \textbf{Comparison of hallucination mitigation methods in MLLMs: effectiveness and general capabilities.}
    This evaluation highlights the best and second-best results in \textbf{bold} and \ul{underlined}, respectively. All comparisons are performed under identical model size constraints. ``Resp.'' and ``Ment.'' denote response-level and mention-level hallucination rates, while ``Hal.'' and ``Cog.'' represent the Hallucination Score and Cognitive Score, respectively. More evaluation details are provided in~\cref{sec:evaluation_details}.
  }
  \vspace{-2mm}
  \label{tab:main_result}
\end{table*}

\subsection{Context-aware Preference Learning}
\label{sec:sentence-level-training}
The preference data generated through the processes outlined in \cref{subsec:candidate-gen} and \cref{subsec:data-prepare} can be formally represented as $(\vec{x}, \vec{c}, \vec{y}_{w}^{+}, \vec{y}_l)$, where $\vec{x}$ is the input, including the image $\vec{v}$ and the prompt $\vec{q}$, $\vec{c}$ denotes the context, $\vec{y}^{+}_{w}$ is the context-coherent positive sample, and $\vec{y}_l$ is the negative sample.

The learning objective is to guide the model, conditioned on the input $\vec{x}$ and the context $\vec{c}$, to maximize the likelihood of generating the contextually coherent positive sample $\vec{y}^{+}_{w}$ while minimizing the likelihood of producing the negative sample $\vec{y}_l$.
To achieve this, we adapt the Direct Preference Optimization (DPO) loss by incorporating the context $\vec{c}$ as part of the input. We term this modified loss as context-aware DPO (C-DPO), which is formulated as follows:

\vspace{-10pt}
\begin{equation}
    \begin{aligned}
        \mathcal{L}_{\text{C-DPO}}(\vec{\theta}) =\, & -\mathbb{E}_{(\vec{x}',\vec{y}^{+}_{w},\vec{y}_l) \sim D} \Biggl[
        \log \sigma \Bigl(\beta \log \frac{\pi_{\vec{\theta}}(\vec{y}^{+}_{w}|\vec{x}')}{\pi_{\text{ref}}(\vec{y}^{+}_{w}|\vec{x}')}                                                            \\
                                                     & \qquad\qquad\qquad\qquad - \beta \log \frac{\pi_{\vec{\theta}}(\vec{y}_l|\vec{x}')}{\pi_{\text{ref}}(\vec{y}_l|\vec{x}')}\Bigr) \Biggr], \\
        \text{where } \vec{x}'                 =     & [\vec{x},\vec{c}] = [\vec{v},\vec{q},\vec{c}].
    \end{aligned}
    \label{eq:cdpo}
\end{equation}
\vspace{-10pt}

In C-DPO, the context $\vec{c}$ is excluded from the loss computation, and gradients are only derived from the discrimination between $\vec{y}^{+}_{w}$ and $\vec{y}_l$. This design ensures that the model focuses on learning the contextual coherence of the positive sample without being directly influenced by the context during gradient updates. Further discussions and comparisons between the proposed C-DPO and the standard DPO are provided in~\cref{subsec:training_objective}.
\section{Experiments}
\label{sec:experiments}

In this section, we conduct comprehensive experiments to evaluate the effectiveness of our SENTINEL in reducing hallucinations while improving the general abilities of the model. We first introduce the experimental setup in~\cref{subsec:exp_setup}, then present the main results in~\cref{subsec:main_results}, and finally conduct ablation studies in~\cref{subsec:ablation} to analyze our method's effectiveness. More results are in~\cref{sec:training_details,sec:evaluation_details}.

\subsection{Experimental Setup}
\label{subsec:exp_setup}

\mypara{Training.}
To ensure a fair comparison, we follow the settings of prior works~\cite{LLaVA_v1_5_2024,VCD_2023,HA_DPO_2023,HSA_DPO_2024,Topic_level_SC_2024,RLAIF_V_2024,HSA_DPO_2024,POVID_2024,CLIP_DPO_2024,EFUF_2024,OPERA_2024,EOS_2024,HALVA_2024,NoiseBoost_2024,Volcano_2024}, using \href{https://huggingface.co/collections/liuhaotian/llava-15-653aac15d994e992e2677a7e}{LLaVA-v1.5} as the reference model across all experiments. For data collection, we prompt the model with detailed image descriptions~\cite{RLHF_V_2024} to generate training data, with images sourced from the Visual Genome dataset~\cite{Visual_Genome_2017}. Model training is conducted using C-DPO (\cref{eq:cdpo}) in combination with LoRA~\cite{LoRA_2021}, and optimized with AdamW~\cite{AdamW_2019}. The 7B and 13B models are trained for one epoch on 8.6K and 7.0K samples, respectively, with learning rates of $2 \times 10^{-7}$ and $3 \times 10^{-7}$. Additional training details are provided in~\cref{sec:training_details}.

\mypara{Evaluation benchmarks.}
We evaluate the hallucination extent and general capabilities of our SENTINEL method across multiple benchmarks. For hallucination evaluation, we use widely adopted benchmarks, including Object HalBench~\cite{Obj_HalBench_2018}, AMBER~\cite{AMBER_2023}, and HallusionBench~\cite{HallusionBench_2023}. To assess general capabilities, we employ VQAv2~\cite{VQA_v2_2017}, TextVQA~\cite{TextVQA_2019}, ScienceQA~\cite{ScienceQA_2022}, and MM-Vet~\cite{MM_Vet_2024}. Further details of these benchmarks are provided in~\cref{subsec:evaluation_benchmarks}.

\mypara{Baselines.}
To show the effectiveness of our method, we compare SENTINEL with several state-of-the-art (SOTA) methods. Specifically, VCD~\cite{VCD_2023}, OPERA~\cite{OPERA_2024}, and DoLa~\cite{DoLa_2024} focus on enhanced decoding strategies, while HA-DPO~\cite{HA_DPO_2023}, POVID~\cite{POVID_2024}, CLIP-DPO~\cite{CLIP_DPO_2024}, RLAIF-V~\cite{RLAIF_V_2024}, and TPO~\cite{Topic_level_SC_2024} leverage preference training.
Additionally, Vanilla DPO applies the original DPO objective~\cref{eq:dpo} using training data from HSA-DPO, while EFUF~\cite{EFUF_2024} is an unlearning-based approach. Details are in~\cref{subsec:evaluation_counterparts}.

\subsection{Main Results}
\label{subsec:main_results}

\mypara{Comparison with recent SOTAs.}
As shown in~\cref{tab:main_result}, we compare our method with baseline methods across several benchmarks. The results demonstrate that SENTINEL significantly reduces the models' hallucination rate. Specifically, for the 7B model, our method achieves a \textbf{4.3} response-level (Resp.) and a \textbf{2.6} mention-level (Ment.) hallucination rate.
Compared to the previous SOTA method, TPO~\cite{Topic_level_SC_2024}, which achieves a 5.6 response-level and 3.2 mention-level hallucination rate, our proposed SENTINEL surpasses it by further reducing hallucinations by a total of 24\% on Object Halbench.
Furthermore, even on the 13B model, compared to the baseline, which achieves 6.9 CHAIR, 31.9 Hallucination score (Hal), and 4.0 Cognitive score (Cog), our proposed SENTINEL significantly improves performance, achieving 2.7 CHAIR, 11.7 Hal score, and 0.9 Cog score, respectively. These results demonstrate that our method is also effective on larger models.

\begin{figure}[t]
    \centering
    \includegraphics[width=0.4\textwidth]{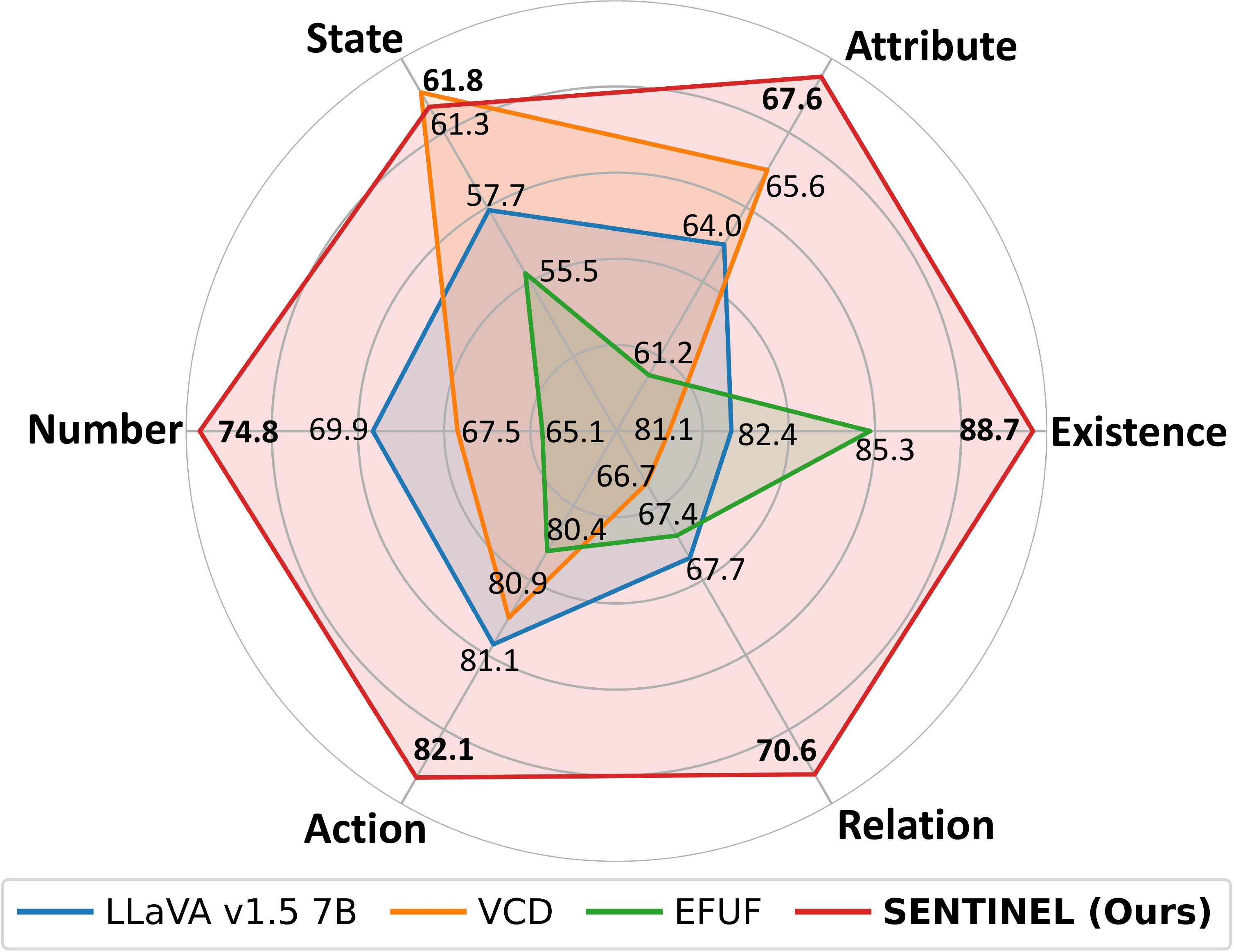}
    \vspace{-5pt}
    \caption{\textbf{Impact on different hallucination types.} Comparison between multiple methods shows that our method reduces hallucination in all six hallucination types.}
    \label{fig:radar}
    \vspace{-10pt}
\end{figure}

\mypara{Comprehensive hallucination mitigation.}
To further evaluate the effect of our method on various hallucination types, we conducted experiments on the discriminative part of the AMBER~\cite{AMBER_2023} benchmark and report the F1 scores for each hallucination type. As shown in~\cref{fig:radar}, LLaVA-v1.5 with SENTINEL outperforms the baseline across all six hallucination types, demonstrating the effectiveness of our method in mitigating various hallucination issues. Notably, for the \textit{Existence} hallucination type, our method improves the 7B model by \textbf{6.3} and the 13B model by \textbf{7.6} compared to the baseline. Detailed results are provided in~\cref{subsec:detailed_evaluation_results}.

\mypara{Improved general capabilities.}
As shown in~\cref{tab:main_result}, SENTINEL enhances the general capabilities of the model on multiple benchmarks.
Specifically, SENTINEL demonstrates stable performance on VQAv2 and TextVQA, whereas previous methods designed for hallucination mitigation suffer from significant performance degradation. Moreover, on ScienceQA and MM-Vet, our method not only surpasses previous hallucination mitigation methods but also outperforms the vanilla LLaVA-v1.5. These results show that SENTINEL not only significantly mitigates hallucinations but also improves general capability.

\begin{figure}[t]
    \centering
    \includegraphics[width=0.47\textwidth]{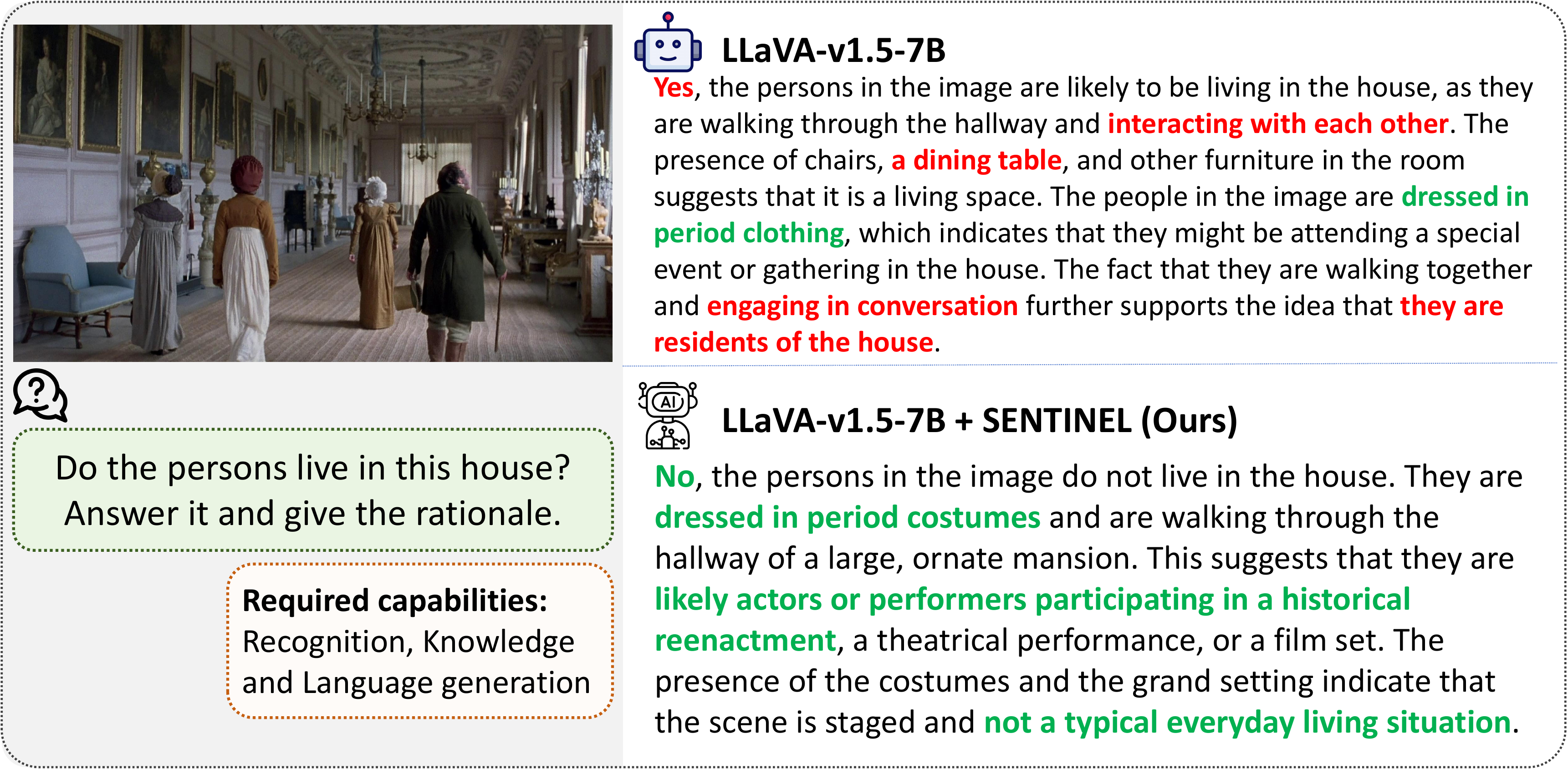}
    \vspace{-7pt}
    \caption{\textbf{Qualitative results of SENTINEL.} Our method can effectively eliminate hallucinations in MLLMs while enhancing the model's general capabilities.}
    \label{fig:single_case}
    \vspace{-7pt}
\end{figure}

\mypara{Qualitative results.}
\label{subsec:case_study}
To further demonstrate the effectiveness of our method, we conduct case studies. As shown in~\cref{fig:single_case}, the baseline model misinterprets the image content, leading to an incorrect conclusion. In contrast, our model effectively understands image content and provides a more detailed and precise description. This example highlights how our approach effectively reduces hallucinations while simultaneously enhancing the model's overall capability. We conduct more case studies in~\cref{sec:additional_case_studies}.

\subsection{Ablation Studies}
\label{subsec:ablation}

In this section, we conduct a series of ablation experiments to further analyze the effectiveness of SENTINEL. More discussions can be found in~\cref{subsec:ablation_study_details}.

\mypara{Effectiveness of data style consistency.}
To analyze the effect of preference data style, we train the model using rewritten data for comparison. Specifically, we instructed GPT-4~\cite{GPT4_2023} to rewrite ($\vec{y}^{+}_{w}$, $\vec{y}_{l}$) while ensuring coherence with the context $\vec{c}$. As shown in~\cref{tab:rewrite_sample_result}, the rewriting results show performance degradation in reducing hallucinations and general ability.
This highlights the advantage of our approach in preserving data style consistency. Furthermore, we conduct a detailed analysis in~\cref{subsec:ablation_study_details}, which shows that models trained on in-domain data converge to a lower preference optimization loss and achieve better differentiation between positive and negative samples, whereas training with rewritten data provides fewer improvements.

\begin{table}
    \centering
    \resizebox{\columnwidth}{!} {%
        \begin{tabular}{ll ccccc}
            \toprule
            \multirow{2}{*}{\vspace{-2mm}\textbf{Method}}          &
            \multicolumn{2}{c}{\textbf{Object HalBench}}           &
            \multicolumn{2}{c}{\textbf{AMBER}}                     &
            \multicolumn{1}{c}{\textbf{MM-Vet}}
            \\
            \cmidrule(lr){2-3}\cmidrule(lr){4-5}\cmidrule(lr){6-6} &
            {Resp. $\downarrow$}                                   &
            {Ment. $\downarrow$}                                   &
            {Acc $\uparrow$}                                       &
            {F1 $\uparrow$}                                        &
            {Overall $\uparrow$}
            \\
            \midrule
            LLaVA-v1.5-7B                                          & 52.7           & 27.9           & 71.5              & 74.1              & 31.1              \\
            Ours (8.6K $(\vec{y}^{+}_{w}, \vec{y}_{l})$)           & \textbf{4.3}   & \textbf{2.6}   & \textbf{76.1}     & \textbf{79.3}     & \textbf{32.6}     \\
            Ours (8.6K Rewrited $(\vec{y}^{+}_{w}, \vec{y}_{l})$)  & 4.8\upred{0.5} & 2.9\upred{0.3} & 75.0\downred{1.1} & 78.0\downred{1.3} & 31.3\downred{1.3} \\
            \bottomrule
        \end{tabular}

    }
    \vspace{-7pt}
    \caption{
        \textbf{Effects of rewritten samples.}
        Rewriting the preference training samples $(\vec{y}^{+}_{w}, \vec{y}_{l})$ results in performance reduction.
    }
    \vspace{-10pt}
    \label{tab:rewrite_sample_result}
\end{table}

\begin{figure}[t]
    \centering
    \includegraphics[width=0.47\textwidth]{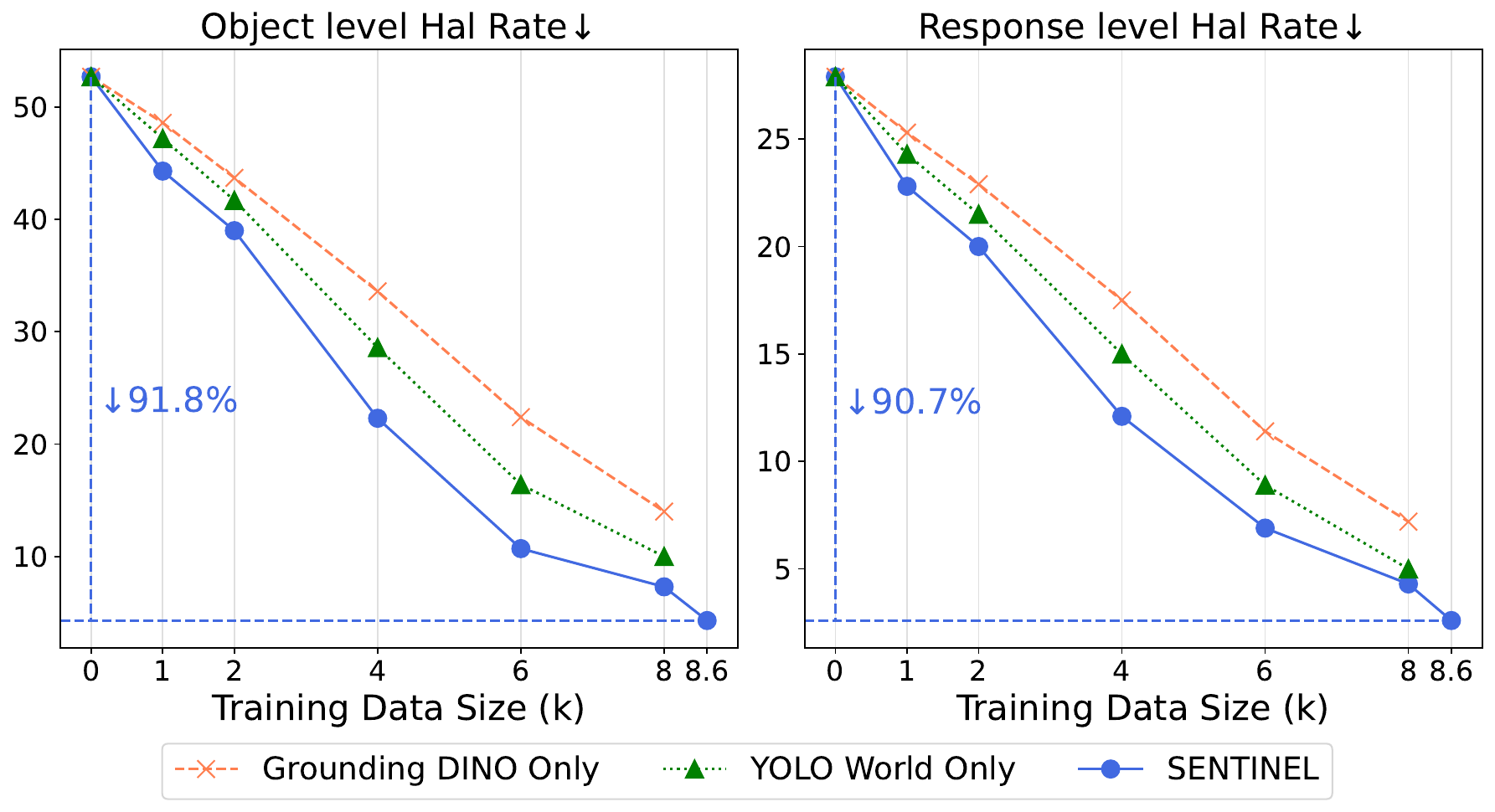}
    \vspace{-5pt}
    \caption{\textbf{Impact of training data quantity on hallucination rate in Object Halbench~\cite{Obj_HalBench_2018}.} The results show that SENTINEL demonstrates better efficiency, effectiveness, and scalability, while effectively reducing hallucination rates across varying data scales.}
    \label{fig:data_scale}
    \vspace{-10pt}
\end{figure}

\mypara{Effectiveness of cross-checking.}
To validate the effectiveness of cross-checking for object presence, we conduct experiments using only the Grounding DINO or YOLO World for detection. In this setting, if the model determines that an object is absent, it is directly classified as hallucinated. As shown in~\cref{fig:data_scale}, leveraging two object detectors for cross-validation significantly outperforms using a single model, effectively reducing the hallucination rate.

\mypara{Effect of different $\vec{y}_{w}$ types on model performance.}
As shown in~\cref{tab:different_yw_type}, we conduct a detailed study on the impact of different types and proportions of the positive data $\vec{y}_{w}$ on model performance. The results show that $\vec{y}^{+}_{w}$ samples, which contain richer contextual information, enhance the model's generalization ability while achieving similar hallucination reduction with less data.

\begin{table}
    \centering
    \resizebox{\columnwidth}{!} {%
        \begin{tabular}{l cccccc}
            \toprule
            \multirow{2}{*}{\vspace{-2mm}\textbf{Method}}                          &
            \multirow{2}{*}{\vspace{-2mm}\textbf{\makecell{Data \nextline Scale}}} &
            \multicolumn{2}{c}{\textbf{Object HalBench}}                           &
            \multicolumn{1}{c}{\textbf{TextVQA}}                                   &
            \multicolumn{1}{c}{\textbf{ScienceQA}}                                 &
            \multicolumn{1}{c}{\textbf{MM-Vet}}
            \\
            \cmidrule(lr){3-4}
            \cmidrule(lr){5-5}
            \cmidrule(lr){6-6}
            \cmidrule(lr){7-7}
                                                                                   &
                                                                                   &
            Resp. $\downarrow$                                                     &
            Ment. $\downarrow$                                                     &
            Acc                                                                    &
            I-Acc$\uparrow$                                                        &
            Overall $\uparrow$
            \\
            \midrule
            LLaVA-v1.5-7B                                                          & -                 & 52.7           & 27.9           & \textbf{58.2}     & 66.8              & 31.1              \\
            $\vec{y}^{+}_{w}$ 100\%                                                & \textbf{8.6K}     & \textbf{4.3}   & \textbf{2.6}   & \textbf{58.2}     & \textbf{69.2}     & \textbf{32.6}     \\
            $\vec{y}^{+}_{w}$ 50\% + $\vec{y}^{-}_{w}$ 50\%                        & 10.0K\upred{1.4K} & 4.8\upred{0.5} & 2.9\upred{0.3} & 58.1\downred{0.1} & 69.0\downred{0.2} & 32.0\downred{0.6} \\
            $\vec{y}^{-}_{w}$ 100\%                                                & 14.0K\upred{5.4K} & 4.6\upred{0.3} & 3.0\upred{0.4} & 58.1\downred{0.1} & 68.7\downred{0.5} & 31.6\downred{1.0} \\
            \bottomrule
        \end{tabular}

    }
    \vspace{-5pt}
    \caption{
        \textbf{Comparison between context-coherent samples $\vec{y}^{+}_{w}$ and context-agnostic samples $\vec{y}^{-}_{w}$.}
        This table reveals that incorporating context-coherent samples $\vec{y}^{+}_{w}$ yields better performance.
    }
    \vspace{-0.2cm}
    \label{tab:different_yw_type}
\end{table}

\mypara{Effect of non-hallucinated sentences as context $\vec{c}$.}
To analyze the impact of using non-hallucinated sentences as context $\vec{c}$, we evaluate three different settings for generating new context: selecting a hallucinated sentence, selecting a non-hallucinated sentence, or directly using a model-generated sentence from greedy decoding. As shown in~\cref{tab:different_context_type}, using a non-hallucinated sentence as context improves the model's ability to distinguish hallucinations and significantly reduces their occurrence in the output. This further demonstrates that intervening at the first instance of hallucination is critical for minimizing its recurrence.

\begin{table}
    \centering
    \resizebox{\columnwidth}{!} {%
        \begin{tabular}{l ccccc}
            \toprule
            \multirow{2}{*}{\vspace{-2mm}\textbf{Method}}                        &
            \multicolumn{2}{c}{\textbf{Object HalBench}\cite{Obj_HalBench_2018}} &
            \multicolumn{3}{c}{\textbf{AMBER}\cite{AMBER_2023}}
            \\
            \cmidrule(lr){2-3}
            \cmidrule(lr){4-6}                                                   &
            Resp. $\downarrow$                                                   &
            Ment. $\downarrow$                                                   &
            CHAIR $\downarrow$                                                   &
            Hal $\downarrow$                                                     &
            Cog $\downarrow$
            \\
            \midrule
            LLaVA-v1.5-7B                                                        & 52.7         & 27.9         & 8.4          & 35.5          & 4.0          \\
            Non-hallucinated context                                             & \textbf{4.3} & \textbf{2.6} & \textbf{2.9} & \textbf{14.6} & \textbf{1.2} \\
            Natural context                                                      & \ul{8.6}     & \ul{4.7}     & \ul{3.3}     & \ul{15.6}     & \ul{1.5}     \\
            Hallucinated context                                                 & 14.3         & 7.1          & 3.9          & 18.6          & 1.8          \\
            \bottomrule
        \end{tabular}

    }
    \vspace{-5pt}
    \caption{
        \textbf{Comparison between different new context formation strategies during the iterative contextual bootstrapping pipeline.}
        Appending non-hallucinated sample $\vec{y}^{+}_{w}$ to the existing context $\vec{c}_{i}$ yields superior performance compared to incorporating hallucinated samples $\vec{y}_{l}$ or greedy decoding contexts, highlighting the effectiveness of our proposed approach.
    }
    \vspace{-10pt}
    \label{tab:different_context_type}
\end{table}

\mypara{Effect of data scale.}
To analyze the impact of the training data scale on our method, we train the model using different dataset sizes (1k/2k/4k/6k/8k) and evaluate its performance on Object Halbench. As shown in~\cref{fig:data_scale}, our method further mitigates model hallucinations as data scale up. This demonstrates the potential and scalability of SENTINEL. Furthermore, since our method does not rely on ultra-large proprietary models or human annotators for dataset construction, it can efficiently collect more training data.

\mypara{Integrating with existing preference learning methods.}
To further demonstrate SENTINEL's generalization, we explore integrating with previous hallucination mitigation approaches. As shown in~\cref{tab:combine_method}, incorporating a subset of our data into the GPT-generated dataset collected by HA-DPO~\cite{HA_DPO_2023} effectively mitigates hallucinations while significantly enhancing the model's generalization. This highlights SENTINEL's complementarity with other preference learning methods and its potential for broader applicability.

\begin{table}
    \centering
    \resizebox{\columnwidth}{!} {%
        \begin{tabular}{l ccccccc}
            \toprule
            \multirow{2}{*}{\vspace{-2mm}\textbf{Method}}                         &
            \multicolumn{2}{c}{\textbf{Object HalBench}}\cite{Obj_HalBench_2018}  &
            \multicolumn{2}{c}{\textbf{AMBER}}\cite{AMBER_2023}                   &
            \multicolumn{1}{c}{\textbf{HallusionBench}}\cite{HallusionBench_2023} &
            \multicolumn{1}{c}{\textbf{TextVQA}}\cite{TextVQA_2019}               &
            \multicolumn{1}{c}{\textbf{MM-Vet}}\cite{MM_Vet_2024}
            \\
            \cmidrule(lr){2-3}
            \cmidrule(lr){4-5}
            \cmidrule(lr){6-6}
            \cmidrule(lr){7-7}
            \cmidrule(lr){8-8}

                                                                                  & {Resp. $\downarrow$}            & {Ment. $\downarrow$}           & Acc$\uparrow$              & F1$\uparrow$               & Question Acc$\uparrow$       & Acc$\uparrow$          & {Overall $\uparrow$}       \\
            \midrule
            LLaVA-v1.5-7B                                                         & 52.7                            & 28.0                           & 71.5                       & 74.1                       & 46.86                        & 58.2                   & 31.0                       \\
            HA-DPO~\cite{HA_DPO_2023}                                             & 37.0\downgreen{29.8\%}          & 20.9\downgreen{25.4\%}         & 74.2\upgreen{2.7}          & 78.0\upgreen{3.9}          & 47.74\upgreen{0.88}          & 56.7\downred{1.5}      & 30.6\downred{0.4}          \\
            HA-DPO + Ours (6K)                                                    & \textbf{8.0} \downgreen{78.4\%} & \textbf{4.6}\downgreen{78.0\%} & \textbf{76.6}\upgreen{2.4} & \textbf{84.2}\upgreen{6.2} & \textbf{48.72}\upgreen{0.98} & \ul{57.1}\upgreen{0.4} & \textbf{33.5}\upgreen{2.9} \\
            \bottomrule
        \end{tabular}
    }
    \vspace{-5pt}
    \caption{
        \textbf{Effectiveness of combining the proposed SENTINEL with HA-DPO.}
        Only a subset of our training data is needed to reduce hallucinations while enhancing generalization effectively.
    }
    \vspace{-10pt}
    \label{tab:combine_method}
\end{table}
\section{Concluding Remarks}
\label{sec:conclusion}

\mypara{Summary.}
In this work, we address the critical challenge of hallucinations in multimodal large language models (MLLMs). While prior methods have shown promise, they often introduce significant computational overhead, rely on costly resources, or create distributional discrepancies. To tackle these issues, we propose SENTINEL, a framework that intervenes early at the onset of hallucinations by leveraging in-domain preference learning. SENTINEL employs an in-domain candidate bootstrapping strategy, context-aware preference data generation, and a context-aware DPO (C-DPO) loss to effectively curb the propagation of hallucinations while preserving the model's intrinsic distribution. Experimental results across multiple benchmarks demonstrate the superiority of SENTINEL, establishing it as a scalable, efficient, and model-agnostic solution for enhancing the reliability of MLLMs.

\mypara{Limitation.}
Currently, as SENTINEL lacks the capability to incorporate spatiotemporal information, it might not be able to effectively address the hallucination issues that require long-term reasoning in video MLLMs. This limitation highlights the need for further research in this area.

\clearpage
\mypara{Acknowledgments.}
This work is supported by the Shenzhen Science and Technology Innovation Program (JCYJ20240813105901003, KJZD20240903102901003), Guangdong Basic and Applied Basic Research Foundation (2025A1515011546), and National Key R\&D Program of China (2024YFE0215300).

{
    \small
    \bibliographystyle{ieeenat_fullname}
    \bibliography{main}

@article{Qwen_VL_2023,
  title   = {{Qwen-VL}: A Versatile Vision-Language Model for Understanding, Localization},
  author  = {Bai, Jinze and Bai, Shuai and Yang, Shusheng and Wang, Shijie and Tan, Sinan and Wang, Peng and Lin, Junyang and Zhou, Chang and Zhou, Jingren},
  journal = {arXiv preprint arXiv:2308.12966},
  year    = {2023}
}

@article{Qwen2_5_VL_2024,
  title   = {{Qwen2.5-VL} Technical Report},
  author  = {Bai, Shuai and Chen, Keqin and Liu, Xuejing and Wang, Jialin and Ge, Wenbin and Song, Sibo and Dang, Kai and Wang, Peng and Wang, Shijie and Tang, Jun and others},
  journal = {arXiv preprint arXiv:2502.13923},
  year    = {2025}
}

@article{Qwen2_VL_2024,
  title   = {{Qwen2-VL}: Enhancing Vision-Language Model's Perception of the World at Any Resolution},
  author  = {Wang, Peng and Bai, Shuai and Tan, Sinan and Wang, Shijie and Fan, Zhihao and Bai, Jinze and Chen, Keqin and Liu, Xuejing and Wang, Jialin and Ge, Wenbin and others},
  journal = {arXiv preprint arXiv:2409.12191},
  year    = {2024}
}

@article{zhong2024lyra,
  title   = {Lyra: An Efficient and Speech-Centric Framework for Omni-Cognition},
  author  = {Zhong, Zhisheng and Wang, Chengyao and Liu, Yuqi and Yang, Senqiao and Tang, Longxiang and Zhang, Yuechen and Li, Jingyao and Qu, Tianyuan and Li, Yanwei and Chen, Yukang and others},
  journal = {arXiv preprint arXiv:2412.09501},
  year    = {2024}
}

@article{qu2025does,
  title   = {Does Your Vision-Language Model Get Lost in the Long Video Sampling Dilemma?},
  author  = {Qu, Tianyuan and Tang, Longxiang and Peng, Bohao and Yang, Senqiao and Yu, Bei and Jia, Jiaya},
  journal = {arXiv preprint arXiv:2503.12496},
  year    = {2025}
}

@article{li2023mgm,
  title   = {{Mini-Gemini}: Mining the Potential of Multi-modality Vision Language Models},
  author  = {Li, Yanwei and Zhang, Yuechen and Wang, Chengyao and Zhong, Zhisheng and Chen, Yixin and Chu, Ruihang and Liu, Shaoteng and Jia, Jiaya},
  journal = {arXiv preprint arXiv:2403.18814},
  year    = {2023}
}

@inproceedings{yang2024unified,
  title     = {Unified Language-driven Zero-shot Domain Adaptation},
  author    = {Yang, Senqiao and Tian, Zhuotao and Jiang, Li and Jia, Jiaya},
  booktitle = {Proceedings of the IEEE Conference on Computer Vision and Pattern Recognition},
  year      = {2024}
}

@article{LLaVA_v1_2023,
  title   = {Visual instruction tuning},
  author  = {Liu, Haotian and Li, Chunyuan and Wu, Qingyang and Lee, Yong Jae},
  journal = {Advances in neural information processing systems},
  year    = {2023}
}

@article{yang2024visionzip,
  title   = {{VisionZip}: Longer is Better but Not Necessary in Vision Language Models},
  author  = {Yang, Senqiao and Chen, Yukang and Tian, Zhuotao and Wang, Chengyao and Li, Jingyao and Yu, Bei and Jia, Jiaya},
  journal = {arXiv preprint arXiv:2412.04467},
  year    = {2024}
}

@inproceedings{lai2024lisa,
  title     = {{LISA}: Reasoning Segmentation via Large Language Model},
  author    = {Lai, Xin and Tian, Zhuotao and Chen, Yukang and Li, Yanwei and Yuan, Yuhui and Liu, Shu and Jia, Jiaya},
  booktitle = {Proceedings of the IEEE Conference on Computer Vision and Pattern Recognition},
  year      = {2024}
}

@article{yang2023lidar,
  title   = {{LiDAR-LLM}: Exploring the Potential of Large Language Models for 3D {LiDAR} Understanding},
  author  = {Yang, Senqiao and Liu, Jiaming and Zhang, Ray and Pan, Mingjie and Guo, Zoey and Li, Xiaoqi and Chen, Zehui and Gao, Peng and Guo, Yandong and Zhang, Shanghang},
  journal = {arXiv preprint arXiv:2312.14074},
  year    = {2023}
}

@inproceedings{shao2024explore,
  title     = {Explore the Potential of {CLIP} for Training-Free Open Vocabulary Semantic Segmentation},
  author    = {Shao, Tong and Tian, Zhuotao and Zhao, Hang and Su, Jingyong},
  booktitle = {Proceedings of the European Conference on Computer Vision},
  year      = {2024}
}

@inproceedings{wang2025declip,
  title     = {{DeCLIP}: Decoupled Learning for Open-Vocabulary Dense Perception},
  author    = {Wang, Junjie and Chen, Bin and Li, Yulin and Kang, Bin and Chen, Yichi and Tian, Zhuotao},
  booktitle = {Proceedings of the Computer Vision and Pattern Recognition Conference},
  year      = {2025}
}

@inproceedings{tian2019learning,
  title     = {Learning shape-aware embedding for scene text detection},
  author    = {Tian, Zhuotao and Shu, Michelle and Lyu, Pengyuan and Li, Ruiyu and Zhou, Chao and Shen, Xiaoyong and Jia, Jiaya},
  booktitle = {Proceedings of the IEEE Conference on Computer Vision and Pattern Recognition},
  pages     = {4234--4243},
  year      = {2019}
}

@article{liu2024typicalness,
  title   = {Typicalness-Aware Learning for Failure Detection},
  author  = {Liu, Yijun and Cui, Jiequan and Tian, Zhuotao and Yang, Senqiao and He, Qingdong and Wang, Xiaoling and Su, Jingyong},
  journal = {arXiv preprint arXiv:2411.01981},
  year    = {2024}
}

@article{yang2023improved,
  title   = {An Improved Baseline for Reasoning Segmentation with Large Language Model},
  author  = {Yang, Senqiao and Qu, Tianyuan and Lai, Xin and Tian, Zhuotao and Peng, Bohao and Liu, Shu and Jia, Jiaya},
  journal = {arXiv preprint arXiv:2312.17240},
  year    = {2023}
}

@inproceedings{LLaVA_v1_5_2024,
  title     = {Improved baselines with visual instruction tuning},
  author    = {Liu, Haotian and Li, Chunyuan and Li, Yuheng and Lee, Yong Jae},
  booktitle = {Proceedings of the IEEE Conference on Computer Vision and Pattern Recognition},
  year      = {2024}
}

@misc{LLaVA_NeXT_2024,
  title  = {{LLaVA-NeXT}: Improved reasoning, {OCR}, and world knowledge},
  author = {Liu, Haotian and Li, Chunyuan and Li, Yuheng and Li, Bo and Zhang, Yuanhan and Shen, Sheng and Lee, Yong Jae},
  year   = {2024},
  url    = {https://llava-vl.github.io/blog/2024-01-30-llava-next/}
}

@article{InstructBLIP_2023,
  title   = {{InstructBLIP}: Towards General-purpose Vision-Language Models with Instruction Tuning},
  author  = {Dai, Wenliang and Li, Junnan and Li, Dongxu and Tiong, Anthony and Zhao, Junqi and Wang, Weisheng and Li, Boyang and Fung, Pascale N and Hoi, Steven},
  journal = {Advances in Neural Information Processing Systems},
  year    = {2023}
}

@article{GPT4_2023,
  title   = {{GPT-4} Technical Report},
  author  = {Achiam, Josh and Adler, Steven and Agarwal, Sandhini and Ahmad, Lama and Akkaya, Ilge and Aleman, Florencia Leoni and Almeida, Diogo and Altenschmidt, Janko and Altman, Sam and Anadkat, Shyamal and others},
  journal = {arXiv preprint arXiv:2303.08774},
  year    = {2023}
}

@misc{OpenAI_GPT4V_2023,
  title  = {{GPT-4V(ision)} System Card},
  author = {OpenAI},
  year   = {2023}
}

@article{HallucinationSurvey_2023,
  title   = {A survey of hallucination in large foundation models},
  author  = {Rawte, Vipula and Sheth, Amit and Das, Amitava},
  journal = {arXiv preprint arXiv:2309.05922},
  year    = {2023}
}

@article{HallucinationSurvey_2024,
  title   = {Hallucination of multimodal large language models: A survey},
  author  = {Bai, Zechen and Wang, Pichao and Xiao, Tianjun and He, Tong and Han, Zongbo and Zhang, Zheng and Shou, Mike Zheng},
  journal = {arXiv preprint arXiv:2404.18930},
  year    = {2024}
}

@article{HallucinationSurvey_202402,
  title   = {A survey on hallucination in large vision-language models},
  author  = {Liu, Hanchao and Xue, Wenyuan and Chen, Yifei and Chen, Dapeng and Zhao, Xiutian and Wang, Ke and Hou, Liping and Li, Rongjun and Peng, Wei},
  journal = {arXiv preprint arXiv:2402.00253},
  year    = {2024}
}

@article{Skipn_2024,
  title   = {Skip$\backslash$n: A Simple Method to Reduce Hallucination in Large Vision-Language Models},
  author  = {Han, Zongbo and Bai, Zechen and Mei, Haiyang and Xu, Qianli and Zhang, Changqing and Shou, Mike Zheng},
  journal = {arXiv preprint arXiv:2402.01345},
  year    = {2024}
}

@article{Step_DPO_2024,
  title   = {{Step-DPO}: Step-wise Preference Optimization for Long-chain Reasoning of {LLMs}},
  author  = {Lai, Xin and Tian, Zhuotao and Chen, Yukang and Yang, Senqiao and Peng, Xiangru and Jia, Jiaya},
  journal = {arXiv preprint arXiv:2406.18629},
  year    = {2024}
}

@article{HA_DPO_2023,
  title   = {Beyond Hallucinations: Enhancing {LVLMs} through Hallucination-Aware Direct Preference Optimization},
  author  = {Zhao, Zhiyuan and Wang, Bin and Ouyang, Linke and Dong, Xiaoyi and Wang, Jiaqi and He, Conghui},
  journal = {arXiv preprint arXiv:2311.16839},
  year    = {2023}
}

@article{HSA_DPO_2024,
  title   = {Detecting and Mitigating Hallucination in Large Vision Language Models via Fine-Grained {AI} Feedback},
  author  = {Xiao, Wenyi and Huang, Ziwei and Gan, Leilei and He, Wanggui and Li, Haoyuan and Yu, Zhelun and Shu, Fangxun and Jiang, Hao and Zhu, Linchao},
  journal = {arXiv preprint arXiv:2404.14233},
  year    = {2024}
}

@article{POVID_2024,
  title   = {Aligning Modalities in Vision Large Language Models via Preference Fine-tuning},
  author  = {Zhou, Yiyang and Cui, Chenhang and Rafailov, Rafael and Finn, Chelsea and Yao, Huaxiu},
  journal = {arXiv preprint arXiv:2402.11411},
  year    = {2024}
}

@article{RLAIF_V_2024,
  title   = {{RLAIF-V}: Open-Source {AI} Feedback Leads to Super {GPT-4V} Trustworthiness},
  author  = {Yu, Tianyu and Zhang, Haoye and Yao, Yuan and Dang, Yunkai and Chen, Da and Lu, Xiaoman and Cui, Ganqu and He, Taiwen and Liu, Zhiyuan and Chua, Tat-Seng and others},
  journal = {arXiv preprint arXiv:2405.17220},
  year    = {2024}
}

@article{Mask_DPO_2025,
  title   = {{Mask-DPO}: Generalizable Fine-grained Factuality Alignment of {LLMs}},
  author  = {Gu, Yuzhe and Zhang, Wenwei and Lyu, Chengqi and Lin, Dahua and Chen, Kai},
  journal = {arXiv preprint arXiv:2503.02846},
  year    = {2025}
}

@inproceedings{RLHF_V_2024,
  title     = {{RLHF-V}: Towards Trustworthy {MLLMs} via Behavior Alignment from Fine-grained Correctional Human Feedback},
  author    = {Yu, Tianyu and Yao, Yuan and Zhang, Haoye and He, Taiwen and Han, Yifeng and Cui, Ganqu and Hu, Jinyi and Liu, Zhiyuan and Zheng, Hai-Tao and Sun, Maosong and others},
  booktitle = {Proceedings of the IEEE Conference on Computer Vision and Pattern Recognition},
  year      = {2024}
}

@inproceedings{M_HalDetect_2024,
  title     = {Detecting and preventing hallucinations in large vision language models},
  author    = {Gunjal, Anisha and Yin, Jihan and Bas, Erhan},
  booktitle = {Proceedings of the AAAI Conference on Artificial Intelligence},
  year      = {2024}
}

@article{LURE_2024,
  title   = {Analyzing and mitigating object hallucination in large vision-language models},
  author  = {Zhou, Yiyang and Cui, Chenhang and Yoon, Jaehong and Zhang, Linjun and Deng, Zhun and Finn, Chelsea and Bansal, Mohit and Yao, Huaxiu},
  journal = {arXiv preprint arXiv:2310.00754},
  year    = {2023}
}

@article{FaithScore_2024,
  title   = {{FaithScore}: Fine-grained Evaluations of Hallucinations in Large Vision-Language Models},
  author  = {Jing, Liqiang and Li, Ruosen and Chen, Yunmo and Du, Xinya},
  journal = {arXiv preprint arXiv:2311.01477},
  year    = {2023}
}

@inproceedings{Grounding_DINO_2024,
  title     = {{Grounding DINO}: Marrying {DINO} with Grounded Pre-Training for Open-Set Object Detection},
  author    = {Liu, Shilong and Zeng, Zhaoyang and Ren, Tianhe and Li, Feng and Zhang, Hao and Yang, Jie and Jiang, Qing and Li, Chunyuan and Yang, Jianwei and Su, Hang and others},
  booktitle = {Proceedings of the European Conference on Computer Vision},
  year      = {2024}
}

@inproceedings{YOLO_World_2024,
  title     = {{YOLO-World}: Real-Time Open-Vocabulary Object Detection},
  author    = {Cheng, Tianheng and Song, Lin and Ge, Yixiao and Liu, Wenyu and Wang, Xinggang and Shan, Ying},
  booktitle = {Proceedings of the IEEE Conference on Computer Vision and Pattern Recognition},
  year      = {2024}
}

@article{FACTUAL_2023,
  title   = {{FACTUAL}: A Benchmark for Faithful and Consistent Textual Scene Graph Parsing},
  author  = {Li, Zhuang and Chai, Yuyang and Zhuo, Terry Yue and Qu, Lizhen and Haffari, Gholamreza and Li, Fei and Ji, Donghong and Tran, Quan Hung},
  journal = {arXiv preprint arXiv:2305.17497},
  year    = {2023}
}

@article{Obj_HalBench_2018,
  title   = {Object hallucination in image captioning},
  author  = {Rohrbach, Anna and Hendricks, Lisa Anne and Burns, Kaylee and Darrell, Trevor and Saenko, Kate},
  journal = {arXiv preprint arXiv:1809.02156},
  year    = {2018}
}

@article{AMBER_2023,
  title   = {{AMBER}: An {LLM}-free Multi-dimensional Benchmark for {MLLMs} Hallucination Evaluation},
  author  = {Wang, Junyang and Wang, Yuhang and Xu, Guohai and Zhang, Jing and Gu, Yukai and Jia, Haitao and Wang, Jiaqi and Xu, Haiyang and Yan, Ming and Zhang, Ji and others},
  journal = {arXiv preprint arXiv:2311.07397},
  year    = {2023}
}

@inproceedings{HallusionBench_2023,
  title     = {{HallusionBench}: An Advanced Diagnostic Suite for Entangled Language Hallucination and Visual Illusion in Large Vision-Language Models},
  author    = {Guan, Tianrui and Liu, Fuxiao and Wu, Xiyang and Xian, Ruiqi and Li, Zongxia and Liu, Xiaoyu and Wang, Xijun and Chen, Lichang and Huang, Furong and Yacoob, Yaser and others},
  booktitle = {Proceedings of the IEEE Conference on Computer Vision and Pattern Recognition},
  year      = {2024}
}

@inproceedings{TextVQA_2019,
  title     = {Towards {VQA} Models That Can Read},
  author    = {Singh, Amanpreet and Natarajan, Vivek and Shah, Meet and Jiang, Yu and Chen, Xinlei and Batra, Dhruv and Parikh, Devi and Rohrbach, Marcus},
  booktitle = {Proceedings of the IEEE Conference on Computer Vision and Pattern Recognition},
  year      = {2019}
}

@article{ScienceQA_2022,
  title   = {Learn to explain: Multimodal reasoning via thought chains for science question answering},
  author  = {Lu, Pan and Mishra, Swaroop and Xia, Tanglin and Qiu, Liang and Chang, Kai-Wei and Zhu, Song-Chun and Tafjord, Oyvind and Clark, Peter and Kalyan, Ashwin},
  journal = {Advances in Neural Information Processing Systems},
  year    = {2022}
}

@article{MM_Vet_2024,
  title   = {{MM-Vet}: Evaluating Large Multimodal Models for Integrated Capabilities},
  author  = {Yu, Weihao and Yang, Zhengyuan and Li, Linjie and Wang, Jianfeng and Lin, Kevin and Liu, Zicheng and Wang, Xinchao and Wang, Lijuan},
  journal = {arXiv preprint arXiv:2308.02490},
  year    = {2023}
}

@article{DPO_2023,
  title   = {Direct preference optimization: Your language model is secretly a reward model},
  author  = {Rafailov, Rafael and Sharma, Archit and Mitchell, Eric and Manning, Christopher D and Ermon, Stefano and Finn, Chelsea},
  journal = {Advances in Neural Information Processing Systems},
  year    = {2023}
}

@article{PPO_2017,
  title   = {Proximal policy optimization algorithms},
  author  = {Schulman, John and Wolski, Filip and Dhariwal, Prafulla and Radford, Alec and Klimov, Oleg},
  journal = {arXiv preprint arXiv:1707.06347},
  year    = {2017}
}

@article{FGAIF_2024,
  title   = {{FGAIF}: Aligning Large Vision-Language Models with Fine-grained {AI} Feedback},
  author  = {Jing, Liqiang and Du, Xinya},
  journal = {arXiv preprint arXiv:2404.05046},
  year    = {2024}
}

@inproceedings{AMP_2024,
  title     = {Automated Multi-level Preference for {MLLMs}},
  author    = {Zhang, Mengxi and Wu, Wenhao and Lu, Yu and Song, Yuxin and Rong, Kang and Yao, Huanjin and Zhao, Jianbo and Liu, Fanglong and Feng, Haocheng and Wang, Jingdong and others},
  booktitle = {The Thirty-eighth Annual Conference on Neural Information Processing Systems},
  year      = {2024}
}

@article{Topic_level_SC_2024,
  title   = {A Topic-level Self-Correctional Approach to Mitigate Hallucinations in {MLLMs}},
  author  = {He, Lehan and Chen, Zeren and Shi, Zhelun and Yu, Tianyu and Shao, Jing and Sheng, Lu},
  journal = {arXiv preprint arXiv:2411.17265},
  year    = {2024}
}

@article{DoLa_2024,
  title   = {{DoLa}: Decoding by Contrasting Layers Improves Factuality in Large Language Models},
  author  = {Chuang, Yung-Sung and Xie, Yujia and Luo, Hongyin and Kim, Yoon and Glass, James and He, Pengcheng},
  journal = {arXiv preprint arXiv:2309.03883},
  year    = {2023}
}

@inproceedings{OPERA_2024,
  title     = {{OPERA}: Alleviating Hallucination in Multi-Modal Large Language Models via Over-Trust Penalty and Retrospection-Allocation},
  author    = {Huang, Qidong and Dong, Xiaoyi and Zhang, Pan and Wang, Bin and He, Conghui and Wang, Jiaqi and Lin, Dahua and Zhang, Weiming and Yu, Nenghai},
  booktitle = {Proceedings of the IEEE Conference on Computer Vision and Pattern Recognition},
  year      = {2024}
}

@article{HALC_2024,
  title   = {{HALC}: Object Hallucination Reduction via Adaptive Focal-Contrast Decoding},
  author  = {Chen, Zhaorun and Zhao, Zhuokai and Luo, Hongyin and Yao, Huaxiu and Li, Bo and Zhou, Jiawei},
  journal = {arXiv preprint arXiv:2403.00425},
  year    = {2024}
}

@inproceedings{VCD_2023,
  title     = {Mitigating object hallucinations in large vision-language models through visual contrastive decoding},
  author    = {Leng, Sicong and Zhang, Hang and Chen, Guanzheng and Li, Xin and Lu, Shijian and Miao, Chunyan and Bing, Lidong},
  booktitle = {Proceedings of the IEEE Conference on Computer Vision and Pattern Recognition},
  year      = {2024}
}

@article{Visual_Genome_2017,
  title   = {{Visual Genome}: Connecting Language and Vision Using Crowdsourced Dense Image Annotations},
  author  = {Krishna, Ranjay and Zhu, Yuke and Groth, Oliver and Johnson, Justin and Hata, Kenji and Kravitz, Joshua and Chen, Stephanie and Kalantidis, Yannis and Li, Li-Jia and Shamma, David A and others},
  journal = {International journal of computer vision},
  year    = {2017}
}

@inproceedings{LoRA_2021,
  title     = {{LoRA}: Low-Rank Adaptation of Large Language Models},
  author    = {Hu, Edward J and Shen, Yelong and Wallis, Phillip and Allen-Zhu, Zeyuan and Li, Yuanzhi and Wang, Shean and Wang, Lu and Chen, Weizhu and others},
  booktitle = {International Conference on Learning Representations},
  year      = {2022}
}

@article{AdamW_2019,
  title   = {Decoupled Weight Decay Regularization},
  author  = {Loshchilov, Ilya and Hutter, Frank},
  journal = {arXiv preprint arXiv:1711.05101},
  year    = {2017}
}

@article{Visual_Perturbation_2022,
  title   = {Visual Perturbation-aware Collaborative Learning for Overcoming the Language Prior Problem},
  author  = {Han, Yudong and Nie, Liqiang and Yin, Jianhua and Wu, Jianlong and Yan, Yan},
  journal = {arXiv preprint arXiv:2207.11850},
  year    = {2022}
}

@inproceedings{Counterfactual_VQA_2021,
  title     = {{Counterfactual VQA}: A Cause-Effect Look at Language Bias},
  author    = {Niu, Yulei and Tang, Kaihua and Zhang, Hanwang and Lu, Zhiwu and Hua, Xian-Sheng and Wen, Ji-Rong},
  booktitle = {Proceedings of the IEEE Conference on Computer Vision and Pattern Recognition},
  year      = {2021}
}

@inproceedings{Distinguishing_VQA_2022,
  title     = {Overcoming Language Priors in Visual Question Answering via Distinguishing Superficially Similar Instances},
  author    = {Wu, Yike and Zhao, Yu and Zhao, Shiwan and Zhang, Ying and Yuan, Xiaojie and Zhao, Guoqing and Jiang, Ning},
  booktitle = {Proceedings of the 29th International Conference on Computational Linguistics},
  year      = {2022}
}

@article{Advancing_Medical_Imaging_2023,
  title   = {Advancing Medical Imaging with Language Models: A Journey from {N-grams} to {ChatGPT}},
  author  = {Hu, Mingzhe and Pan, Shaoyan and Li, Yuheng and Yang, Xiaofeng},
  journal = {arXiv preprint arXiv:2304.04920},
  year    = {2023}
}

@inproceedings{Driving_with_LLMs_2024,
  title     = {Driving with {LLMs}: Fusing Object-Level Vector Modality for Explainable Autonomous Driving},
  author    = {Chen, Long and Sinavski, Oleg and H{\"u}nermann, Jan and Karnsund, Alice and Willmott, Andrew James and Birch, Danny and Maund, Daniel and Shotton, Jamie},
  booktitle = {IEEE International Conference on Robotics and Automation (ICRA)},
  year      = {2024}
}

@article{Embodied_Task_Planning_2023,
  title   = {Embodied Task Planning with Large Language Models},
  author  = {Wu, Zhenyu and Wang, Ziwei and Xu, Xiuwei and Lu, Jiwen and Yan, Haibin},
  journal = {arXiv preprint arXiv:2307.01848},
  year    = {2023}
}

@inproceedings{Fine_tuning_Factuality_2023,
  title     = {Fine-tuning language models for factuality},
  author    = {Tian, Katherine and Mitchell, Eric and Yao, Huaxiu and Manning, Christopher D and Finn, Chelsea},
  booktitle = {The Twelfth International Conference on Learning Representations},
  year      = {2023}
}

@article{FLAME_2024,
  title   = {{FLAME}: Factuality-Aware Alignment for Large Language Models},
  author  = {Lin, Sheng-Chieh and Gao, Luyu and Oguz, Barlas and Xiong, Wenhan and Lin, Jimmy and Yih, Scott and Chen, Xilun},
  journal = {Advances in Neural Information Processing Systems},
  year    = {2024}
}

@article{LRV_Instruction_2024,
  title   = {Mitigating Hallucination in Large Multi-Modal Models via Robust Instruction Tuning},
  author  = {Liu, Fuxiao and Lin, Kevin and Li, Linjie and Wang, Jianfeng and Yacoob, Yaser and Wang, Lijuan},
  journal = {arXiv preprint arXiv:2306.14565},
  year    = {2023}
}

@article{Woodpecker_2023,
  title   = {Woodpecker: Hallucination Correction for Multimodal Large Language Models},
  author  = {Yin, Shukang and Fu, Chaoyou and Zhao, Sirui and Xu, Tong and Wang, Hao and Sui, Dianbo and Shen, Yunhang and Li, Ke and Sun, Xing and Chen, Enhong},
  journal = {Science China Information Sciences},
  year    = {2024}
}

@article{MiniGPT_4_2024,
  title   = {{MiniGPT-4}: Enhancing Vision-Language Understanding with Advanced Large Language Models},
  author  = {Zhu, Deyao and Chen, Jun and Shen, Xiaoqian and Li, Xiang and Elhoseiny, Mohamed},
  journal = {arXiv preprint arXiv:2304.10592},
  year    = {2023}
}

@inproceedings{CLIP_DPO_2024,
  title     = {{CLIP-DPO}: Vision-Language Models as a Source of Preference for Fixing Hallucinations in {LVLMs}},
  author    = {Ouali, Yassine and Bulat, Adrian and Martinez, Brais and Tzimiropoulos, Georgios},
  booktitle = {Proceedings of the European Conference on Computer Vision},
  year      = {2024}
}

@inproceedings{CLIP_2021,
  title     = {Learning Transferable Visual Models From Natural Language Supervision},
  author    = {Radford, Alec and Kim, Jong Wook and Hallacy, Chris and Ramesh, Aditya and Goh, Gabriel and Agarwal, Sandhini and Sastry, Girish and Askell, Amanda and Mishkin, Pamela and Clark, Jack and others},
  booktitle = {International conference on machine learning},
  year      = {2021}
}

@article{EFUF_2024,
  title   = {{EFUF}: Efficient Fine-grained Unlearning Framework for Mitigating Hallucinations in Multimodal Large Language Models},
  author  = {Xing, Shangyu and Zhao, Fei and Wu, Zhen and An, Tuo and Chen, Weihao and Li, Chunhui and Zhang, Jianbing and Dai, Xinyu},
  journal = {arXiv preprint arXiv:2402.09801},
  year    = {2024}
}

@article{Silkie_2023,
  title   = {Silkie: Preference Distillation for Large Visual Language Models},
  author  = {Li, Lei and Xie, Zhihui and Li, Mukai and Chen, Shunian and Wang, Peiyi and Chen, Liang and Yang, Yazheng and Wang, Benyou and Kong, Lingpeng},
  journal = {arXiv preprint arXiv:2312.10665},
  year    = {2023}
}

@inproceedings{ZeRO_2020,
  title     = {{ZeRO}: Memory Optimizations Toward Training Trillion Parameter Models},
  author    = {Rajbhandari, Samyam and Rasley, Jeff and Ruwase, Olatunji and He, Yuxiong},
  booktitle = {SC20: International Conference for High Performance Computing, Networking, Storage and Analysis},
  year      = {2020}
}

@article{Llama_2_2023,
  title   = {Llama 2: Open Foundation and Fine-Tuned Chat Models},
  author  = {Touvron, Hugo and Martin, Louis and Stone, Kevin and Albert, Peter and Almahairi, Amjad and Babaei, Yasmine and Bashlykov, Nikolay and Batra, Soumya and Bhargava, Prajjwal and Bhosale, Shruti and others},
  journal = {arXiv preprint arXiv:2307.09288},
  year    = {2023}
}

@article{EOS_2024,
  title   = {{Less is More}: Mitigating Multimodal Hallucination from an {EOS} Decision Perspective},
  author  = {Yue, Zihao and Zhang, Liang and Jin, Qin},
  journal = {arXiv preprint arXiv:2402.14545},
  year    = {2024}
}

@article{HALVA_2024,
  title   = {Data-augmented phrase-level alignment for mitigating object hallucination},
  author  = {Sarkar, Pritam and Ebrahimi, Sayna and Etemad, Ali and Beirami, Ahmad and Ar{\i}k, Sercan {\"O} and Pfister, Tomas},
  journal = {arXiv preprint arXiv:2405.18654},
  year    = {2024}
}

@article{GPT_4o_2024,
  title   = {{GPT-4o} System Card},
  author  = {Hurst, Aaron and Lerer, Adam and Goucher, Adam P and Perelman, Adam and Ramesh, Aditya and Clark, Aidan and Ostrow, AJ and Welihinda, Akila and Hayes, Alan and Radford, Alec and others},
  journal = {arXiv preprint arXiv:2410.21276},
  year    = {2024}
}

@inproceedings{VQA_v2_2017,
  title     = {Making the {V} in {VQA} Matter: Elevating the Role of Image Understanding in Visual Question Answering},
  author    = {Goyal, Yash and Khot, Tejas and Summers-Stay, Douglas and Batra, Dhruv and Parikh, Devi},
  booktitle = {Proceedings of the IEEE Conference on Computer Vision and Pattern Recognition},
  year      = {2017}
}

@article{NoiseBoost_2024,
  title   = {{NoiseBoost}: Alleviating Hallucination with Noise Perturbation for Multimodal Large Language Models},
  author  = {Wu, Kai and Jiang, Boyuan and Jiang, Zhengkai and He, Qingdong and Luo, Donghao and Wang, Shengzhi and Liu, Qingwen and Wang, Chengjie},
  journal = {arXiv preprint arXiv:2405.20081},
  year    = {2024}
}

@article{Volcano_2024,
  title   = {Volcano: mitigating multimodal hallucination through self-feedback guided revision},
  author  = {Lee, Seongyun and Park, Sue Hyun and Jo, Yongrae and Seo, Minjoon},
  journal = {arXiv preprint arXiv:2311.07362},
  year    = {2023}
}

@inproceedings{NLTK_2002,
  title     = {{NLTK}: The Natural Language Toolkit},
  author    = {Bird, Steven},
  booktitle = {Proceedings of the COLING/ACL 2006 interactive presentation sessions},
  year      = {2006}
}

@article{OmDet_2024,
  title     = {{OmDet}: Large-scale vision-language multi-dataset pre-training with multimodal detection network},
  author    = {Zhao, Tiancheng and Liu, Peng and Lee, Kyusong},
  journal   = {IET Computer Vision},
  year      = {2024},
  publisher = {Wiley Online Library}
}

@inproceedings{LlamaFactory_2024,
  title     = {{LlamaFactory}: Unified Efficient Fine-Tuning of 100+ Language Models},
  author    = {Yaowei Zheng and Richong Zhang and Junhao Zhang and Yanhan Ye and Zheyan Luo and Zhangchi Feng and Yongqiang Ma},
  booktitle = {Proceedings of the 62nd Annual Meeting of the Association for Computational Linguistics (Volume 3: System Demonstrations)},
  publisher = {Association for Computational Linguistics},
  year      = {2024}
}

@article{SimPO_2024,
  title   = {{SimPO}: Simple Preference Optimization with a Reference-Free Reward},
  author  = {Meng, Yu and Xia, Mengzhou and Chen, Danqi},
  journal = {Advances in Neural Information Processing Systems},
  year    = {2024}
}

@article{Uni_DPO_2026,
  title   = {{Uni-DPO}: A Unified Paradigm for Dynamic Preference Optimization of {LLMs}},
  author  = {Peng, Shangpin and Wang, Weinong and Tian, Zhuotao and Yang, Senqiao and Wu, Xing and Xu, Haotian and Zhang, Chengquan and Isobe, Takashi and Hu, Baotian and Zhang, Min},
  journal = {arXiv preprint arXiv:2506.10054},
  year    = {2025}
}

@article{DPO_positive_2024,
  title   = {Smaug: Fixing Failure Modes of Preference Optimisation with {DPO-Positive}},
  author  = {Pal, Arka and Karkhanis, Deep and Dooley, Samuel and Roberts, Manley and Naidu, Siddartha and White, Colin},
  journal = {arXiv preprint arXiv:2402.13228},
  year    = {2024}
}
}

\clearpage
\appendix
\setcounter{page}{1}
\maketitlesupplementary

\section*{Overview}
This material provides supplementary details to the main paper, including the following sections:
\vspace{0.3cm}
\begin{itemize}[itemsep=0pt, parsep=3pt, leftmargin=*]
    \item (\ref{sec:motivation_details}) \textbf{Motivation Details}
          \begin{itemize}
              \item (\ref{subsec:object_position_distribution}) Object Position Distribution
              \item (\ref{subsec:decode_based_early_intervention}) Decode Based Early Intervention
          \end{itemize}
    \item (\ref{sec:method_details}) \textbf{Method Details}
          \begin{itemize}
              \item (\ref{subsec:object_extraction_details}) Object Extraction
              \item (\ref{subsec:iterative_contextual_booststrapping}) Iterative Contextual Booststrapping
              \item (\ref{subsec:object_detector_selection}) Selection of Object Detector
              \item (\ref{subsec:uncertain_objects_treatment}) Treatment of Uncertain Objects
          \end{itemize}
    \item (\ref{sec:training_details}) \textbf{Training Details}
          \begin{itemize}
              \item (\ref{subsec:training_dataset}) Training Dataset
              \item (\ref{subsec:training_setup}) Training Setup
              \item (\ref{subsec:training_objective}) Training Objective
          \end{itemize}
    \item (\ref{sec:evaluation_details}) \textbf{Evaluation Details}
          \begin{itemize}
              \item (\ref{subsec:evaluation_benchmarks}) Evaluation Benchmarks
              \item (\ref{subsec:evaluation_counterparts}) Evaluation Counterparts
              \item (\ref{subsec:detailed_evaluation_settings}) Evaluation Settings
              \item (\ref{subsec:detailed_evaluation_results}) Evaluation Results
              \item (\ref{subsec:ablation_study_details}) Details of Ablation Study
          \end{itemize}
    \item (\ref{sec:supp_sentinel_other_baselines}) \textbf{SENTINEL with Other Baselines}
    \item (\ref{sec:supp_related_works})  \textbf{Related Work}
    \item (\ref{sec:additional_case_studies}) \textbf{Additional Case Studies}
\end{itemize}

\section{Motivation Details}
\label{sec:motivation_details}

In this section, we deepen the discussion supporting the key observations from the main paper.

\subsection{Object Position Distribution}
\label{subsec:object_position_distribution}

Following the approach of Caption Hallucination Assessment with Image Relevance~\cite{Obj_HalBench_2018}, we select 300 images from the \href{https://cocodataset.org/#home}{COCO2014} dataset and use the provided captions and segmentation annotations as references to determine whether the objects described by the model exist in the images. As shown in the main paper Fig.~2, as the model generates longer outputs, the number of real objects described decreases while hallucinated objects increase, indicating that hallucinations of the model become more severe with output length. Notably, towards the end of the generation (around the last 10\% tokens), both the number of hallucinated and real objects decreases. This is because, towards the end of the image description, the model tends to conclude with abstract summaries about the atmosphere or emotions rather than providing concrete object descriptions.

\subsection{Decode Based Early Intervention}
\label{subsec:decode_based_early_intervention}

As a preliminary investigation, we explore a training-free approach to mitigating object hallucinations in MLLMs. In essence, our method dynamically verifies each generated sentence against the image content and filters out any hallucinated ones before proceeding. Specifically, for the image captioning task, we sample multiple candidate sentences ($n\! =\! 5$) from the model's output, stopping generation at the first period. These candidate sentences are then parsed using SceneGraphParser~\cite{FACTUAL_2023} to extract mentioned objects. We subsequently employ an open-vocabulary object detector, Grounding DINO~\cite{Grounding_DINO_2024}, to verify the existence of these objects in the image. A sentence without hallucinated objects is selected as the current generated sentence, and then continues generating the subsequent content.

\begin{figure}[t!]
    \centering
    \includegraphics[width=0.47\textwidth]{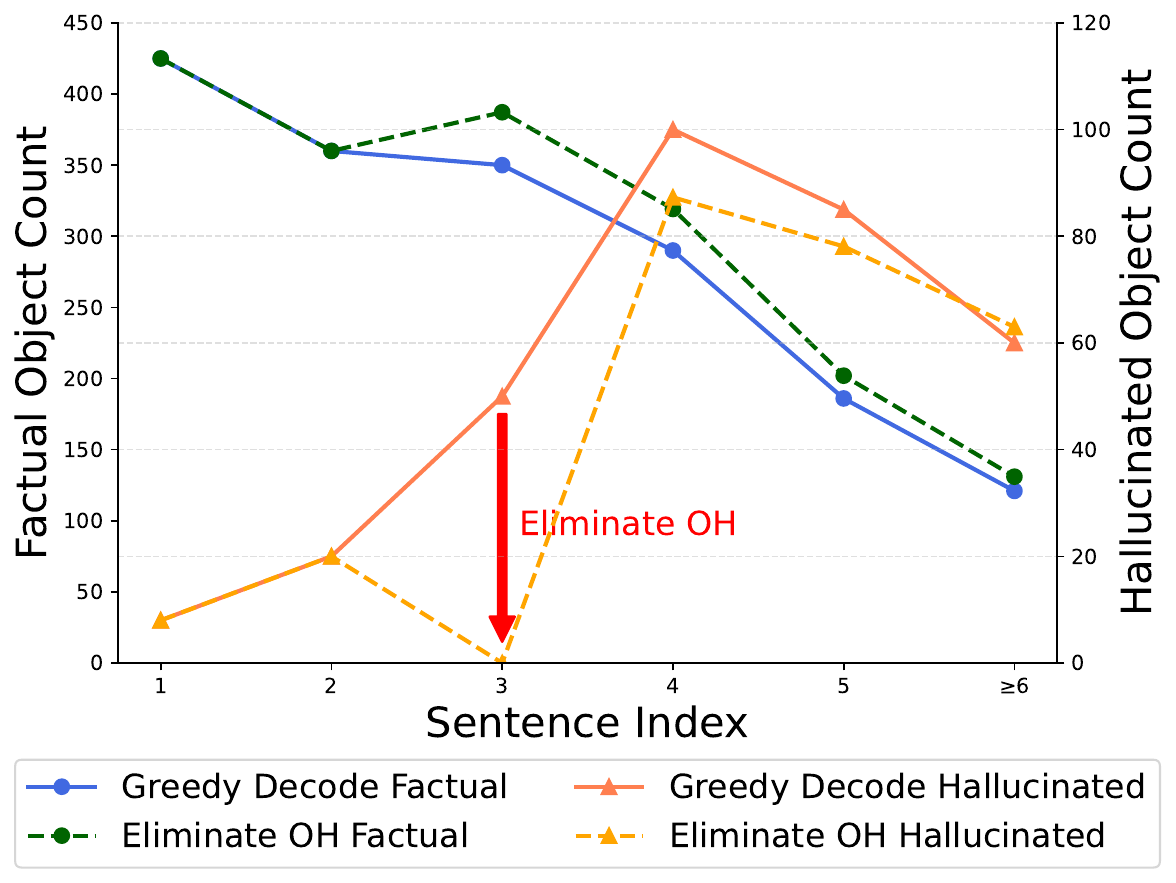}
    \vspace{-10pt}
    \caption{\textbf{Effect of intermediate hallucination mitigation on subsequent generations.} Showing the effectiveness of early-stage intervention in mitigating the propagation of hallucinations.}
    \label{fig:eliminate_sent_3}
    \vspace{-5pt}
\end{figure}

\begin{figure}[t!]
    \centering
    \includegraphics[width=0.47\textwidth]{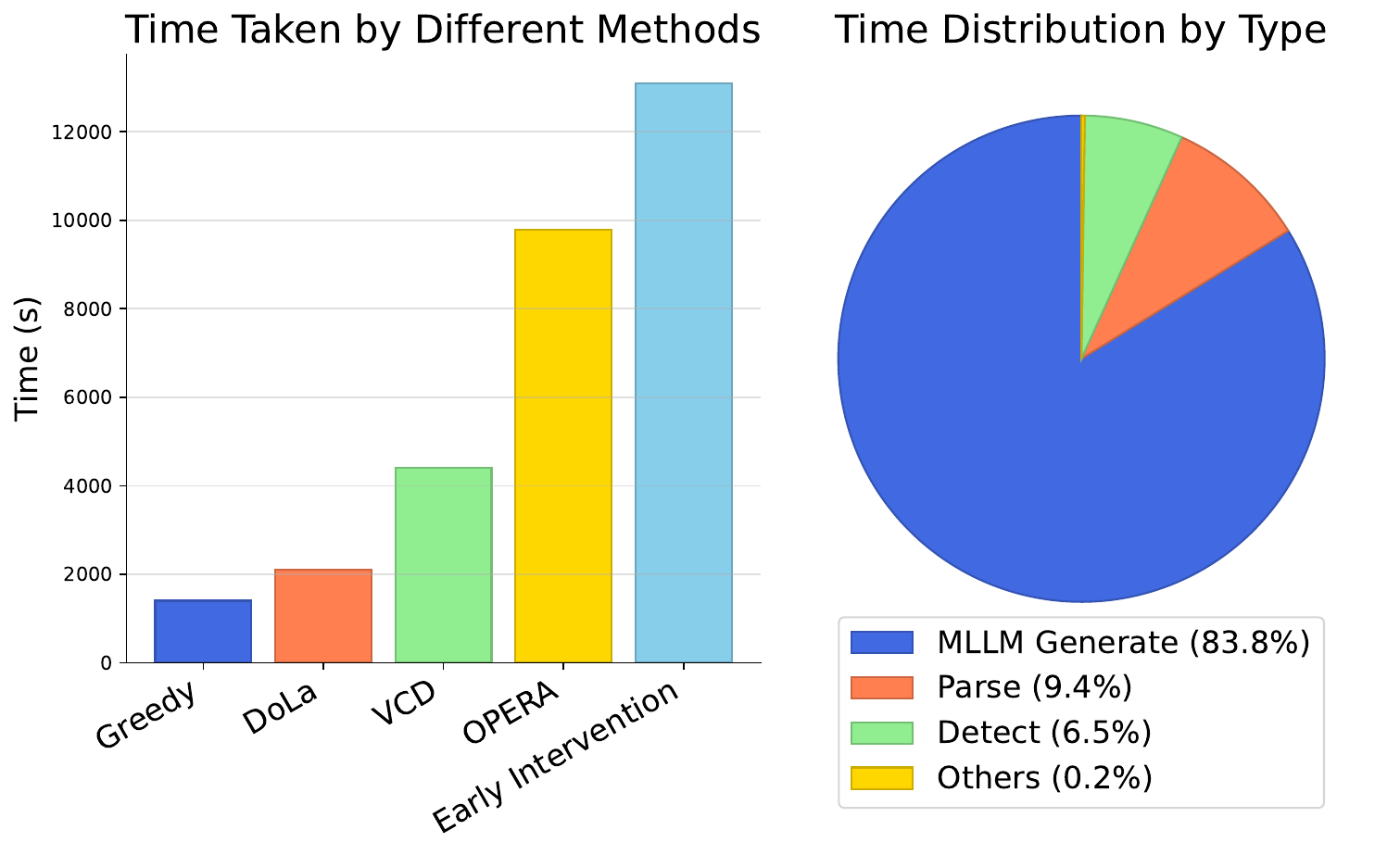}
    \vspace{-10pt}
    \caption{\textbf{Time cost analysis of decode-based methods.} Decode-based early intervention increases inference time, primarily due to the additional generation steps required by MLLM sampling, whereas the object detector remains highly efficient.}
    \label{fig:time_taken}
    \vspace{-10pt}
\end{figure}

\begin{table}[t!]
  \centering
  \resizebox{\columnwidth}{!} {%
    \begin{tabular}{ll ccccc}
      \toprule
      \multirow{2}{*}{\vspace{-2mm}\textbf{Model}}          &
      \multirow{2}{*}{\vspace{-2mm}\textbf{Method}}         &
      \multicolumn{2}{c}{\textbf{Object HalBench}}          &
      \multicolumn{3}{c}{\textbf{AMBER}}
      \\
      \cmidrule(lr){3-4}
      \cmidrule(lr){5-7}
                                                            &
                                                            &
      Resp. $\downarrow$                                    &
      Ment. $\downarrow$                                    &
      CHAIR $\downarrow$                                    &
      Hal $\downarrow$                                      &
      Cog $\downarrow$
      \\
      \midrule
      \multirow{7}{*}{LLaVA-v1.5-7B~\cite{LLaVA_v1_5_2024}} & baseline                                                         & 52.7          & 27.9          & 8.4          & 35.5          & 4.0          \\
                                                            & Woodpecker~\cite{Woodpecker_2023}                                & 39.6          & 26.4          & -            & -             & -            \\
                                                            & VCD~\cite{VCD_2023}                                              & 52.7          & 27.3          & 9.1          & 39.8          & 4.2          \\
                                                            & OPERA~\cite{OPERA_2024}                                          & 40.0          & 21.9          & 6.5          & 28.5          & 3.1          \\
                                                            & EOS~\cite{EOS_2024}                                              & 40.0          & 22.2          & 6.4          & 27.4          & 2.6          \\
                                                            & HA-DPO~\cite{HA_DPO_2023}                                        & 37.0          & 20.9          & 6.7          & 30.9          & 3.3          \\
                                                            & \makecell{Decode based\phantom{aaa}\nextline early intervention} & \textbf{33.5} & \textbf{17.6} & \textbf{5.5} & \textbf{26.8} & \textbf{2.6} \\
      \bottomrule
    \end{tabular}
  }
  \vspace{-10pt}
  \caption{
    \textbf{Effectiveness of decode based early intervention.}
  }
  \vspace{-15pt}
  \label{tab:decode_early_intervention}
\end{table}

This approach effectively prevents the further propagation of hallucinations. As shown in the main paper Fig.~2b, even when applied at just a single sentence, eliminating hallucinations as early as the second sentence significantly reduces the likelihood of generating hallucinated objects in subsequent outputs. A similar effect is observed when intervention occurs only at the third sentence, as shown in~\cref{fig:eliminate_sent_3}.

When this early intervention strategy is applied throughout the entire caption generation process, as shown in~\cref{tab:decode_early_intervention}, it effectively mitigates object hallucinations when evaluated on the Object Halbench~\cite{Obj_HalBench_2018} benchmark. However, as illustrated in~\cref{fig:time_taken}, it increases inference time, primarily due to the additional sampling time of the MLLM, while the object detector remains highly efficient. These findings highlight the detector's role as both an effective and computationally efficient component, reinforcing its potential for constructing high-quality training data for hallucination mitigation.

\section{Method Details}
\label{sec:method_details}
In this section, we detail our methods for extracting concrete objects from models' outputs in~\cref{subsec:object_extraction_details}, and propose iterative contextual bootstrapping (ICB) to enhance robustness in~\cref{subsec:iterative_contextual_booststrapping}. In~\cref{subsec:object_detector_selection}, we discuss the selection of the object detector. Finally, in~\cref{subsec:uncertain_objects_treatment}, we describe how we handle uncertain objects. Our approach reduces hallucinations efficiently without relying on large auxiliary models.

\subsection{Object Extraction}
\label{subsec:object_extraction_details}
In this section, we detail our approach to extracting the mentioned objects from the model's output automatically and efficiently. Our objective is to obtain identifiable and concrete entity descriptions, following a structured pipeline.

First, we employ SceneGraphParser~\cite{FACTUAL_2023} to convert the input descriptions into a series of triplets representing relationships within the scene. Specifically, each triplet is treated as a (subject, predicate, object) tuple. For example:
\begin{quote}
    \textit{``A little black cat sits on a chair next to a table.''}
\end{quote}
is parsed into the following structured triplets:
\[
    \begin{aligned}
         & (\text{cat, is, little})       & (\text{cat, is, black})     \\
         & (\text{chair, next to, table}) & (\text{cat, sit on, chair})
    \end{aligned}
\]

Next, we extract entities from these triplets. We apply the following rules:
\begin{itemize}
    \item If the predicate belongs to \{``is", ``are"\}, it represents an attribute relationship. In this case, we consider only the subject as a potential entity.
    \item Otherwise, both the subject and object are considered potential entities.
\end{itemize}

To refine the entity extraction process, we leverage the \href{https://spacy.io/usage}{SpaCy} natural language processing library to analyze the part of speech (POS) of the extracted candidates and filter out words that are neither nouns nor proper nouns. Furthermore, we utilize NLTK's WordNet Lemmatizer~\cite{NLTK_2002} in conjunction with a lexicographic filtering mechanism to exclude non-entity nouns. Specifically, we examine the lexicographer category of each word, and if it falls within the following non-concrete categories, it is removed:

$$
    \begin{array}{llll}
        \text{noun.feeling},       & \text{noun.attribute}, \\ \text{noun.state}, & \text{noun.shape}, \\
        \text{noun.time},          & \text{noun.quantity},  \\ \text{noun.cognition}, & \text{noun.event}, \\
        \text{noun.communication}, & \text{noun.relation},  \\ \text{noun.act}, & \text{noun.location}.
    \end{array}
$$

Our method effectively extracts entities without the need for large auxiliary models such as GPT-4~\cite{GPT4_2023} or LLaMA-2-70B~\cite{Llama_2_2023}. Instead, it relies solely on lightweight NLP tools and libraries, ensuring both high extraction accuracy and maintaining an open-vocabulary nature.

\subsection{Iterative Contextual Booststrapping}
\label{subsec:iterative_contextual_booststrapping}

To ensure robustness across different contexts, we introduce the iterative contextual bootstrapping (ICB) strategy, as shown in the main paper Fig.~5. By leveraging contextually bootstrapped data, early intervention can be seamlessly integrated into diverse contexts, effectively mitigating hallucinations and enhancing robustness.

To further investigate the impact of iterative contextual bootstrapping (ICB), we conduct an ablation study where we exclude ICB and instead sample a non-hallucinated description $\vec{y}^{+}_{w}$ only at the first occurrence of hallucination, using it as the positive sample during constructing pairs, while the original hallucinated description serves as the negative sample $\vec{y}_{l}$. We then train the model using the same method and dataset size mentioned in the main paper. The results, as presented in~\cref{tab:iter_result}, demonstrate that our approach, when incorporating ICB, exhibits greater robustness and effectively reduces hallucinations across different scenarios.
\begin{table}
    \centering
    \resizebox{\columnwidth}{!} {%
        \begin{tabular}{l ccccccc}
            \toprule
            \multirow{2}{*}{\vspace{-2mm}\textbf{Method}} &
            \multicolumn{2}{c}{\textbf{Object HalBench}}  &
            \multicolumn{3}{c}{\textbf{AMBER}}            &
            \multicolumn{1}{c}{\textbf{MM-Vet}}
            \\
            \cmidrule(lr){2-3}
            \cmidrule(lr){4-6}
            \cmidrule(lr){7-7}                            &
            Resp. $\downarrow$                            &
            Ment. $\downarrow$                            &
            CHAIR $\downarrow$                            &
            Hal $\downarrow$                              &
            Cog $\downarrow$                              &
            Overall $\uparrow$
            \\
            \midrule
            LLaVA-v1.5-7B                                 & 52.7           & 27.9           & 8.4            & 35.5            & 4.0            & 31.1              \\
            Ours w/ ICB                                   & \textbf{4.3}   & \textbf{2.6}   & \textbf{2.9}   & \textbf{14.6}   & \textbf{1.2}   & \textbf{32.6}     \\
            Ours w/o ICB                                  & 5.3\upred{1.0} & 3.2\upred{0.6} & 3.1\upred{0.2} & 14.9\upred{0.3} & 1.4\upred{0.2} & 31.8\downred{0.8} \\
            \bottomrule
        \end{tabular}
    }
    \vspace{-10pt}
    \caption{
        \textbf{Effect of Iterative Contextual Booststrapping.}
        Iterative Contextual Bootstrapping (ICB) enables early intervention to be seamlessly integrated into various contexts, effectively mitigating hallucinations and ensuring robustness across different scenarios.
    }
    \vspace{-10pt}
    \label{tab:iter_result}
\end{table}

\subsection{Selection of Object Detector}
\label{subsec:object_detector_selection}

Detectors are more cost-effective for providing training guidance for MLLMs than human annotators. SENTINEL is not constrained to particular detectors; any model with open-world recognition ability can be employed. As shown in~\cref{tab:detector_type}, more effective detectors lead to superior performance, and the cross-validation technique effectively mitigates the phenomenon of false positives.

\begin{table}
    \centering
    \renewcommand{\arraystretch}{0.87}
    \resizebox{\linewidth}{!}{%
        \begin{tabular}{lcc}
            \toprule
            \multirow{2}{*}{\vspace{-1.5mm}\textbf{Method}}
                                                                                          & \multicolumn{2}{c}{\textbf{Object HalBench}}
            \\
            \cmidrule(lr){2-3}
                                                                                          & Resp.\,$\downarrow$                          & Ment.\,$\downarrow$ \\
            \midrule
            LLaVA-v1.5-7B                                                                 & 52.7                                         & 28.0                \\
            OmDet~\cite{OmDet_2024}                                                       & 19.3                                         & 9.9                 \\
            Grounding DINO~\cite{Grounding_DINO_2024}                                     & 14.3                                         & 7.7                 \\
            YOLO World~\cite{YOLO_World_2024}                                             & 12.3                                         & 6.9                 \\
            Grounding DINO~\cite{Grounding_DINO_2024} + YOLO World~\cite{YOLO_World_2024} & \textbf{6.6}                                 & \textbf{3.8}        \\
            \bottomrule
        \end{tabular}
    }
    \vspace{-9pt}
    \caption{
        \textbf{Results with different detectors.}
        We observe that detector OmDet~\cite{OmDet_2024} often produces false positives, identifying objects that do not exist in the images, which may lead to less reliable results. Generally, detectors with more human-like real-world perception abilities yield better performance.
    }
    \label{tab:detector_type}
    \vspace{-5pt}
\end{table}

\subsection{Treatment of Uncertain Objects}
\label{subsec:uncertain_objects_treatment}

As mentioned in the main paper, we ignore uncertain objects to maintain data quality and reduce detector bias. We also conduct ablation studies that treat uncertain objects alternately as factual or hallucinated. \cref{tab:uncertain_objects_treatment} shows that ignoring uncertain objects yields better results. We hypothesize that it is because 1) `\texttt{uncertain}'$\Rightarrow$`\texttt{factual}' may bring hallucinations to the context during iterative contextual bootstrapping (ICB), contradicting the early intervention strategy based on the hallucination-free contexts. 2) `\texttt{uncertain}'$\Rightarrow$`\texttt{hallucinated}' may introduce noisy and ambiguous negative samples for preference learning.

\begin{table}
    \centering
    \renewcommand{\arraystretch}{0.87}
    \resizebox{\linewidth}{!}{%
        \begin{tabular}{l cccc}
            \toprule
            \multirow{2}{*}{\vspace{-2mm}\textbf{Method}} &
            \multicolumn{2}{c}{\textbf{Object HalBench}}  &
            \multicolumn{1}{c}{\textbf{MM-Vet}}
            \\
            \cmidrule(lr){2-3} \cmidrule(lr){4-4}         &
            Resp. $\downarrow$                            &
            Ment. $\downarrow$                            &
            Overall $\uparrow$
            \\
            \midrule
            LLaVA-v1.5-7B                                 & 52.7         & 28.0         & 31.0          \\
            Ignore uncertain                              & \textbf{4.3} & \textbf{2.6} & \textbf{32.6} \\
            Uncertain as factual                          & 10.3         & 6.9          & 31.8          \\
            Uncertain as hallucinated                     & 8.3          & 5.0          & 32.0          \\
            \bottomrule
        \end{tabular}
    }
    \vspace{-9pt}
    \caption{
        \textbf{Treatments of uncertain objects.}
        Ignoring uncertain objects can improve the quality of training data, thereby enhancing final model performance.
    }
    \label{tab:uncertain_objects_treatment}
    \vspace{-9pt}
\end{table}

\section{Training Details}
\label{sec:training_details}

In this section, we provide a detailed overview of the preference training process. The dataset used for training is described in~\cref{subsec:training_dataset}, the training setup is outlined in~\cref{subsec:training_setup}, and the training objective is analyzed in detail in~\cref{subsec:training_objective}.

\subsection{Training Dataset}
\label{subsec:training_dataset}

\mypara{Visual Genome.}
\href{https://homes.cs.washington.edu/~ranjay/visualgenome/index.html}{Visual Genome (VG)}~\cite{Visual_Genome_2017} is a publicly available large-scale vision-language dataset that provides dense annotations for about 108K images, with each image containing an average of 21 objects, 18 attributes, and 18 object relationships. In addition to object annotations, VG includes 1.7 million visual question-answering pairs in a multi-choice format, covering six question types: What, Where, When, Who, Why, and How. Compared to traditional VQA datasets, VG offers a more balanced distribution of question types while also serving as one of the most comprehensive resources for bridging visual concepts with language. In our study, VG images are utilized for constructing the training dataset.

\mypara{Training Data.}
We use approximately 4K images from VG for training dataset construction, selected based on their appropriate information density and appropriate level of object diversity. Notably, we do not utilize any labels or ground-truth annotations from VG or other datasets when constructing the preference dataset. Instead, our approach automatically and efficiently generates highly discriminative preference training data in a cost-effective manner.

\subsection{Training Setup}
\label{subsec:training_setup}

We strictly follow the official setup provided by \href{https://github.com/haotian-liu/LLaVA}{LLaVA} to ensure reproducibility. The details of the training hyperparameters used in our training are presented in~\cref{tab:training_details}.

\begin{table}[t!]
  \centering
  \renewcommand{\arraystretch}{0.93}
  \resizebox{\columnwidth}{!} {%
    \begin{tabular}{l|cc}
      \toprule
      \diagbox{Setting}{Model}                   &
      \multicolumn{1}{c}{\textbf{LLaVA-v1.5-7B}} &
      \multicolumn{1}{c}{\textbf{LLaVA-v1.5-13B}}
      \\
      \midrule
      LLM                                        & \multicolumn{1}{c}{Vicuna-v1.5-7B}                                     & Vicuna-v1.5-13B \\
      Vision encoder                             & \multicolumn{2}{c}{CLIP ViT-L\textsubscript{336px/14}\cite{CLIP_2021}}                   \\
      Projector                                  & \multicolumn{2}{c}{mlp2x\_gelu}                                                          \\
      \midrule
      Learning rate                              & \multicolumn{1}{c}{2e-6}                                               & 3e-6            \\
      Batch size per GPU                         & \multicolumn{1}{c}{16}                                                 & 8               \\
      Trainable parameters                       & \multicolumn{2}{c}{LoRA trains only LLM's linear layers.}                                \\
      LoRA rank $r$                              & \multicolumn{2}{c}{128}                                                                  \\
      LoRA alpha $\alpha$                        & \multicolumn{2}{c}{256}                                                                  \\
      LoRA beta $\beta$                          & \multicolumn{2}{c}{0.1}                                                                  \\
      Projector lr                               & \multicolumn{2}{c}{0}                                                                    \\
      Learning rate  scheduler                   & \multicolumn{2}{c}{Cosine}                                                               \\
      Optimizer                                  & \multicolumn{2}{c}{AdamW~\cite{AdamW_2019}}                                              \\
      Model max lenght                           & \multicolumn{2}{c}{2048}                                                                 \\
      Weight decay                               & \multicolumn{2}{c}{0.}                                                                   \\
      Epochs                                     & \multicolumn{2}{c}{1}                                                                    \\
      Global batch size                          & \multicolumn{2}{c}{64}                                                                   \\
      Memory optimization                        & \multicolumn{2}{c}{ZeRO stage 2~\cite{ZeRO_2020}}                                        \\
      \bottomrule
    \end{tabular}
  }
  \vspace{-10pt}
  \caption{
    \textbf{Training hyperparameters used in our experiments.}
  }
  \vspace{-10pt}
  \label{tab:training_details}
\end{table}

\subsection{Training Objective}
\label{subsec:training_objective}

As shown in the main paper Eq.~(2), we employ the context-aware DPO (C-DPO) objective to train the model to differentiate between hallucinated and non-hallucinated content at the first occurrence of hallucination, aiming to mitigate its propagation. In this section, we provide a detailed analysis of (1) the rationale for excluding context $\vec{c}$ from the loss computation, (2) the key differences between our proposed C-DPO and the standard DPO, and (3) a comparison between our training objective and Mask-DPO~\cite{Mask_DPO_2025}.

\mypara{Why mask context in loss calculation?}
We implemented a pseudocode for calculation based on the context-aware DPO (C-DPO) formula. As shown in~\cref{alg:cdpo_calculation}, to compute the C-DPO loss, we need to evaluate the log probabilities (logps) of the output tokens given an input. If we do not mask out the context during loss computation, the context $\vec{c}$ remains identical in both positive and negative samples. Since the context and its preceding tokens are the same, for the policy model, the logps of the context tokens will be the same across both forward passes. This adds an identical term $C$ to both policy\_chosen\_logps and policy\_rejected\_logps, which cancels out in the policy\_logratios computation at line 7, leaving the loss unaffected.

From a gradient perspective, since $C$ is derived from the same model parameters $\theta$ based on identical preceding tokens in both forward passes, its gradient remains the same due to the autoregressive nature of the model. As a result, this gradient term cancels out as well and does not affect model training. Therefore, to reduce unnecessary computation and mitigate potential numerical errors, we exclude the context $\vec{c}$ from the loss calculation in C-DPO.

\begin{algorithm}[h]
    \caption{Pseudocode for C-DPO Training}
    \begin{algorithmic}[1]
        \scriptsize
        \Statex \textbf{Input:} Training sample $(\vec{v}, \vec{q}, \vec{c}, \vec{y}^{+}_{w}, \vec{y}_{l})$
        \Statex \textbf{Output:} C-DPO loss
        \Statex
        \State import torch
        \State import torch.nn.functional as F
        \State
        \State def get\_cdpo\_loss(self, $(\vec{v}, \vec{q}, \vec{c}, \vec{y}^{+}_{w}, \vec{y}_{l})$) $\to$ torch.Tensor:
        \State \hspace{\algorithmicindent}\textbf{\# policy model forward pass}
        \State \hspace{\algorithmicindent}policy\_chosen\_logps = model.dpo\_forward($(\vec{v}, \vec{q}, \vec{c}, \vec{y}^{+}_{w})$)
        \State \hspace{\algorithmicindent}policy\_rejected\_logps = model.dpo\_forward($(\vec{v}, \vec{q}, \vec{c}, \vec{y}_{l})$)
        \State \hspace{\algorithmicindent}policy\_logratios = policy\_chosen\_logps - policy\_rejected\_logps
        \State
        \State \hspace{\algorithmicindent}\textbf{\# reference model forward pass}
        \State \hspace{\algorithmicindent}with torch.no\_grad():
        \State \hspace{\algorithmicindent}\hspace{\algorithmicindent}ref\_chosen\_logps = ref\_model.dpo\_forward($(\vec{v}, \vec{q}, \vec{c}, \vec{y}^{+}_{w})$)
        \State \hspace{\algorithmicindent}\hspace{\algorithmicindent}ref\_rejected\_logps = ref\_model.dpo\_forward($(\vec{v}, \vec{q}, \vec{c}, \vec{y}_{l})$)
        \State \hspace{\algorithmicindent}ref\_logratios = ref\_chosen\_logps - ref\_rejected\_logps
        \State
        \State \hspace{\algorithmicindent}\textbf{\# compute C-DPO loss}
        \State \hspace{\algorithmicindent}logits = policy\_logratios - ref\_logratios
        \State \hspace{\algorithmicindent}loss = -F.logsigmoid(dpo\_beta * logits)
        \State \hspace{\algorithmicindent}return loss.mean()
        \Statex
        \Statex \Comment{model.dpo\_forward() returns the \textbf{sum} of the log probabilities of all tokens that have not been masked out.}
    \end{algorithmic}
    \label{alg:cdpo_calculation}
\end{algorithm}

\mypara{Comparison with Standard DPO.}
To validate the effectiveness of our proposed context-aware DPO (C-DPO), we conducted an additional experiment using a standard DPO for training. In this setup, no context $\vec{c}$ was included, and both $\vec{y}_{w}$ and $\vec{y}_{l}$ are complete image captions based on the given image $\vec{v}$ and prompt $\vec{q}$. $\vec{y}_{w}$ consisted of sentences with minimal hallucinations (using non-hallucinated context and ensuring the current sentence itself is hallucination-free until the end of generating), while $\vec{y}_{l}$ contained sentences with maximal hallucinations (using hallucinated context and ensuring the current sentence itself contained hallucinations until the end of generating). Both methods were trained on the same scale of data (8.6K samples).

As shown in~\cref{tab:standard_dpo}, our proposed context-aware DPO (C-DPO) more effectively guides the model in distinguishing hallucinated content from non-hallucinated content, leading to improved hallucination suppression while maintaining generalization capabilities.

\begin{figure}[htbp]
    \centering
    \begin{subfigure}[b]{\linewidth}
        \includegraphics[width=\linewidth]{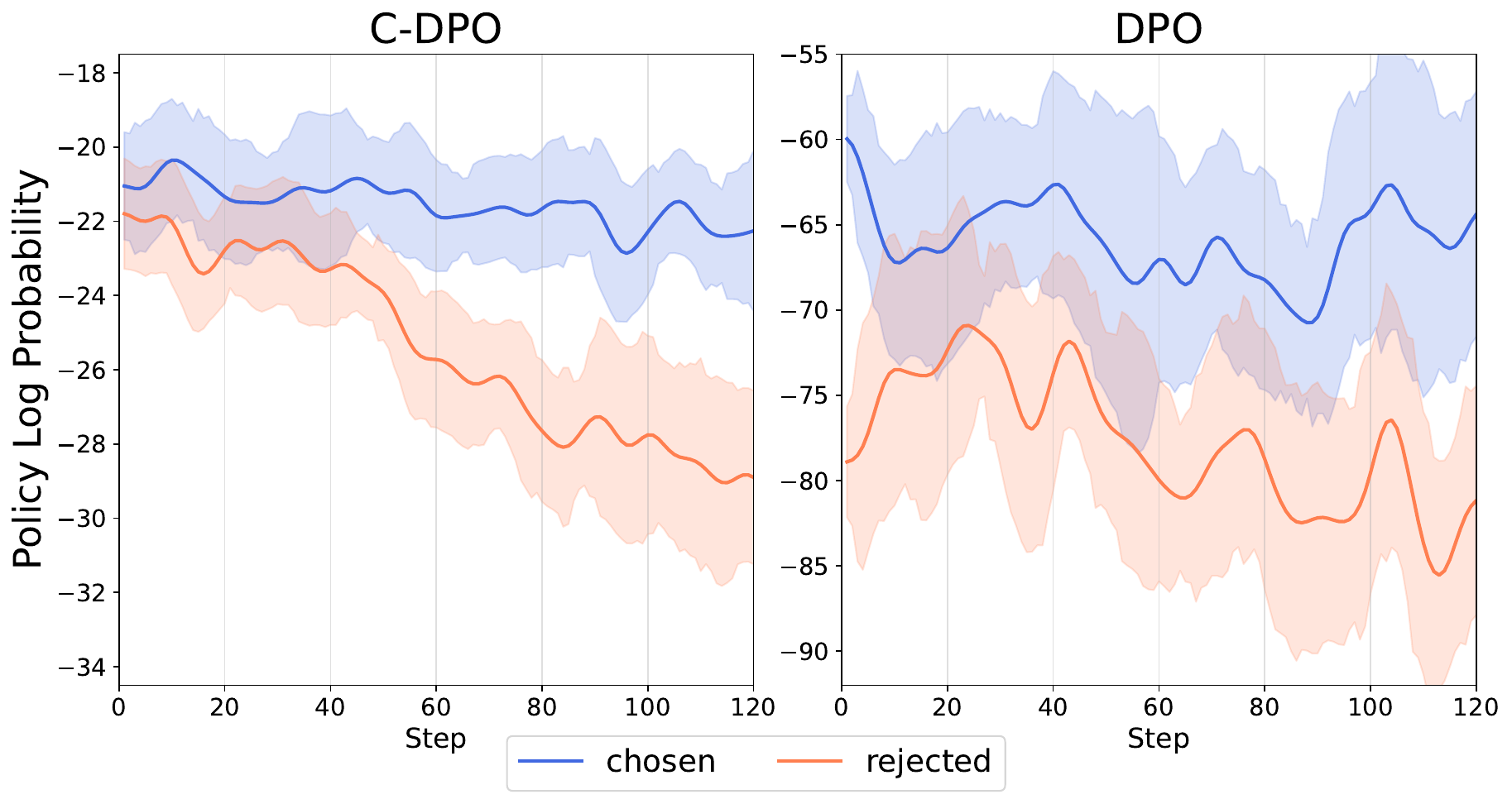}
        \vspace{-15pt}
        \caption{Logps comparison}
        \label{fig:logps_dpo}
    \end{subfigure}
    \vspace{12pt}
    \begin{subfigure}[b]{\linewidth}
        \includegraphics[width=\linewidth]{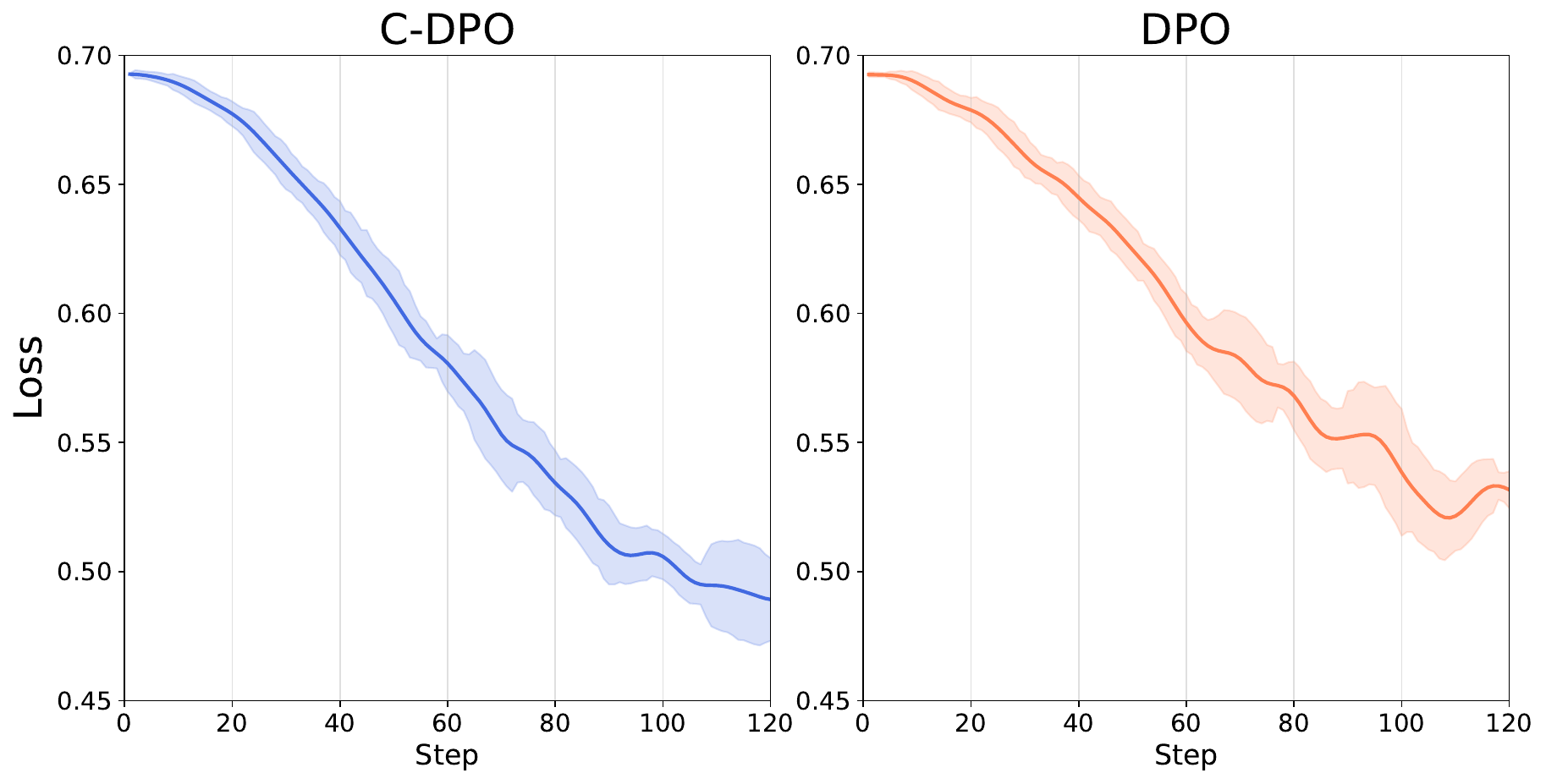}
        \vspace{-12pt}
        \caption{Loss comparison}
        \label{fig:loss_dpo}
    \end{subfigure}
    \vspace{-30pt}
    \caption{\textbf{Comparison between C-DPO and standard DPO during model training.} The proposed C-DPO promotes more stable gradient updates, enhancing training stability.}
    \label{fig:logps_loss_dpo}
    \vspace{-15pt}
\end{figure}

To further analyze the underlying reasons, we track the training dynamics of both objectives, including policy model log probabilities (logps) and loss. As illustrated in~\cref{fig:logps_loss_dpo}, the standard DPO exhibits greater logps variations between $\vec{y}_{w}$ and $\vec{y}_{l}$ during training due to the substantial differences between sentence pairs. Prior studies by~\citet{DPO_2023} and~\citet{HA_DPO_2023} suggest that such variability can dominate gradient updates, potentially compromising training stability. This instability may hinder the model's ability to capture long-range dependencies, leading to slower convergence and a more gradual reduction in training loss.

\begin{table}[h]
    \centering
    \resizebox{\columnwidth}{!} {%
        \begin{tabular}{lcccc}
            \toprule
            \multirow{2}{*}{\vspace{-2mm}\textbf{Method}}                        &
            \multicolumn{2}{c}{\textbf{Object HalBench}\cite{Obj_HalBench_2018}} &
            \multicolumn{1}{c}{\textbf{TextVQA}\cite{TextVQA_2019}}              &
            \multicolumn{1}{c}{\textbf{MM-Vet}\cite{MM_Vet_2024}}                                                                                           \\
            \cmidrule(lr){2-3}\cmidrule(lr){4-4}\cmidrule(lr){5-5}               &
            {Resp. $\downarrow$}                                                 &
            {Ment. $\downarrow$}                                                 &
            {Acc $\uparrow$}                                                     &
            {Overall $\uparrow$}
            \\
            \midrule
            LLaVA-v1.5-7B                                                        & 52.7            & 27.9           & 58.2              & 31.0              \\
            C-DPO~\cref{eq:cdpo} (8.6K data)                                     & \textbf{4.3}    & \textbf{2.6}   & \textbf{58.2}     & \textbf{32.6}     \\
            Standard DPO~\cref{eq:dpo} (8.6K data)                               & 10.1\upred{5.8} & 5.5\upred{2.9} & 58.1\downred{0.1} & 31.7\downred{0.9} \\
            \bottomrule
        \end{tabular}

    }
    \vspace{-10pt}
    \caption{
        \textbf{Effectiveness of C-DPO.}
        Compared to standard DPO, C-DPO enables the model to better learn to distinguish between correct and incorrect responses at the onset of hallucination, effectively mitigating hallucinations from the outset.
    }
    \vspace{-5pt}
    \label{tab:standard_dpo}
\end{table}

\mypara{Differences between our objective and DPO with Mask.}
\citet{Mask_DPO_2025} propose a preference learning approach that selectively retains factually correct sentences from preferred samples while avoiding penalties on factual content within inferior samples, thereby mitigating ambiguity issues inherent in preference learning. While this method effectively prioritizes high-quality samples, it primarily relies on masking certain parts of the training data without fully considering their potential impact on model learning. As demonstrated in the main paper Tab.~4, within our workflow, the choice of context $\vec{c}$—which is masked from loss calculation during training—plays a crucial role in shaping the final training outcomes. This highlights the importance of carefully considering these factors to ensure that our approach effectively guides the model toward learning accurate and reliable knowledge.

\section{Evaluation Details}
\label{sec:evaluation_details}

In this section, we provide detailed information about the evaluation process. The evaluation benchmarks we used are described in~\cref{subsec:evaluation_benchmarks}, where we showcase the strong performance of our method. In~\cref{subsec:evaluation_counterparts}, we outline the counterparts used for comparison. In~\cref{subsec:detailed_evaluation_settings}, we present the detailed evaluation setup. In~\cref{subsec:detailed_evaluation_results}, we provide detailed results from some of the experiments. Additionally, in~\cref{subsec:ablation_study_details}, we present specific details of the ablation studies.

\subsection{Evaluation Benchmarks}
\label{subsec:evaluation_benchmarks}

We provide a detailed description of the evaluation benchmarks used in our study.

\begin{itemize}[leftmargin=*]
    \vspace{+0.5mm}
    \item \textbf{Object HalBench.}
          Object HalBench~\cite{Obj_HalBench_2018} is a widely used benchmark for assessing common object hallucinations in detailed image descriptions. Following~\cite{RLAIF_V_2024}, we incorporate eight diverse prompts to enhance evaluation stability. We report two key metrics: the response-level hallucination rate (Resp.), which measures the proportion of responses containing hallucinations, and the mention-level hallucination rate (Ment.), which quantifies the percentage of hallucinated object mentions.
          \vspace{+0.5mm}
    \item \textbf{AMBER.}
          AMBER~\cite{AMBER_2023} is a widely used metric for hallucination evaluation, assessing the frequency of hallucinatory objects in model-generated responses.

          In the generative component of AMBER, hallucination is quantified using the following three metrics:

          CHAIR score:
          {
          \setlength{\abovedisplayskip}{6pt}
          \setlength{\belowdisplayskip}{6pt}
          \begin{equation}
              \text{\textit{CHAIR(R)}} = 1 - \frac{len(R^{'}_{obj} \cap A_{obj})}{len(R^{'}_{obj})}.
          \end{equation}
          }
          where $R'_{obj}$ represents the set of objects mentioned in the model's response, and $A_{obj}$ denotes the set of objects that actually exist in the image.

          Hal score: Measures the proportion of responses containing hallucinations. A response is considered hallucinatory if $\textit{CHAIR(R)} \neq 0$. It is computed as:
          {
          \setlength{\abovedisplayskip}{6pt}
          \setlength{\belowdisplayskip}{6pt}
          \begin{equation}
              \text{\textit{Hal(R)}} =
              \begin{cases}
                  1 & \text{if $\text{\textit{CHAIR(R)}} \neq 0$}, \\
                  0 & \text{otherwise}.
              \end{cases}
          \end{equation}
          }
          Cog score: This metric assesses the alignment between model-generated hallucinations and human cognitive tendencies. It measures the probability of the model generating objects from a predefined set of hallucinatory target objects $H_{obj}$, calculated as:
          {
          \setlength{\abovedisplayskip}{6pt}
          \setlength{\belowdisplayskip}{6pt}
          \begin{equation}
              \text{\textit{Cog(R)}} = \frac{len(R^{'}_{obj} \cap H_{obj})}{len(R^{'}_{obj})}.
          \end{equation}
          }
          In the discriminative component of AMBER, hallucination severity is evaluated based on six factors: object existence, attributes, relationships, state, number, and actions. We report the F1 score to assess the model's performance across these aspects.
          \vspace{+0.5mm}
    \item \textbf{HallusionBench.}
          HallusionBench~\cite{HallusionBench_2023} is a benchmark designed to assess multimodal large language models (MLLMs) in image-context reasoning, specifically focusing on hallucination and illusion phenomena. By incorporating a carefully curated set of challenging reasoning tasks, HallusionBench enables a systematic evaluation of both language-based hallucinations and vision-driven illusions. To quantify model performance, we report the overall accuracy across all questions, covering both straightforward and complex cases.
    \item \textbf{VQAv2.}
          VQAv2~\cite{VQA_v2_2017} is a widely used general visual question answering benchmark that enhances dataset balance by collecting complementary images for each question.
    \item \textbf{TextVQA.}
          TextVQA~\cite{TextVQA_2019} is a benchmark designed for text-rich visual question answering, requiring models to not only recognize textual content within images but also reason about the extracted information. This task evaluates a model's ability to accurately identify text characters while effectively handling the inherent noise present in OCR-generated outputs.
    \item \textbf{ScienceQA.}
          ScienceQA~\cite{ScienceQA_2022} is a multiple-choice benchmark designed to evaluate zero-shot generalization in scientific question answering. It features multimodal questions covering a diverse range of science topics, with annotated answers supported by corresponding lectures and explanations. These annotations provide general external knowledge and specific reasoning for deriving the correct answer. In our study, we conduct experiments on the image subset of ScienceQA to assess model performance in multimodal scientific reasoning.
    \item \textbf{MM-Vet.}
          MM-Vet~\cite{MM_Vet_2024} is a comprehensive benchmark designed to assess a model's ability to engage in visual conversations across diverse tasks. It evaluates response \textbf{correctness} and \textbf{helpfulness} through GPT-4~\cite{GPT4_2023} scoring. The dataset includes a wide range of image types, such as real-world scenes, artworks, statistical graphs, and memes, paired with open-ended questions that require multimodal reasoning. MM-Vet focuses on six core evaluation capabilities: recognition, knowledge, optical character recognition (OCR), spatial awareness, language generation, and math.
\end{itemize}

\subsection{Evaluation Counterparts}
\label{subsec:evaluation_counterparts}

We compare our SENTINEL approach with various methods designed to mitigate hallucinations in MLLMs, all of which are trained on or applied to LLaVA-v1.5~\cite{LLaVA_v1_5_2024} to ensure fairness.

\begin{itemize}[leftmargin=*]
    \vspace{+0.5mm}
    \item \textbf{VCD.}
          VCD~\cite{VCD_2023} is a training-free method designed to mitigate hallucinations in vision-language models by enhancing their focus on image content. It achieves this by contrasting output distributions derived from both original and distorted visual inputs. This contrastive approach helps the model better align its responses with actual image content rather than relying on spurious correlations. The computational cost of a single inference step using VCD is approximately twice that of standard greedy decoding.
    \item \textbf{OPERA.}
          OPERA~\cite{OPERA_2024} addresses hallucination in multimodal language models through two strategies: Over-Trust Penalty and Retrospection-Allocation. The Over-Trust Penalty reduces overconfidence by adjusting model logits during beam search, while Retrospection-Allocation revisits previously generated tokens to correct potential errors, improving response accuracy.
    \item \textbf{DoLa.}
          DoLa~\cite{DoLa_2024} enhances factual accuracy by leveraging contrastive decoding across different model layers. This approach effectively reduces the generation of incorrect facts and consistently improves truthfulness in model responses.
    \item \textbf{EFUF.}
          EFUF~\cite{EFUF_2024} mitigates hallucinations without requiring paired data by employing gradient ascent and three specialized loss functions. It applies gradient descent when encountering real objects and gradient ascent when detecting hallucinated ones, effectively refining the model's output.
    \item \textbf{HA-DPO.}
          HA-DPO~\cite{HA_DPO_2023} formulates hallucination mitigation as a preference selection task, training the model to prefer non-hallucinated responses when given two outputs for the same image. To ensure training stability, it incorporates a causal language modeling objective into the DPO loss. Additionally, both positive and negative samples are rewritten in GPT-4's style to maintain stylistic consistency.
    \item \textbf{POVID.}
          POVID~\cite{POVID_2024} highlights the role of inferior responses in training and enhances them by modifying images and introducing extra hallucinations via GPT-4V~\cite{OpenAI_GPT4V_2023}. The approach then fine-tunes LLaVA-1.5-7B using a set of 17K preference data.
          \vspace{+0.5mm}
    \item \textbf{RLAIF-V.}
          RLAIF-V~\cite{RLAIF_V_2024} employs a ``Feedback From Peer'' strategy, where the overall response score is derived by aggregating scores from decomposed sub-responses, reducing reliance on costly, ultra-large proprietary models like GPT4. The model is trained using an iterative alignment approach, conducting DPO training over four iterations, with each iteration consisting of four epochs.
          \vspace{+0.5mm}
    \item \textbf{TPO.}
          TPO~\cite{Topic_level_SC_2024} is a self-correction approach that enables the model to mitigate its hallucinations at the topic level. Using a deconfounded strategy, it replaces each topic in the response with either the best or worst alternatives generated by the model. This process creates more contrasting pairwise preference feedback, improving the quality of feedback without requiring human intervention or proprietary models.
    \item \textbf{HSA-DPO.}
          HSA-DPO~\cite{HSA_DPO_2024} first trains a hallucination detection model using datasets constructed by GPT-4V~\cite{GPT4_2023}. This model is then leveraged in a detect-then-rewrite pipeline to generate 6K preference data for training. Finally, MLLMs are aligned using the proposed hallucination severity-aware DPO method.
\end{itemize}

\subsection{Evaluation settings}
\label{subsec:detailed_evaluation_settings}

Our overall evaluation setup strictly follows the guidelines provided by LLaVA-v1.5~\cite{LLaVA_v1_5_2024}, with certain hyperparameter settings detailed in~\cref{tab:eval_settings}.

\subsection{Evaluation Results}
\label{subsec:detailed_evaluation_results}

\mypara{Detailed results of MM-Vet.}
We present the detailed results of the MM-Vet~\cite{MM_Vet_2024} benchmark in~\cref{tab:mmvet_details}. The results indicate that, compared to existing methods, our approach achieves the most significant improvement on the 7B model, with an increase of 1.6 points. Notably, for the 13B model, while other methods exhibit varying degrees of performance degradation, our method continues to yield improvements. This demonstrates the effectiveness of our approach in enhancing both the correctness and helpfulness of model responses.

\begin{table}[t!]
    \centering
    \resizebox{\columnwidth}{!} {
        \begin{tabular}{l|cc}
            \toprule
            \textbf{Method} & \textbf{Parameters}                          & \textbf{Value}      \\
            \midrule
            \multirow{3}{*}{VCD~\cite{VCD_2023}}
                            & Amplification Factor $\alpha$                & 1.0                 \\
                            & Adaptive Plausibility Threshold              & 0.1                 \\
                            & Diffusion Noise Step                         & 500                 \\
            \midrule
            \multirow{3}{*}{DoLa~\cite{DoLa_2024}}
                            & Repetition Penalty $\theta$                  & 1.2                 \\
                            & Adaptive Plausibility Threshold $\beta$      & 0.1                 \\
                            & Pre-mature Layers                            & $[0, 2 \cdots, 32]$ \\
            \midrule
            \multirow{4}{*}{OPERA~\cite{OPERA_2024}}
                            & Self-attention Weights Scale Factor $\theta$ & 50                  \\
                            & Attending Retrospection Threshold            & 15                  \\
                            & Beam Size                                    & 3                   \\
                            & Penalty Weights                              & 1                   \\
            \bottomrule
        \end{tabular}
    }
    \vspace{-10pt}
    \caption{Evaluation hyperparameters of decode-based methods.}
    \vspace{-5pt}
    \label{tab:eval_settings}
\end{table}
\begin{table}[t!]
    \centering
    \resizebox{\columnwidth}{!} {%
        \begin{tabular}{l ccccccc}
            \toprule
            \multicolumn{1}{l}{\textbf{Method}}           &
            \multicolumn{1}{c}{Rec}                       &
            \multicolumn{1}{c}{OCR}                       &
            \multicolumn{1}{c}{Know}                      &
            \multicolumn{1}{c}{Gen}                       &
            \multicolumn{1}{c}{Spat}                      &
            \multicolumn{1}{c}{Math}                      &
            \multicolumn{1}{c}{\textbf{Overall}}                                                                                 \\
            \midrule
            LLaVA-v1.5-7B~\cite{LLaVA_v1_5_2024}          & 35.9 & 23.3 & 17.1 & 22.0 & 25.9 & 11.5 & 31.0\pmval{0.2}            \\
            VCD~\cite{VCD_2023}                           & 34.5 & 21.9 & 18.3 & 20.6 & 24.8 & 3.8  & 29.8\downred{1.2}          \\
            OPERA~\cite{OPERA_2024}                       & 34.9 & 21.6 & 18.7 & 21.1 & 25.7 & 7.7  & 30.3\downred{0.7}          \\
            DoLa~\cite{DoLa_2024}                         & 36.1 & 21.3 & 19.4 & 20.9 & 26.9 & 7.7  & 30.8\downred{0.2}          \\
            EFUF~\cite{EFUF_2024}                         & 36.5 & 21.4 & 17.1 & 19.5 & 27.9 & 7.7  & 31.2\upgreen{0.2}          \\
            HA-DPO~\cite{HA_DPO_2023}                     & 35.5 & 22.1 & 18.3 & 21.9 & 26.3 & 7.7  & 30.6\downred{0.4}          \\
            POVID\textsuperscript{\dag}~\cite{POVID_2024} & -    & -    & -    & -    & -    & -    & 31.8\upgreen{0.8}          \\
            RLAIF-V~\cite{RLAIF_V_2024}                   & 34.4 & 23.4 & 18.7 & 23.7 & 27.7 & 7.3  & 29.9\downred{1.1}          \\
            TPO~\cite{Topic_level_SC_2024}                & 31.8 & 15.4 & 16.7 & 19.6 & 22.1 & 7.7  & 25.7\downred{5.3}          \\
            Ours                                          & 37.7 & 23.1 & 22.7 & 25.6 & 26.8 & 7.7  & \ul{32.6}\upgreen{1.6}     \\
            Ours + HA-DPO~\cite{HA_DPO_2023}              & 38.4 & 25.0 & 21.2 & 23.7 & 29.3 & 7.7  & \textbf{33.5}\upgreen{2.5} \\
            \midrule
            LLaVA-v1.5-13B~\cite{LLaVA_v1_5_2024}         & 39.7 & 28.8 & 23.2 & 24.2 & 34.5 & 11.5 & \ul{36.0}                  \\
            VCD~\cite{VCD_2023}                           & 38.7 & 24.4 & 22.4 & 26.4 & 30.1 & 7.7  & 33.7\downred{2.3}          \\
            vanilla-DPO~\cite{HSA_DPO_2024}               & 38.4 & 29.7 & 17.9 & 21.0 & 35.6 & 11.5 & 35.0\downred{1.0}          \\
            HSA-DPO~\cite{HSA_DPO_2024}                   & 35.9 & 28.4 & 16.4 & 18.9 & 34.5 & 15.0 & 33.7\downred{2.3}          \\
            Ours                                          & 38.9 & 30.2 & 22.6 & 23.1 & 32.7 & 15.0 & \textbf{36.2}\upgreen{0.2} \\
            \bottomrule
        \end{tabular}
    }
    \vspace{-10pt}
    \caption{\textbf{Full evaluation results of MM-Vet benchmark.} \dag indicates that the results are from~\cite{POVID_2024}.}
    \vspace{-10pt}
    \label{tab:mmvet_details}
\end{table}

\mypara{Detailed results of AMBER.} We present the detailed results of the discriminative part of the AMBER~\cite{MM_Vet_2024} benchmark in~\cref{tab:amber_dis_details}. The results show that some of the existing methods may experience a decline in performance across certain specific hallucination categories. In contrast, our approach demonstrates improvements in every specific hallucination category, regardless of whether the 7B or 13B model is used. Notably, for the Existence hallucination type, our method improves the 7B model by \textbf{6.3} and the 13B model by \textbf{7.6} compared to the baseline.

\begin{table}[ht!]
    \centering
    \resizebox{\columnwidth}{!} {%
        \begin{tabular}{l ccccccc}
            \toprule
            \multicolumn{1}{l}{\textbf{Method}} &
            \multicolumn{1}{c}{Existence}       &
            \multicolumn{1}{c}{Attribute}       &
            \multicolumn{1}{c}{State}           &
            \multicolumn{1}{c}{Number}          &
            \multicolumn{1}{c}{Action}          &
            \multicolumn{1}{c}{Relation}        &
            \multicolumn{1}{c}{\textbf{Overall}}
            \\
            \midrule
            LLaVA-v1.5-7B                       & 82.4              & 64.0              & 57.7              & 69.9              & 81.1              & 67.7               & 74.1                       \\
            VCD~\cite{VCD_2023}                 & 81.1\downred{1.3} & 65.6              & 61.8              & 67.5\downred{2.4} & 80.9\downred{0.2} & 66.7\downred{1.0}  & 73.9\downred{0.2}          \\
            DoLa~\cite{DoLa_2024}               & 87.6              & 67.5              & 62.1              & 72.8              & 82.4              & 56.3\downred{11.4} & 77.8                       \\
            EFUF~\cite{EFUF_2024}               & 85.3              & 61.2\downred{2.8} & 55.5\downred{2.2} & 65.1\downred{4.8} & 80.4\downred{0.7} & 67.4\downred{0.3}  & 75.0                       \\
            HA-DPO~\cite{HA_DPO_2023}           & 88.2              & 66.1              & 56.5\downred{1.2} & 78.5              & 82.3              & 68.7               & \ul{78.0}                  \\
            Ours                                & 88.7\upgreen{6.3} & 67.6\upgreen{3.6} & 61.3\upgreen{3.6} & 74.8\upgreen{4.9} & 82.1\upgreen{1.0} & 70.6\upgreen{2.9}  & \textbf{79.3}\upgreen{5.2} \\
            \midrule
            LLaVA-v1.5-13B                      & 78.5              & 70.0              & 66.0              & 74.2              & 82.2              & 44.9               & 73.1                       \\
            VCD~\cite{VCD_2023}                 & 78.5              & 71.7              & 69.0              & 73.6\downred{0.6} & 81.6\downred{0.6} & 45.6               & \ul{73.8}                  \\
            Ours                                & 86.1\upgreen{7.6} & 72.6\upgreen{2.6} & 66.6\upgreen{0.6} & 81.6\upgreen{7.4} & 82.6\upgreen{0.4} & 51.5\upgreen{6.6}  & \textbf{78.7}\upgreen{5.6} \\
            \bottomrule
        \end{tabular}
    }
    \vspace{-10pt}
    \caption{
        \textbf{Full evaluation results of AMBER's descriminative part.}
        We report F1 scores for each category and the overall score.
    }
    \vspace{-10pt}
    \label{tab:amber_dis_details}
\end{table}

\subsection{Details of Ablation Study}
\label{subsec:ablation_study_details}

In this section, we provide more specific details of the ablation studies to validate the effectiveness of our method.

\mypara{Effect of style consistency.}
Many preference training methods adopt rewriting techniques to construct non-hallucinated training samples~\cite{HA_DPO_2023, HSA_DPO_2024, Topic_level_SC_2024}. To validate the negative impact of rewritten training data on the model's generalization performance, we follow the approach of HA-DPO~\cite{HA_DPO_2023} and design prompts to instruct GPT to rewrite the preference training samples. Specifically, we prompt GPT-4\cite{GPT4_2023} to rewrite $\vec{y}_{w}$ and $\vec{y}_{l}$ in a different style while ensuring coherence with the given context. The prompt template is provided in~\cref{tab:prompt_rewrite}, and the results are presented in the main paper Tab.~2.

To evaluate how our in-domain training data affects the model's linguistic qualities, we adopt the approach from \cite{HALVA_2024} and use GPT-4o-mini~\cite{GPT_4o_2024} as a judge. Responses are rated on a scale of 0 to 10 across four aspects: grammatical correctness, fluency, detailedness, and choice of words. We assess the model's performance on 300 image description tasks from Object HalBench~\cite{Obj_HalBench_2018}. The evaluation prompt is shown in~\cref{tab:prompt_linguistic}. As demonstrated in~\cref{tab:linguistic_qualities}, our training not only preserves the model's linguistic capabilities but also improves the detailedness of the descriptions.

\begin{table}[h]
    \centering
    \resizebox{\columnwidth}{!} {%
        \begin{tabular}{l ccccc}
            \toprule
            \multirow{1}{*}{\textbf{Method}}            &
            \multicolumn{1}{c}{Grammatical Correctness} &
            \multicolumn{1}{c}{Fluency}                 &
            \multicolumn{1}{c}{Detailedness}            &
            \multicolumn{1}{c}{Choice Of Words}
            \\
            \midrule
            LLaVA-v1.5-7B                               & 9.92               & 9.28               & 8.21               & 8.94               \\
            SENTINEL                                    & 9.97\upgreen{0.05} & 9.53\upgreen{0.25} & 8.32\upgreen{0.11} & 8.97\upgreen{0.03} \\
            \midrule
            LLaVA-v1.5-13B                              & 9.95               & 9.44               & 8.29               & 8.95               \\
            SENTINEL                                    & 9.98\upgreen{0.03} & 9.60\upgreen{0.16} & 8.40\upgreen{0.11} & 8.98\upgreen{0.03} \\
            \bottomrule
        \end{tabular}

    }
    \vspace{-10pt}
    \caption{
        \textbf{Language quality evaluation results.}
        Our in-domain training data preserves the model's language quality in image detail description tasks while improving the level of detailedness.
    }
    \vspace{-5pt}
    \label{tab:linguistic_qualities}
\end{table}

\begin{figure}[htbp]
    \centering
    \begin{subfigure}[b]{\linewidth}
        \includegraphics[width=\linewidth]{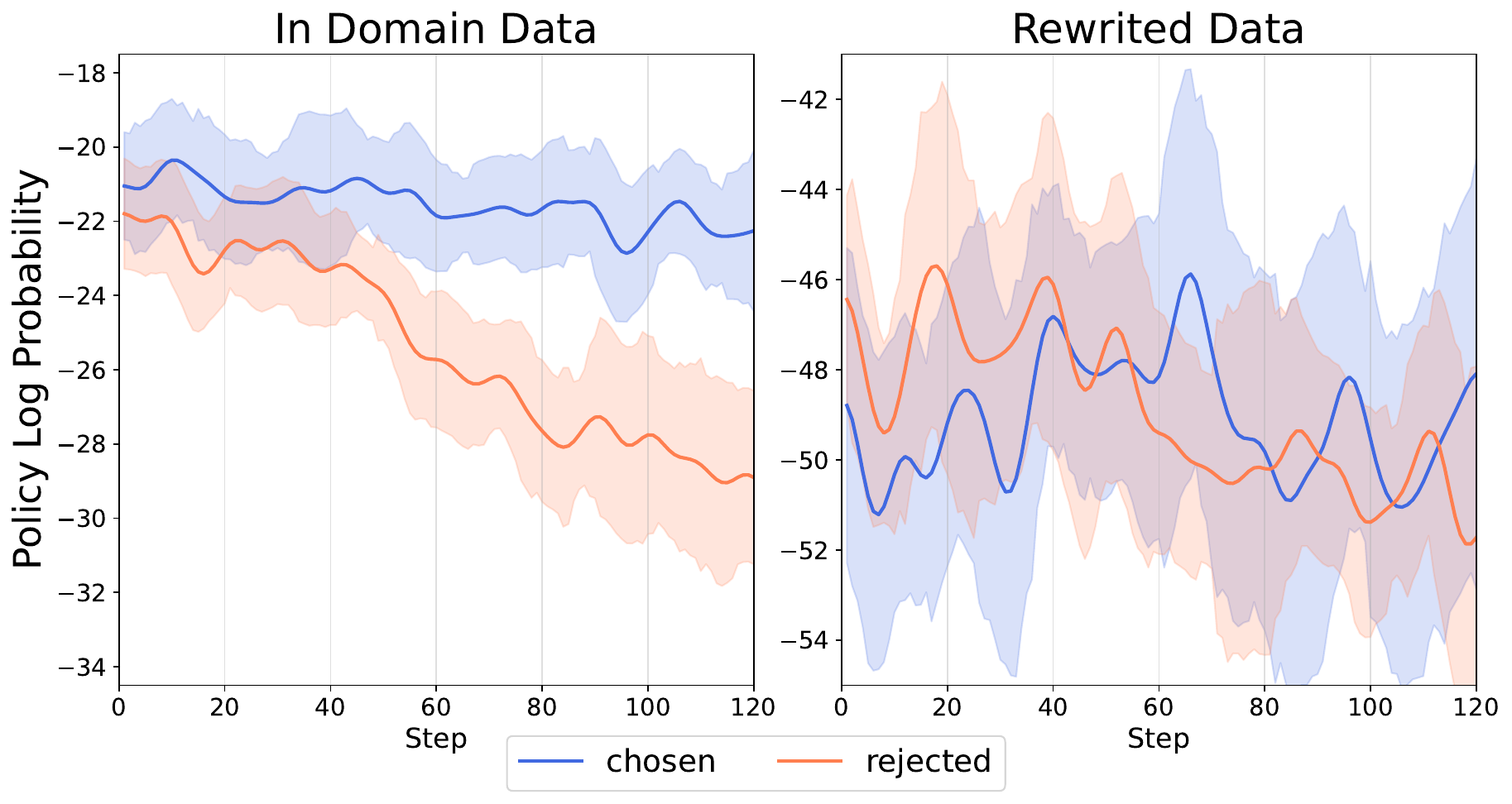}
        \vspace{-15pt}
        \caption{Training log probabilities (logps) comparison}
        \label{fig:logps_rewrited}
    \end{subfigure}
    \vspace{12pt}
    \begin{subfigure}[b]{\linewidth}
        \includegraphics[width=\linewidth]{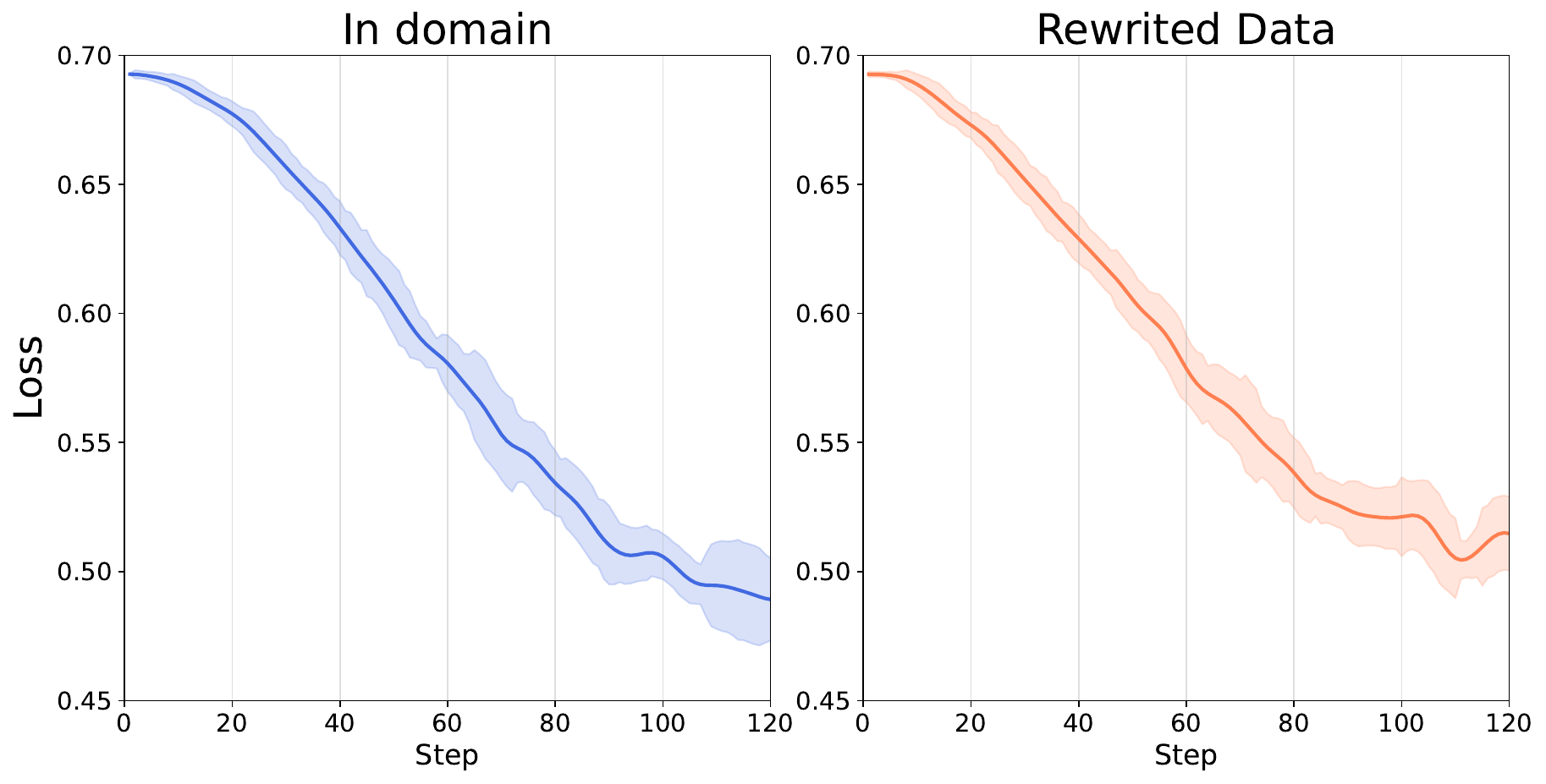}
        \vspace{-16pt}
        \caption{Training loss comparison}
        \label{fig:loss_rewrited}
    \end{subfigure}
    \vspace{-32pt}
    \caption{\textbf{Impact of rewriting on the training process.} Training with rewritten data fails to achieve the same level of convergence, resulting in higher final loss and weaker differentiation between positive and negative samples, demonstrating the necessity of in-domain training data.}
    \label{fig:logps_loss_rewrited}
    \vspace{-14pt}
\end{figure}

To further investigate the impact of rewritten data on training, we analyze the log probabilities (logps) and loss trends of the policy model when trained with in-domain data versus rewritten data, as shown in~\cref{fig:logps_loss_rewrited}.  Our observations indicate that the rewritten data, due to its deviation from the model's original output style and linguistic domain, significantly lowers the logps of both positive and negative samples. Additionally, the rewriting process obscures the fundamental distinction between positive and negative samples (i.e., whether hallucinations are present), thereby weakening the model's ability to distinguish between them and diminishing the effectiveness of the training signal. As a result, models trained on in-domain data converge to a lower loss and achieve superior differentiation between positive and negative samples, whereas training with rewritten data fails to provide comparable improvements.

\mypara{Effect of data scaling up.}
Since our proposed SENTINEL does not rely on ultra-large proprietary models~\cite{AMP_2024, FGAIF_2024, HA_DPO_2023, HSA_DPO_2024, POVID_2024, Woodpecker_2023} or human annotators~\cite{RLHF_V_2024, M_HalDetect_2024} for preference learning dataset construction, it can efficiently collect more training data. As shown in~\cref{tab:data_scale_up}, although RLHF-V~\cite{RLHF_V_2024} leverages high-quality human-annotated training data to achieve a lower hallucination rate with fewer training samples, their high cost limits the scalability of the training data. Our method enables cost-effective scaling up, leading to improved model performance.

\begin{table}[ht]
    \centering
    \resizebox{\columnwidth}{!} {%
        \begin{tabular}{l ccccc}
            \toprule
            \multirow{2}{*}{\vspace{-2mm}\textbf{Method}}                        &
            \multirow{2}{*}{\vspace{-2mm}\textbf{Data Scale}}                    &
            \multicolumn{2}{c}{\textbf{Object HalBench}\cite{Obj_HalBench_2018}} &
            \multicolumn{2}{c}{\textbf{AMBER}\cite{AMBER_2023}}
            \\
            \cmidrule(lr){3-4} \cmidrule(lr){5-6}                                &      & Resp. $\downarrow$ & Ment. $\downarrow$ & Acc $\uparrow$ & F1 $\uparrow$ \\
            \midrule
            LLaVA-v1.5-7B                                                        & -    & 52.7               & 28.0               & 71.5           & 74.1          \\
            RLHF-V~\cite{RLHF_V_2024}                                            & 1.4K & \ul{12.2}          & \ul{7.5}           & \ul{72.6}      & \ul{75.0}     \\
            \midrule
            SENTINAL                                                             & 2.0K & 39.0               & 20.0               & 72.2           & 74.9          \\
            SENTINAL                                                             & 8.6K & \textbf{4.3}       & \textbf{2.6}       & \textbf{76.1}  & \textbf{79.3} \\
            \bottomrule
        \end{tabular}
    }
    \vspace{-10pt}
    \caption{
        \textbf{Impact of training data quantity.}
        The results show that SENTINEL demonstrates better efficiency and scalability.
    }
    \vspace{-10pt}
    \label{tab:data_scale_up}
\end{table}

\mypara{Complement with existing preference learning methods.}
HA-DPO~\cite{HA_DPO_2023} employs a GPT-4~\cite{GPT4_2023}-based rewriting approach to modify both positive and negative samples in the preference training data, ensuring stylistic consistency between them. However, this rewriting process introduces stylistic discrepancies between the training data and the target model’s original outputs, potentially affecting its generalization ability.

To assess the effectiveness of in-domain preference learning data, we augment the HA-DPO~\cite{HA_DPO_2023} training dataset (approximately 4.4K samples) with a subset of our constructed dataset (6K samples from the full 8.6K) and train \href{https://huggingface.co/liuhaotian/llava-v1.5-7b}{LLaVA-v1.5-7B} under the same training settings as HA-DPO. As shown in the main paper Tab.~5, integrating even a partial set of our training data significantly reduces hallucinations while enhancing the model's overall performance. These results further demonstrate that our sentence-level preference training approach is complementary to existing sample-level preference learning methods.

\begin{table}[ht!]
    \centering
    \begin{tcolorbox} [colback=gray!5, boxsep=2pt,left=5pt, right=5pt, top=3pt, bottom=3pt]
        \scriptsize
        Following is a detailed image description.\\
        Your task is to assess the response on the following criteria:\\
        1. Grammatical Correctness: Analyze the response for grammar, punctuation, and syntax accuracy.\\
        2. Fluency: Evaluate whether the response flows smoothly, reads naturally, and maintains coherence throughout.\\
        3. Detailedness: Check if the response provides sufficient and relevant detail to address the topic comprehensively, without redundancy or unnecessary information.\\
        4. Choice of Words: Assess if the words used are appropriate, varied, and effectively convey the intended message.
        Rate each criterion on a scale from 0 to 10, where 0 indicates poor quality and 10 signifies an excellent response.\\

        Here is the image description to evaluate:\\

        \textcolor{blue}{\{description\}}\\

        Your response should be in this format:\\
        Grammatical Correctness: SCORE\\
        Fluency: SCORE\\
        Detailedness: SCORE\\
        Choice of Words: SCORE
    \end{tcolorbox}
    \vspace{-0.5cm}
    \caption{
        \textbf{Prompts for linguistic quality evaluation.}
        Responses are rated on a scale of 0 to 10 across four aspects: grammatical correctness, fluency, detailedness, and choice of words.
    }
    \vspace{-10px}
    \label{tab:prompt_linguistic}
\end{table}

\begin{table}[ht!]
  \centering
  \begin{tcolorbox} [colback=gray!5, boxsep=2pt,left=5pt, right=5pt, top=3pt, bottom=3pt]
    \scriptsize
    \textbf{Rewrite training data} \vspace{0.1cm} \\
    Please help me rewrite the given sentences in a style different from the original.\\
    You will be provided with three parts: ``context'' refers to the previously generated sentences, and ``option one'' and ``option two'' represent two choices for the sentence that follows the context.\\
    Your goal is to make the new versions from the original while preserving all details and information. \\
    Avoid adding any new information or changing the original meaning.\\
    Please rewrite the two options that differ in tone, structure, word choice, and phrasing compared to the original, while ensuring coherence and natural flow with the given context.\\

    The format of your output should be: \\
    Option one: ... \\
    Option two: ... \\

    The sentences are: \\
    Context: \textcolor{blue}{\{context\}}\\
    Option one: \textcolor{blue}{\{y\_win\}}\\
    Option two: \textcolor{blue}{\{y\_lose\}}
  \end{tcolorbox}
  \vspace{-0.5cm}
  \caption{
    \textbf{Prompts for rewriting.}
    We prompt GPT-4~\cite{GPT4_2023} to rewrite $\vec{y}^{+}_{w}$ and $\vec{y}_{l}$ in a style different from the original while ensuring coherence with the given context $\vec{c}$ to show the effect of rewriting on the model's generalization performance.
  }
  \vspace{-15px}
  \label{tab:prompt_rewrite}
\end{table}

\section{SENTINEL with Other Baselines}
\label{sec:supp_sentinel_other_baselines}

\begin{table*}[t!]
    \centering
    \resizebox{\textwidth}{!}{%
        \begin{tabular}{ll c c |c c c c c}
            \toprule
            \multirow{3}{*}{\vspace{-2mm}\textbf{Model}}                           &
            \multicolumn{3}{c}{\textbf{Hallucination benchmarks}}                  &
            \multicolumn{4}{c}{\textbf{General benchmarks}}
            \\
                                                                                   &
            \multicolumn{2}{c}{\textbf{Object HalBench}~\cite{Obj_HalBench_2018}}  &
            \multicolumn{1}{c}{\textbf{HallusionBench}~\cite{HallusionBench_2023}} &
            \multicolumn{1}{c}{\textbf{VQAv2}~\cite{VQA_v2_2017}}                  &
            \multicolumn{1}{c}{\textbf{TextVQA}~\cite{TextVQA_2019}}               &
            \multicolumn{1}{c}{\textbf{ScienceQA}~\cite{ScienceQA_2022}}           &
            \multicolumn{1}{c}{\textbf{MM-Vet}~\cite{MM_Vet_2024}}
            \\
            \cmidrule(lr){2-3}\cmidrule(lr){4-4}\cmidrule(lr){5-5}
            \cmidrule(lr){6-6}\cmidrule(lr){7-7}\cmidrule(lr){8-8}
                                                                                   &
            Resp.\,$\downarrow$                                                    &
            Ment.\,$\downarrow$                                                    &
            Question Acc.\,$\uparrow$                                              &
            Acc.\,$\uparrow$                                                       &
            Acc.\,$\uparrow$                                                       &
            Image Acc.\,$\uparrow$                                                 &
            Overall\,$\uparrow$
            \\
            \midrule
            \multirow{1}{*}{LLaVA-v1.6-vicuna-7B}                                  & 15.3$\to$5.0 & 10.1$\to$3.4 & 36.73$\to$37.80 & 81.5$\to$81.5 & 59.4$\to$59.4 & 74.3$\to$74.2 & 40.9$\to$45.4 \\

            \midrule
            \multirow{1}{*}{LLaVA-v1.6-vicuna-13B}                                 & 13.7$\to$4.0 & 7.7$\to$2.6  & 41.10$\to$41.36 & 82.2$\to$82.2 & 63.6$\to$63.5 & 77.7$\to$78.0 & 47.8$\to$48.5 \\
            \midrule
            \multirow{1}{*}{Qwen2-VL-2B-Instruct}                                  & 15.3$\to$2.3 & 8.6$\to$1.7  & 41.28$\to$42.16 & 81.5$\to$81.5 & 78.3$\to$78.5 & 76.9$\to$77.4 & 49.4$\to$49.8 \\
            \midrule
            \multirow{1}{*}{Qwen2-VL-7B-Instruct}                                  & 14.3$\to$4.8 & 8.5$\to$4.0  & 51.55$\to$53.41 & 83.7$\to$83.8 & 82.2$\to$82.2 & 85.7$\to$86.9 & 62.7$\to$62.8 \\
            \midrule
            \multirow{1}{*}{Qwen2.5-VL-7B-Instruct}                                & 15.0$\to$4.7 & 9.2$\to$2.8  & 52.00$\to$52.08 & 84.0$\to$84.0 & 77.7$\to$77.7 & 88.6$\to$88.5 & 72.0$\to$72.2 \\
            \bottomrule
        \end{tabular}
    }
    \vspace{-2mm}
    \caption{
        \textbf{Comparison of hallucination mitigation methods with other baseline models: effectiveness and general capabilities (baseline $\!\to\!$ SENTINEL).}
        This evaluation highlights the best and second-best results in \textbf{bold} and \ul{underlined}, respectively. All comparisons are performed under identical model size constraints. ``Resp.'' and ``Ment.'' denote response-level and mention-level hallucination rates, while ``Hal.'' and ``Cog.'' represent the Hallucination Score and Cognitive Score, respectively.
    }
    \label{tab:sentinel_other_baselines}
\end{table*}

In this section, we explore the effectiveness of our SENTINEL approach when applied to other baselines, specifically \href{https://huggingface.co/collections/llava-hf/llava-next-65f75c4afac77fd37dbbe6cf}{LLaVA-v1.6}~\cite{LLaVA_NeXT_2024}, \href{https://huggingface.co/collections/Qwen/qwen2-vl-66cee7455501d7126940800d}{Qwen2-VL}~\cite{Qwen2_5_VL_2024} and \href{https://huggingface.co/collections/Qwen/qwen25-vl-6795ffac22b334a837c0f9a5}{Qwen2.5-VL}~\cite{Qwen2_5_VL_2024}. The results are presented in~\cref{tab:sentinel_other_baselines}. The findings indicate that our SENTINEL approach consistently reduces hallucinations across a range of model families and sizes, while preserving or even enhancing overall performance, thereby demonstrating its robustness and effectiveness.

During these experiments, to generate training data for each target model, we simply replace the sampling model within the SENTINEL framework with the corresponding model, thereby demonstrating SENTINEL's model-agnostic design. For training, we employ the widely used \href{https://github.com/hiyouga/LLaMA-Factory}{LLaMA-Factory}~\cite{LlamaFactory_2024} framework to ensure fairness and reproducibility. Evaluation follows the same protocol described above\footnote{For efficiency, in this set of experiments we use the GPT-4o~\cite{GPT_4o_2024} model for HallusionBench~\cite{HallusionBench_2023} evaluation, which makes these results not directly comparable to those reported for the benchmark in the main paper.}. All training data, configuration details, and associated resources will be released publicly.

\section{Related Work}
\label{sec:supp_related_works}
\mypara{Multimodal large language models.}
In recent years, vision-language models (VLMs) have made remarkable progress~\cite{CLIP_2021, shao2024explore, wang2025declip, tian2019learning, liu2024typicalness, yang2024unified}. With the advancement of large language models (LLMs), multimodal large language models (MLLMs) have achieved impressive alignment between visual and textual representations through cross-modal feature integration, marking a crucial milestone toward truly general-purpose AI systems~\cite{Qwen_VL_2023, Qwen2_VL_2024, Qwen2_5_VL_2024, LLaVA_v1_2023, LLaVA_v1_5_2024, LLaVA_NeXT_2024, InstructBLIP_2023, OpenAI_GPT4V_2023, MiniGPT_4_2024, qu2025does, yang2023improved, zhong2024lyra, yang2023lidar, li2023mgm, lai2024lisa, yang2024visionzip}. However, mitigating hallucination and building reliable models for real-world scenarios remain significant challenges.

\mypara{Object Hallucination.}
Object Hallucination (OH) refers to the phenomenon where MLLMs generate text that is semantically coherent but misaligned with the given image~\cite{HallucinationSurvey_2023,HallucinationSurvey_2024,HallucinationSurvey_202402}. Prior studies suggest that this issue may arise during generation due to an over-reliance on linguistic priors and insufficient attention to visual features~\cite{Visual_Perturbation_2022, Counterfactual_VQA_2021, Distinguishing_VQA_2022}. Furthermore, research indicates that hallucination tends to intensify over time~\cite{LURE_2024, FaithScore_2024}.

\mypara{Mitigate OH with improved decoding strategies.}
Several approaches have explored enhanced decoding strategies to mitigate object hallucination. VCD~\cite{VCD_2023} enhances the model's focus on image content during generation by applying contrastive decoding between the original image and a noise-corrupted version. DoLa~\cite{DoLa_2024} improves factual accuracy by leveraging contrastive decoding across layers to better surface factual knowledge and reduce incorrect outputs. OPERA~\cite{OPERA_2024} introduces Over-Trust Penalty and Retrospection-Allocation to address hallucination in multimodal language models. HALC~\cite{HALC_2024} reduces object hallucination through an adaptive focal-contrast decoding approach, incorporating a dynamic auto-focal grounding mechanism for real-time token correction and a refined beam search strategy to effectively suppress hallucinations while maintaining text quality.

\mypara{Mitigate OH by preference learning.}
Preference learning is a powerful paradigm for aligning large language models with human judgments and values. Recently, Direct Preference Optimization (DPO)~\cite{DPO_2023} and its variations~\cite{SimPO_2024, DPO_positive_2024, Uni_DPO_2026} have made preference learning more accessible and easier to integrate. Another line of research on mitigating OH employs preference learning to tackle object hallucination by reformulating it as a preference optimization problem. These approaches construct high-quality, stylistically consistent positive-negative sample pairs to enhance model training. Rewriting is an effective method for obtaining ``non-hallucinated'' training data. HA-DPO~\cite{HA_DPO_2023} utilizes GPT~\cite{GPT4_2023} to directly detect and rewrite the model's original output, ensuring that both positive and negative samples undergo rewriting. HSA-DPO~\cite{HSA_DPO_2024} distills a smaller hallucination detection model from the proprietary model GPT and applies it to detect hallucinations and refine responses through rewriting. In contrast, RLAIF~\cite{RLAIF_V_2024} does not employ rewriting; instead, it constructs datasets using the ``Feedback from Peer'' approach, leveraging open-source models' outputs as feedback. This method directly utilizes the model's full outputs as both positive and negative samples.

\mypara{Eliminating OH via unlearning.}
EFUF~\cite{EFUF_2024} employs an unlearning-based approach to mitigate OH in MLLMs. Specifically, it applies gradient ascent when hallucinated objects are detected during the model's generation, effectively suppressing hallucinations in a targeted and efficient manner. This method enables a precise removal of hallucinated content during captioning without compromising the model's overall performance.

\section{Additional Case Studies}
\label{sec:additional_case_studies}

In this section, we provide more case studies to further illustrate the effectiveness of our proposed SENTINEL in mitigating hallucinations and enhancing its generalization capabilities. To ensure a fair comparison, results for both LLaVA-v1.5 and SENTINEL are obtained using greedy decoding, while GPT-4V refers to gpt-4-vision-preview~\cite{OpenAI_GPT4V_2023}.

\begin{figure*}[t]
    \centering
    \includegraphics[width=0.90\textwidth]{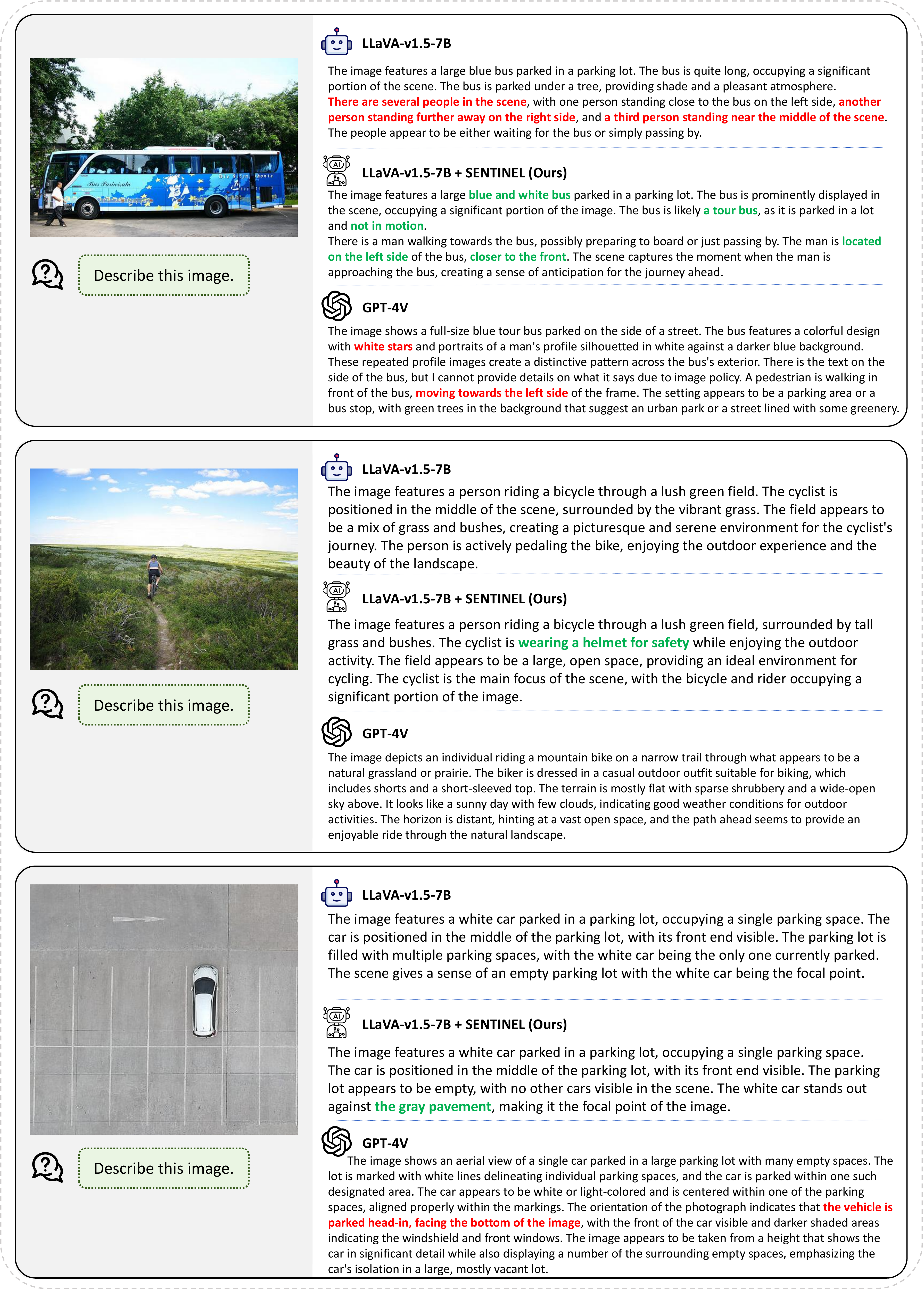}
    \vspace{-5pt}
    \caption{\textbf{Comparing general image description results between SENTINAL and its base model LLaVA-v1.5-7B.} Our method effectively mitigates hallucinations while enhancing the general performance of the base model, providing a more detailed description.}
    \label{fig:case_7B_1}
    \vspace{-10pt}
\end{figure*}
\begin{figure*}[t]
    \centering
    \includegraphics[width=0.90\textwidth]{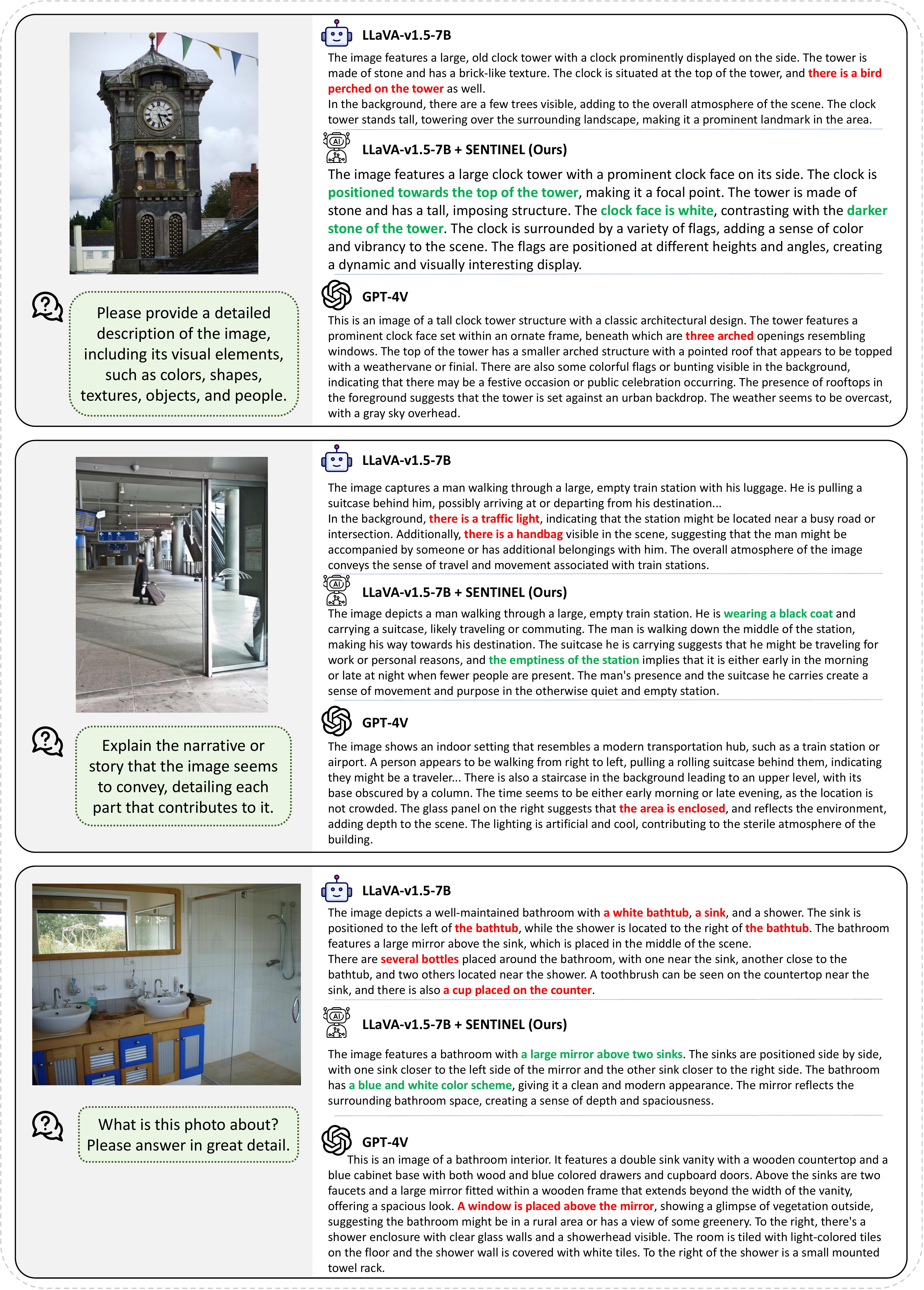}
    \vspace{-5pt}
    \caption{\textbf{Comparing detailed image description results between SENTINAL and its base model LLaVA-v1.5-7B.} Our method effectively mitigates hallucinations while enhancing the general performance of the base model, providing a more detailed description.}
    \label{fig:case_7B_2}
    \vspace{-10pt}
\end{figure*}
\begin{figure*}[t]
    \centering
    \includegraphics[width=0.89\textwidth]{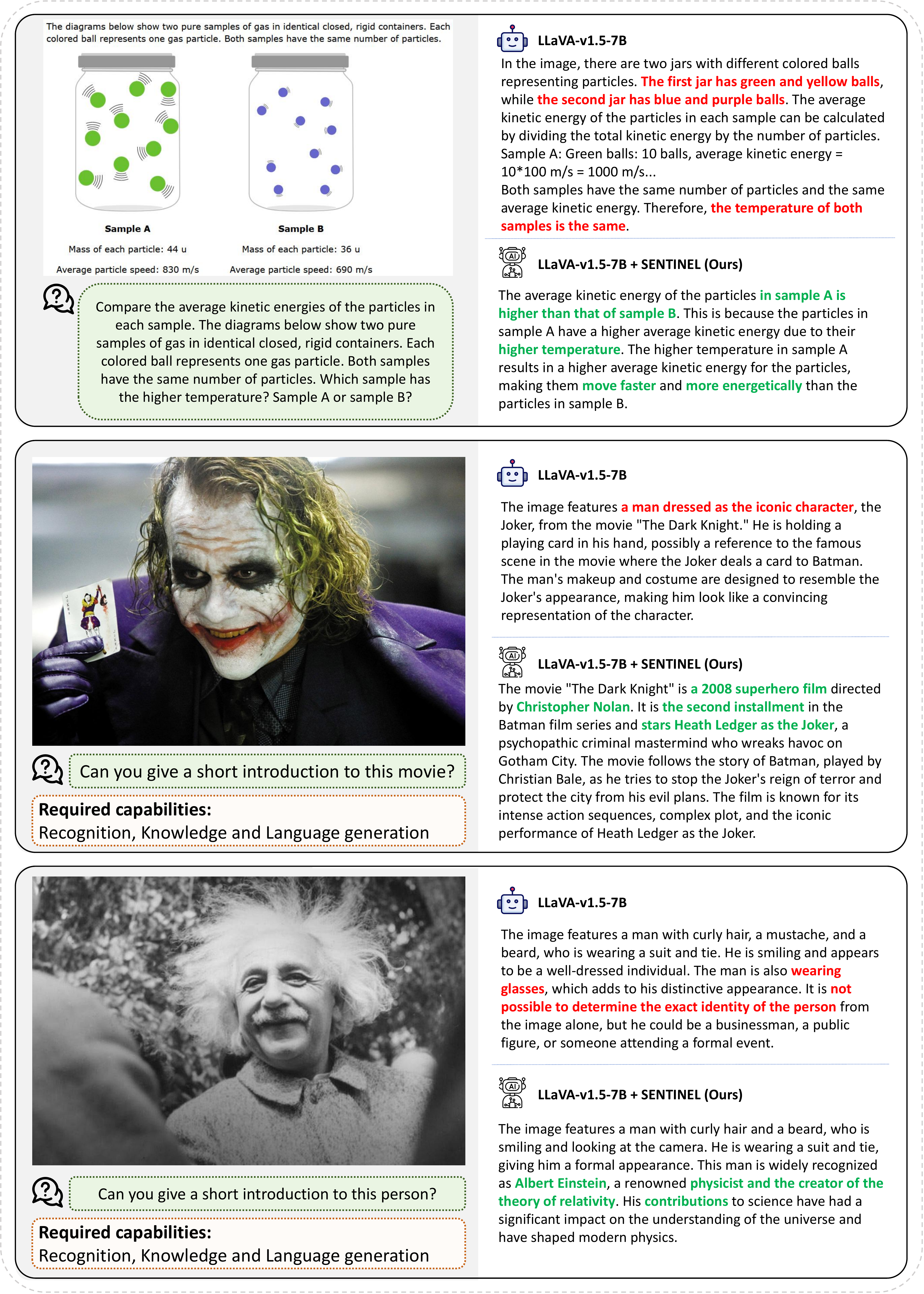}
    \vspace{-5pt}
    \caption{\textbf{Comparing visual question answering results between SENTINAL and LLaVA-v1.5-7B.} Our method effectively mitigates hallucinations while enhancing the general performance of the base model, leading to more accurate and detailed answers.}
    \label{fig:case_7B_3}
    \vspace{-10pt}
\end{figure*}
\begin{figure*}[t]
    \centering
    \includegraphics[width=0.90\textwidth]{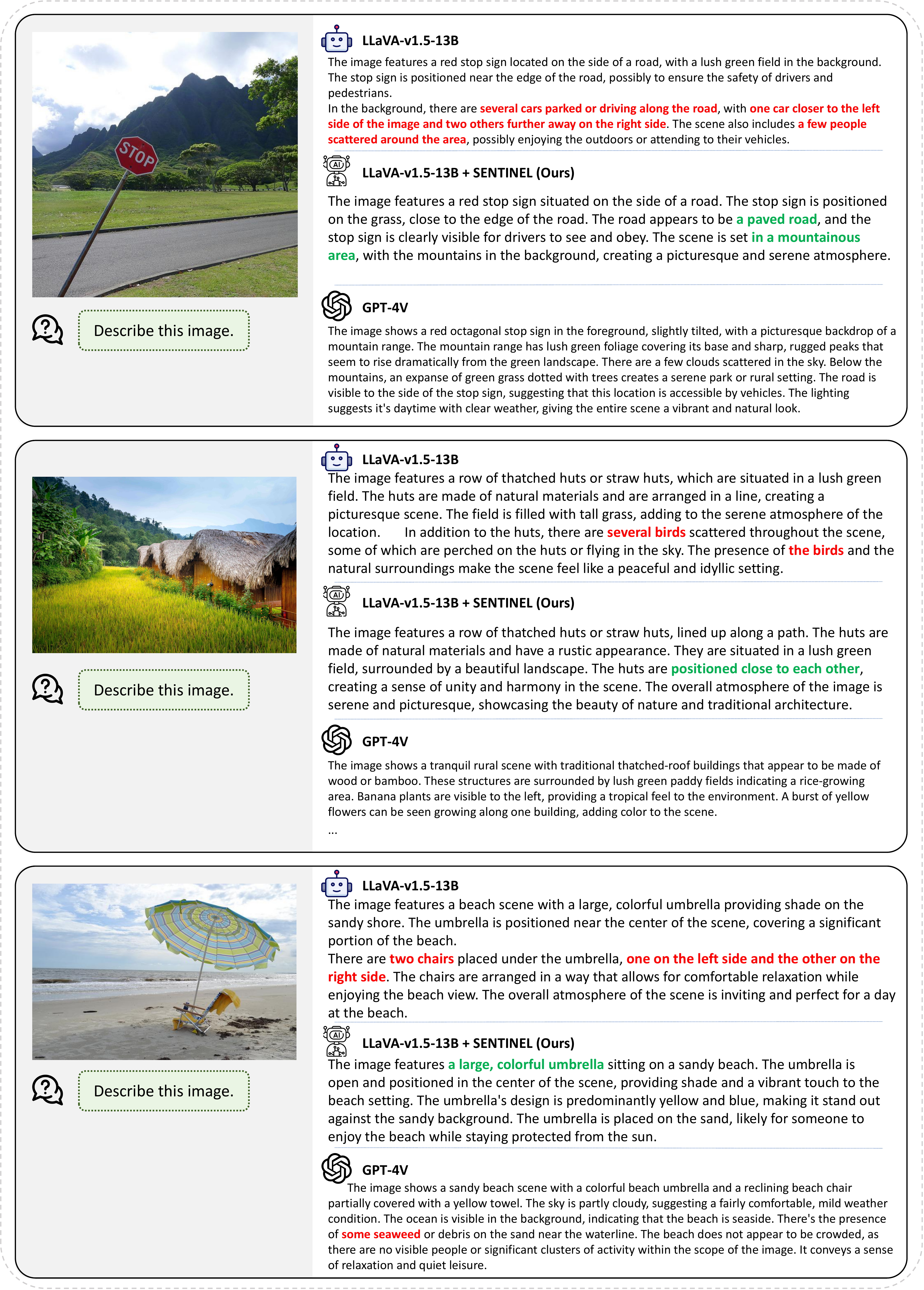}
    \vspace{-5pt}
    \caption{\textbf{Comparing general image descriptions between SENTINAL and its base model LLaVA-v1.5-13B.} Our method effectively mitigates hallucinations while enhancing the general performance of the base model, providing a more detailed description.}
    \label{fig:case_13B_1}
    \vspace{-10pt}
\end{figure*}
\begin{figure*}[t]
    \centering
    \includegraphics[width=0.90\textwidth]{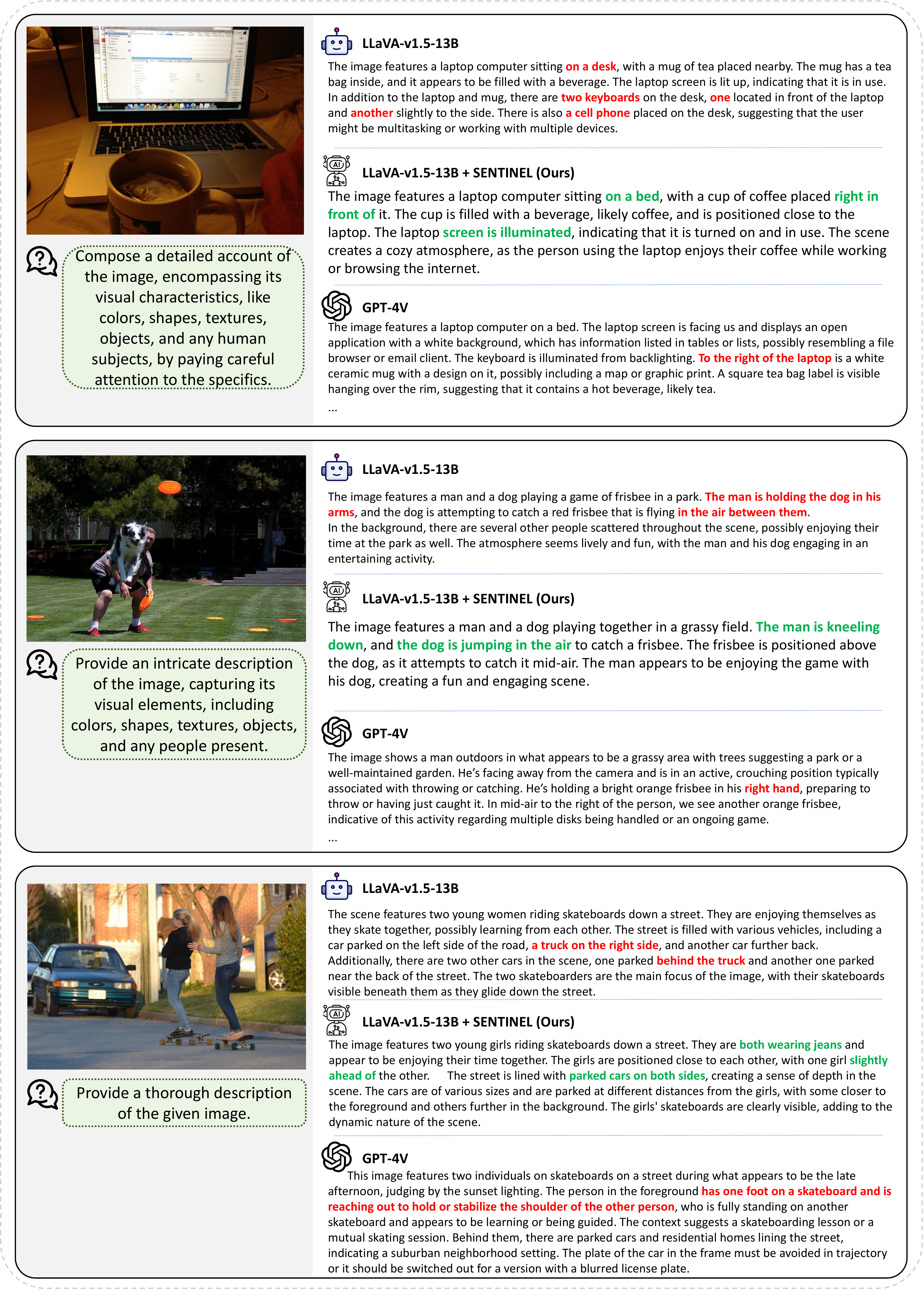}
    \vspace{-5pt}
    \caption{\textbf{Comparing detailed image descriptions between SENTINAL and its base model LLaVA-v1.5-13B.} Our method effectively mitigates hallucinations while enhancing the general performance of the base model, providing a more detailed description.}
    \label{fig:case_13B_2}
    \vspace{-10pt}
\end{figure*}
\begin{figure*}[t]
    \centering
    \includegraphics[width=0.90\textwidth]{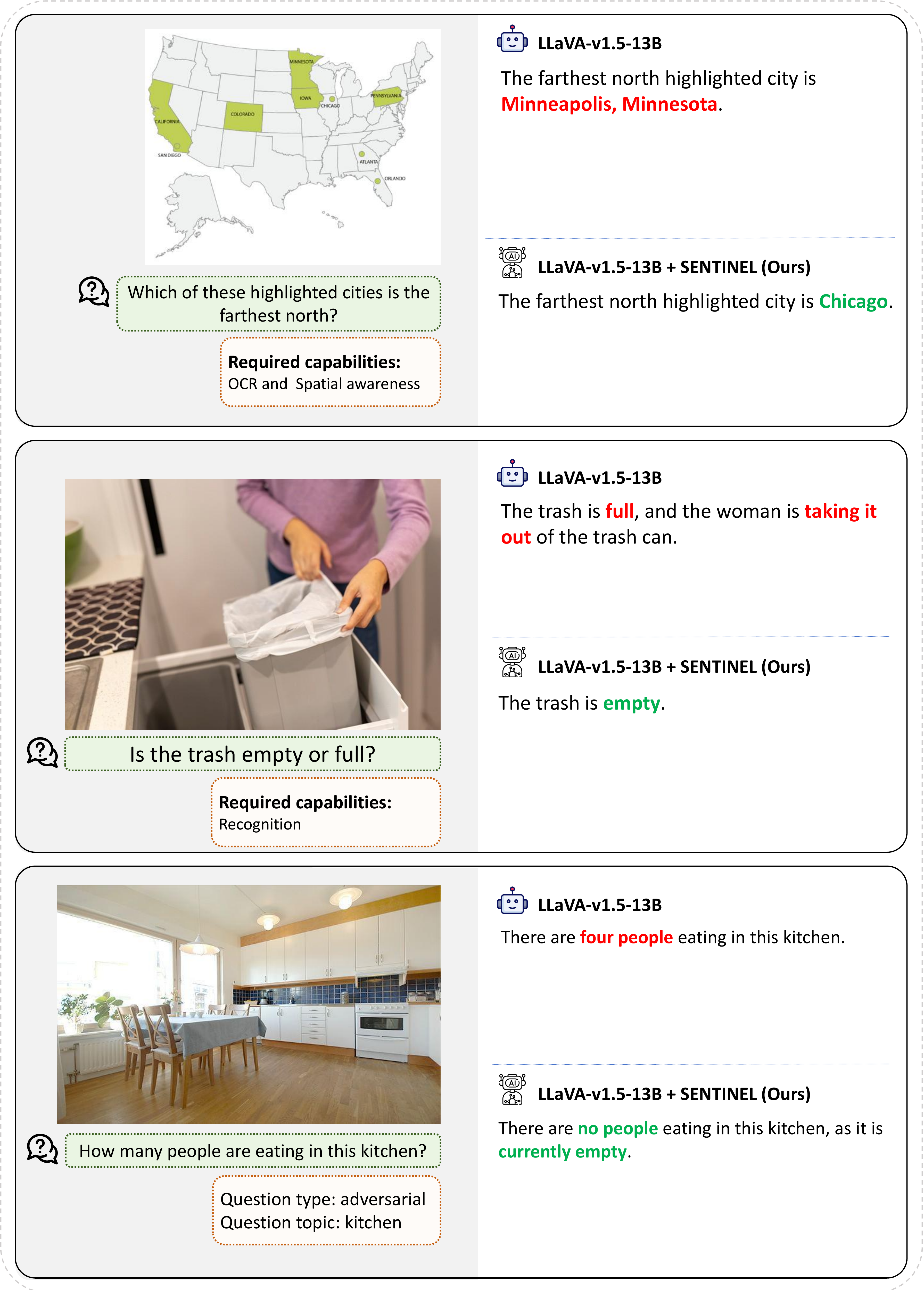}
    \vspace{-5pt}
    \caption{\textbf{Comparing visual question answering between SENTINAL and its base model LLaVA-v1.5-13B.} Our method effectively mitigates hallucinations while enhancing the general performance of the base model, leading to more accurate answers.}
    \label{fig:case_13B_3}
    \vspace{-10pt}
\end{figure*}

\end{document}